\documentclass{article}

\usepackage[main,preprint]{neurips_2026}

\usepackage[utf8]{inputenc} %
\usepackage[T1]{fontenc}    %
\usepackage{hyperref}       %
\usepackage{url}            %
\usepackage{booktabs}       %
\usepackage{amsfonts}       %
\usepackage{nicefrac}       %
\usepackage{microtype}      %
\usepackage{xcolor}         %

\usepackage{graphicx}
\usepackage{subcaption}
\usepackage{amsmath}
\usepackage{amssymb}
\usepackage{mathtools}
\usepackage{amsthm}

\usepackage{multirow}
\usepackage{enumitem}
\usepackage{placeins}
\usepackage{tabularx}

\usepackage{algorithm}
\usepackage{algpseudocode}

\usepackage[most]{tcolorbox}
\usepackage{listings}

\usepackage[capitalize,noabbrev]{cleveref}

\definecolor{DarkRoyalBlue}{RGB}{0,51,102}
\definecolor{Teal}{RGB}{0,128,128}
\definecolor{ForestGreen}{RGB}{34,139,34}
\definecolor{lightgray}{RGB}{245,245,245}

\newcommand{\best}[1]{\textcolor{red}{\textbf{#1}}}
\newcommand{\gain}[1]{\textcolor{ForestGreen}{\textbf{#1}}}

\lstdefinestyle{pddlStyle}{
  basicstyle=\ttfamily\small\color{black},
  keywordstyle=\color{DarkRoyalBlue}\bfseries,
  identifierstyle=\color{Teal},
  commentstyle=\color{ForestGreen},
  frame=none,
  breaklines=true,
  showstringspaces=false,
  tabsize=2
}

\tcbset{
  myboxstyle/.style={
    enhanced,
    colback=white,
    colframe=DarkRoyalBlue,
    boxrule=1pt,
    arc=1.5mm,
    boxsep=5pt,
    left=8pt,
    right=8pt,
    top=5pt,
    bottom=5pt,
    fonttitle=\bfseries,
    coltitle=white,
    colbacktitle=DarkRoyalBlue,
    title={A PDDL Problem Example Generated by an LLM}
  }
}

\title{Grasp-Then-Plan with Failure Attribution: \\
  A Closed Two-Stage Framework for Precise and Generalizable Robotic Manipulation}

\author{
{\small\bfseries Jiahao Xu$^{1}$\thanks{Equal contributions.} \enspace Peiyuan Wang$^{1}$\footnotemark[1] \enspace Hanzhuo Zhang$^{1}$ \enspace Zihao Yu$^{1}$ \enspace Tianyu Fu$^{1}$} \\
{\small\bfseries Hao Chen$^{1}$ \enspace Xuanhao Xiang$^{1}$ \enspace Zixuan Li$^{2}$ \enspace Jianbo Yu$^{1}$ \enspace Chenchen Fu$^{1}$ \enspace Wanyuan Wang$^{1}$\thanks{Corresponding Author.}} \\
{\small\normalfont $^{1}$School of Computer Science and Engineering, Southeast University, China} \\
{\small\normalfont $^{2}$School of Computer Science and Software Engineering, Hohai University, China} \\
}

\makeatletter
\renewcommand{\@noticestring}{Accepted at NeurIPS 2026.}
\makeatother
\hypersetup{hidelinks}

\begin{document}

\maketitle

\begin{abstract}
  In robotic manipulation, the tight coupling between grasping and motion planning often obscures the true source of failure, leading to inefficient trial-and-error. To enable efficient long-horizon manipulation, we propose GTP-FA (Grasp-Then-Plan with Failure Attribution), a task-oriented two-stage ‘grasp-then-plan’ framework that generates grasp candidates and performs downstream motion planning conditioned on the selected grasp. Given a failed manipulation trajectory, we learn a failure attribution model that generalizes to unseen grasps and produces a stable distribution over failure modes for diagnosis-guided optimization. Based on these attribution results, we then optimize both modules in a diagnosis-driven manner: on the grasping side, we inject task-level priors and risk penalties into grasp candidate scoring and optimization to suppress unstable or task-incompatible grasps; on the planning side, we target high-risk initial states through data collection and fine-tuning to address genuine planning bottlenecks. We evaluate the proposed framework in both simulation and real-robot experiments, and show that GTP-FA improves the corresponding base learners across RL, IL, diffusion-policy, and VLA-based settings, achieving substantially higher overall task success rates. Project page: \href{https://sites.google.com/view/gtp-fa/}{here}.
  
\end{abstract}

\section{Introduction}
\label{sec:intro}

In long-horizon robotic manipulation, the grasp often determines whether a task is feasible~\citep{mao2025universal,ma2024generalizing,tang2025foundationgrasp,wei2024grasp}. For constrained placement or tool-use tasks, even a strong downstream planning/control policy can fail if the selected grasp occludes critical interaction regions, restricts the reachable motion space, or otherwise conflicts with the task objective~\citep{tang2023task}. In practice, such grasp-induced failures are often observed only as ``planning failures'': the policy is forced to act from an infeasible or unfavorable initial condition, leading to collisions, compensatory behaviors, and inefficient learning signals~\citep{huang2025fail2progress,duan2024aha,lin2025failsafe}.

A natural remedy is to decompose manipulation into ``grasp-then-plan.'' However, decomposition alone does not remove the coupling between grasping and downstream execution, because grasping is not task-agnostic: the task dictates how the object should be grasped~\citep{wen2022catgrasp}. Optimizing only geometric feasibility or grasp stability may produce grasps that are stable but task-incompatible. Moreover, friction, material properties, and execution noise can make grasp stability fragile under distribution shift~\citep{mousavian20196,fang2020graspnet,weng2022neural}. Although reinforcement learning, imitation learning, and vision-language-action models have substantially improved robotic policy learning~\citep{chen2025pirl,xu2025stare,black2410pi0,hu2024data,escoriza2025multi,intelligence2025pi,intelligence2025pi_,tao2024reverse,ball2023efficient}, failure data in strongly coupled grasp--execution settings still entangles grasp-side and downstream-policy causes. Without explicit failure attribution, end-to-end fine-tuning may incorrectly assign grasp errors to the downstream policy, contaminating training data and weakening generalization. Thus, improving robustness requires answering a basic question: \emph{should a failure be attributed to grasping or to downstream planning/control?}

We propose \textbf{GTP-FA} (\textbf{G}rasp-\textbf{T}hen-\textbf{P}lan with \textbf{F}ailure \textbf{A}ttribution), a closed-loop framework that links grasp selection to downstream outcomes and uses attribution to apply optimization pressure to the correct module. Concretely, we learn a failure attribution model that maps an execution outcome to a structured distribution over failure modes and estimates how responsibility is shared between grasping and planning. To make attribution reliable under unseen grasps, we impose a local-consistency prior in the diagnostic space. This diagnostic signal then guides what to optimize (grasping vs. planning), what data to collect, and which failure types to prioritize.

Our contributions are threefold: (i) we introduce a task-centric Grasp-Then-Plan closed-loop framework that explicitly binds grasp selection to downstream success; (ii) we develop a failure attribution mechanism and improve its stability under unseen grasps to ensure optimization pressure is applied to the correct module; and (iii) we propose a diagnosis-driven bidirectional optimization strategy that routes updates to the grasp side or downstream policy side based on the attributed failure source, improving task success across RL, imitation-learning, diffusion-policy, and VLA-based learners.

\section{Related Work}
\label{sec:related_work}

\textbf{Six-DoF grasp synthesis and task-oriented grasping.}
Large-scale benchmarks have driven rapid progress in 6-DoF grasping, yielding RGB-D/point-cloud-based methods for grasp estimation and generation, including generative and implicit formulations for handling occlusion and clutter \citep{fang2020graspnet,ma2024generalizing,mousavian20196,weng2022neural}.
However, grasp stability alone does not ensure task success. Task-oriented grasping therefore incorporates task context, e.g., conditioning grasps on language or vision--language inputs \citep{tang2023task,wei2024grasp,wen2022catgrasp}, or injecting semantic and geometric priors via foundation models for improved generalization \citep{tang2025foundationgrasp,huang2024copa}. VLM-based grasp correction further revises grasps at execution time \citep{lee2025graspcorrect}.
In contrast, our work links grasp selection and its induced failure modes to downstream planning failures through explicit diagnostic signals, enabling more targeted optimization of both grasping and planning.

\textbf{Failure detection, diagnosis, reasoning, and recovery.}
Prior work on long-horizon manipulation addresses failure handling through detection \citep{inceoglu2023multimodal}, diagnosis and attribution \citep{sagar2024mystery}, and reasoning or recovery \citep{lin2025failsafe,duan2024aha,agia2024unpacking,huang2025fail2progress}. While these approaches improve failure observability and recovery, they offer limited system-level credit assignment in grasp--plan pipelines, leaving unclear whether failures stem from grasping or planning.
In contrast, we introduce grasp–planning-aware failure attribution that explicitly distinguishes grasp-side and planning-side failures, and use the attribution both as supervision and as an optimization-routing signal, enabling a diagnosis-driven bidirectional improvement loop.

\textbf{Coupling between grasp selection and downstream policy execution.}
Grasp--planning systems are commonly implemented as two-stage pipelines that select a grasp and then plan motions conditioned on it, often leading to mismatches such as feasible grasps with infeasible plans, or vice versa. To reduce this coupling, some methods jointly optimize grasp and motion using unified objectives or differentiable costs \citep{urain2022se}. Others address long-horizon decision making via reinforcement or imitation learning, including behavior cloning, diffusion policies, and RL-based policy optimization, leveraging demonstrations or online adaptation to improve robustness and generalization \citep{sontakke2023roboclip,kim2025subtask,ma2023liv,yuan2024policy,lu2024koi}. In parallel, recent vision--language--action models improve generalization through instruction tuning and structured prompting \citep{niu2024llarva,liu2025robodexvlm}.
However, existing methods are often tied to a specific grasp--motion optimization formulation or a particular class of policy learner, and lack a general failure-diagnosis and optimization-routing mechanism that can be shared across base algorithms. In contrast, GTP-FA provides a base-learner-agnostic, diagnosis-driven interface: it distinguishes grasp-side failures from downstream-policy failures, routes optimization pressure to the corresponding module, and can be applied to RL, imitation-learning, diffusion-policy, and VLA-based learners.

\section{Problem Definition}
\label{sec:Problem Definition}
We study \textbf{task-oriented grasp-then-plan} manipulation with a two-finger gripper and a fixed grasp during downstream execution (see Appendix~\ref{app:limitations_impacts} for scope). Given a language instruction $\tau$ and a visual observation $o$ (possibly multi-view RGB/RGBD), the system first selects a grasp
\begin{equation}
g \in \mathcal{G} \subset \mathrm{SE}(3)\times\mathbb{R},
\end{equation}
where $\mathrm{SE}(3)$ specifies the 6-DoF gripper pose, and the additional $\mathbb{R}$ dimension denotes the gripper width or other continuous grasp parameters. Conditioned on the fixed grasp, a downstream planning/control policy then produces an action sequence $a_{1:H}$:
\begin{equation}
g \sim \pi_g(\,\cdot \mid \tau,o), 
\qquad
a_t \sim \pi(\,\cdot \mid s_t,o_t,\tau,g).
\end{equation}
The outcome of an episode is denoted by $y\in\{0,1\}$, indicating whether the task is successfully completed.

Because grasping and planning are tightly coupled, failures are often difficult to attribute directly. We therefore structure failures using a failure-mode set
\begin{equation}
\mathcal{M}=\{\mathrm{FM\text{-}G1},\mathrm{FM\text{-}G2},\mathrm{FM\text{-}P}\},
\end{equation}
where \textbf{FM-G1} denotes task-irrelevant grasp contact/functionality, \textbf{FM-G2} denotes unstable grasping or slippage during execution, and \textbf{FM-P} denotes  downstream planning/control or policy insufficiency. Our goal is to learn an attribution model $D$ that outputs a probability distribution over failure modes for a given rollout:
\begin{equation}
D(\tau,o,g,\xi)\ \rightarrow\ p(m\mid \tau,o,g,\xi),
\qquad
m\in\mathcal{M},
\end{equation}
where $\xi$ is an observable execution summary (e.g., drop/collision indicators and end-effector goal errors). This soft distribution allocates operational responsibility for targeted optimization; its categories need not correspond to mutually exclusive physical causes.

Overall, we aim to maximize task success under task and environment variations:
\begin{equation}
\max_{\pi_g,\pi}\ 
\mathbb{E}_{(\tau,o)\sim \mathcal{D}_{\mathrm{task}}}
\!\left[\Pr\!\left(y=1\mid \tau,o\right)\right],
\end{equation}
where $\mathcal{D}_{\mathrm{task}}$ denotes the task and environment distribution.

\section{Method}\label{sec:Method}

\subsection{Overview}

\begin{figure*}[t]
    \centering
    \includegraphics[width=\linewidth]{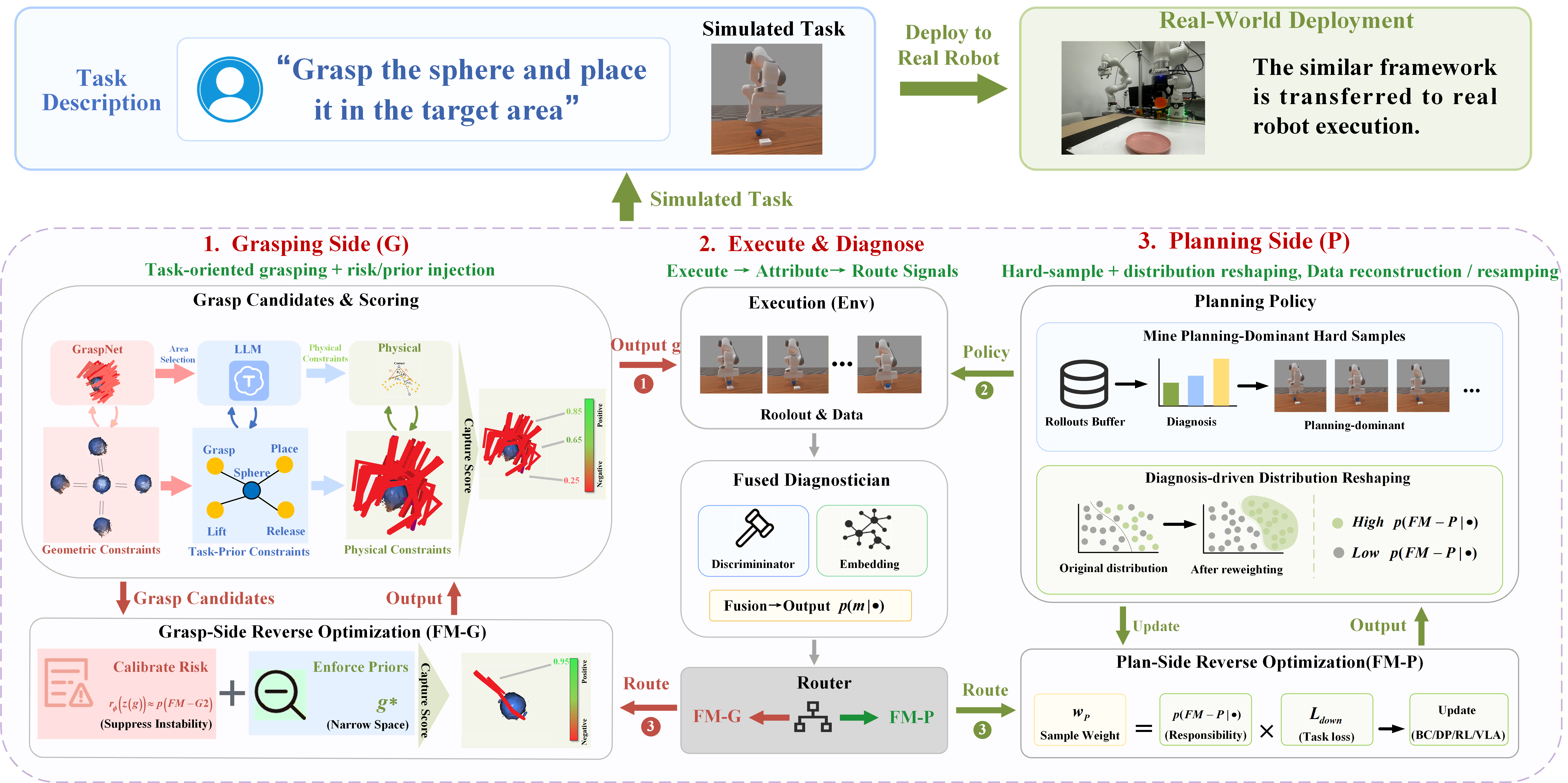}
    \caption{
Overview of GTP-FA.
Given a task description, GTP-FA performs task-aware grasp optimization, executes the selected grasp, diagnoses the resulting outcome, and routes optimization signals to either the grasp side or the planning side in a closed-loop execute--diagnose--update pipeline.
The optimized grasp-then-plan interface is then transferred to real-robot deployment.
}
    \label{fig:ove}
\end{figure*}

GTP-FA formulates task-oriented grasp-then-plan manipulation as a closed-loop execute--diagnose--update process. Because the same observable failure may arise from grasp-functional mismatch, grasp instability, or downstream policy insufficiency, naive end-to-end optimization can misattribute grasp-side errors to the planning/policy side. To address this issue, GTP-FA first uses a Failure Attribution Discriminator $D$ to predict a structured distribution over failure modes from a single execution, and introduces a local-consistency prior through a Grasp-Conditioned Representation Space $E$ to stabilize attribution for unseen grasps. The fused diagnosis is then used as an optimization-routing signal: when failures are mainly attributed to the grasp side, the system updates grasp selection with task priors and risk penalties; when failures are mainly attributed to the planning/policy side, the system improves the downstream module through hard-P mining, distribution reshaping, data resampling, and responsibility-weighted learning. The overall GTP-FA optimization pipeline is shown in Figure~\ref{fig:ove}.

\subsection{Failure Attribution Discriminator $D$}
\label{sec:failure_attribution}

Following the failure-mode definition in Sec.~\ref{sec:Problem Definition}, we learn a Failure Attribution Discriminator $D$ to perform post-hoc diagnosis for each execution. Given the task instruction, observation, selected grasp, and execution summary, $D$ estimates whether a failure mainly originates from grasp-functional mismatch, grasp instability, or downstream planning/control insufficiency. The execution summary $\xi$ includes events such as slip/drop indicators, collision events, and end-effector errors. This diagnostic distribution is used to assign responsibility between the grasp side and the downstream policy side, avoiding the incorrect use of grasp-induced failures as supervision for downstream policy learning.

Since manually obtaining consistent failure-attribution labels can be expensive and subjective, we construct weak supervision through controlled repeated trials under fixed grasp conditions. For each grasp condition $g$, we run the downstream policy $K$ times while only changing the random seed or adding small perturbations, and compute the empirical terminal success rate:
\begin{equation}
\hat{q}_{\mathrm{end}}(g)
=
\frac{1}{K}
\sum_{k=1}^{K}
\mathbb{I}\{y_{\mathrm{end}}^{(k)}=1\}.
\end{equation}
Intuitively, $\hat{q}_{\mathrm{end}}(g)$ measures whether the task is usually achievable under the current grasp condition. We then generate high-confidence pseudo-labels: clear functional/contact-region mismatch is labeled as FM-G1; For failed executions under a non-G1 grasp condition, a high terminal success rate indicates that the task is generally realizable under this grasp; therefore, the observed failure is more likely attributed to FM-P. In contrast, a low terminal success rate together with slip or drop events indicates FM-G2. Ambiguous samples are discarded or handled with soft labels to reduce noise.

After training, the output of $D$ is used for failure explanation, responsibility assignment, and optimization routing. When $p_D(\mathrm{FM\text{-}P})$ is high, the system emphasizes downstream planning/policy updates; when $p_D(\mathrm{FM\text{-}G2})$ is high, the system suppresses unstable grasps and triggers grasp-side optimization. Details of weak-label construction, threshold sensitivity, discriminator inputs, and training are provided in Appendix~\ref{app:fa_details}.

\subsection{Failure Attribution Generalization to Unseen Grasps}
\label{sec:unseen_grasp_generalization}

Since the failure attribution discriminator $D$ is trained from weak pseudo-labels collected on a finite set of post-grasp initializations, its predictions can become unstable when test-time grasps involve unseen contact regions, wrist poses, or post-grasp states. To improve attribution reliability under such grasp distribution shifts, we introduce a Grasp-Conditioned Diagnostic Representation Space $E$. The key idea is that grasp conditions inducing similar post-grasp initialization states should exhibit similar local failure-mode distributions.

For each grasp condition $g$, we extract a grasp-conditioned initialization feature $x_g$, which describes the post-grasp state induced by the selected grasp, such as object pose, relative gripper pose, contact region, or other observable state features. We map this feature into a diagnostic embedding space:
\begin{equation}
e = f_{\theta}(x_g) \in \mathbb{R}^{d}.
\end{equation}
The embedding model is trained with weak failure labels and success/failure signals so that diagnostically similar grasp conditions are locally close in the representation space.

Given an unseen grasp at test time, we retrieve its $k$ nearest neighbors $\mathcal{N}_k(e)$ from an embedding bank built from training and rollout samples, and construct a neighborhood diagnostic prior by weighted voting:
\begin{equation}
\tilde{p}_{E}(m\mid e)
=
\sum_{j\in\mathcal{N}_k(e)}
w_j \mathbb{I}\{m_j=m\},
\qquad
w_j =
\frac{\exp(\mathrm{sim}(e,e_j)/t)}
{\sum_{\ell\in\mathcal{N}_k(e)}\exp(\mathrm{sim}(e,e_{\ell})/t)} .
\end{equation}
Here, $m_j$ denotes the failure-mode label of a neighboring sample, and $w_j$ assigns larger weights to more similar grasp conditions. This prior provides a smoother local diagnostic estimate when $D$ is uncertain on an unseen grasp.

\begin{figure*}[t]
    \centering
    \includegraphics[width=0.9\linewidth]{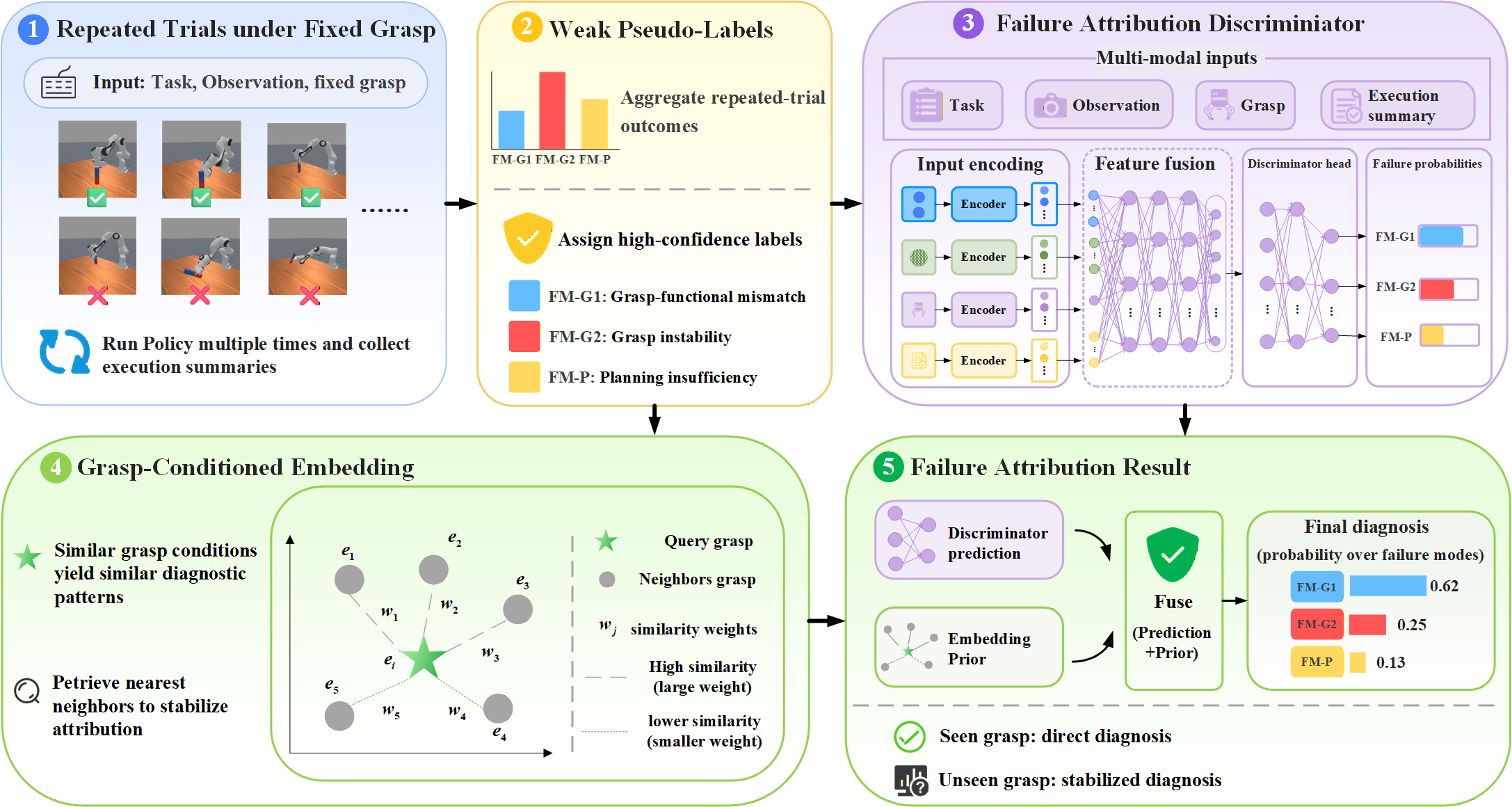}
    \caption{
Failure attribution and diagnostic signal generation.
Given a task instruction, visual observation, candidate grasp, and execution summary, the discriminator $D$ performs post-hoc diagnosis for a single execution and outputs a probability distribution over three failure modes.
To improve attribution stability for unseen grasps, the system uses the grasp-conditioned representation space $E$ to construct a neighborhood diagnostic prior, which is fused with the output of $D$ to produce a reliable diagnostic signal for grasp-side optimization and planning-side downstream policy learning.
}
    \label{fig:pa1}
\end{figure*}

We then fuse this neighborhood prior with the discriminator prediction:
\begin{equation}
p_{\mathrm{fuse}}(m\mid \tau,o,g,\xi,e)
=
(1-\alpha)p_D(m\mid\tau,o,g,\xi)
+
\alpha \tilde{p}_E(m\mid e),
\end{equation}
where $\alpha$ is determined by the confidence of $D$. When $D$ is confident, the system relies more on the direct discriminator prediction; when $D$ is uncertain, it relies more on the neighborhood diagnostic prior. The fused attribution $p_{\mathrm{fuse}}$ provides a more stable diagnostic signal for the grasp-side and planning-side optimization routing in Sec.~\ref{sec:bidirectional_optimization}. Figure~\ref{fig:pa1} illustrates the overall attribution pipeline, including weak-label construction, discriminator prediction, grasp-conditioned embedding, and D/E fusion. Details of the embedding objective, nearest-neighbor prior, and fusion implementation are provided in Appendix~\ref{app:de_fusion}.

\subsection{Diagnosis-driven Bidirectional Optimization}
\label{sec:bidirectional_optimization}

We use the fused attribution result from Sec.~\ref{sec:unseen_grasp_generalization} as an optimization-routing signal to decide whether a failure should be mainly assigned to the grasp side or the downstream planning/policy side. Given an execution sample $(\tau,o,g,\xi,y)$, the fused diagnostic model outputs a failure-mode distribution $p_{\mathrm{fuse}}(m)$. For brevity, we write $p_{\mathrm{fuse}}(m)$ to denote 
$p_{\mathrm{fuse}}(m\mid \tau,o,g,\xi,e)$ for the current sample. We define the grasp-side and planning-side responsibility weights as

\begin{equation}
w_G
=
p_{\mathrm{fuse}}(\mathrm{FM\text{-}G1})
+
p_{\mathrm{fuse}}(\mathrm{FM\text{-}G2}),
\qquad
w_P
=
p_{\mathrm{fuse}}(\mathrm{FM\text{-}P}).
\end{equation}

Here, $w_G$ indicates that the failure is more likely caused by grasp-functional mismatch or grasp instability, while $w_P$ indicates downstream planning/control insufficiency. GTP-FA routes optimization pressure according to these weights, instead of applying undifferentiated end-to-end updates to all failed samples.

\paragraph{Grasp-side update.}
When $w_G$ is high, the system treats the failure as more likely induced by the grasp condition and updates grasp selection and scoring. Specifically, we first train a lightweight risk predictor from the fused diagnostic signal to estimate the instability risk of each candidate grasp:
\begin{equation}
r_\phi(z(g))
\approx
p_{\mathrm{fuse}}(\mathrm{FM\text{-}G2}\mid \tau,o,g,\xi,e),
\end{equation}
where $z(g)$ denotes the internal representation or scoring feature of the candidate grasp. We then augment the original grasp score $s_{\mathrm{base}}(g;o)$ with a task-prior term and a diagnostic risk penalty:
\begin{equation}
g^*
=
\arg\max_{g\in\mathcal{G}(o)}
\Big[
s_{\mathrm{base}}(g;o)
+
\lambda s_{\mathrm{prior}}(g;\pi_{\mathrm{VLM}}(\tau))
-
\beta r_{\phi}(z(g))
\Big].
\end{equation}
Here, $s_{\mathrm{prior}}$ denotes a coarse task-prior constraint produced by the VLM prior, which suppresses grasps that occupy functional regions, block release motions, or hinder subsequent manipulation; 
$\pi_{\mathrm{VLM}}(\tau)$ denotes a task-prior generator that maps the language instruction to coarse functional-region or release-motion constraints.
$r_\phi(z(g))$ down-weights unstable grasps that are likely to slip or drop; and $\lambda$ and $\beta$ control the strengths of the task prior and risk suppression, respectively. Thus, the grasp module optimizes not only geometric stability, but also task feasibility and downstream execution requirements. More details on grasp-side risk calibration, task-prior scoring, and VLM prompt construction are provided in Appendices~\ref{app:grasp_side_scoring} and~\ref{app:vlm_task_prior}.

\paragraph{Planning-side update.}
When $w_P$ is high and the grasp-side risk is low, the sample is more likely to correspond to a planning-dominant hard start. We aggregate the planning-failure risk for each candidate grasp condition $g$ from rollouts:
\begin{equation}
r_P(g)
=
\mathbb{E}
\left[
p_{\mathrm{fuse}}(\mathrm{FM\text{-}P})
\mid g
\right],
\end{equation}
and combine it with low FM-G1/FM-G2 risks to obtain the planning-dominant hard-P subset $\mathcal{G}_{\mathrm{hardP}}$. This subset represents initial states where the grasp side is relatively feasible but downstream planning or policy execution remains difficult. Detailed screening criteria and threshold settings are provided in Appendix~\ref{app:diag_bidir_formulas}.

During downstream policy training, we increase the coverage of hard-P samples through distribution reshaping:
\begin{equation}
g
\sim
(1-\rho)\,\mathrm{Unif}(\mathcal{G}_{\mathrm{full}})
+
\rho\,\mathrm{Unif}(\mathcal{G}_{\mathrm{hardP}}),
\end{equation}
where $\rho$ controls the degree to which the training distribution is biased toward planning-bottleneck samples. During standard evaluation, we always set $\rho=0$, ensuring that all methods are compared under the same evaluation distribution. For interactive reinforcement-learning methods such as PPO and SAC, this corresponds to start-state resampling during training. For data-driven methods such as BC, DP, and VLA, it corresponds to diagnosis-driven reconstruction and resampling of training data, rollout data, or fine-tuning data.

We further use the planning-side responsibility weight $w_P$ to reweight the policy objective:
\begin{equation}
\min_{\theta}
\mathbb{E}
\left[
w_P \cdot
\mathcal{L}_{\mathrm{down}}(\theta;\tau,o,g,\xi,y)
\right],
\end{equation}
where $\mathcal{L}_{\mathrm{down}}$ denotes the training objective of the corresponding base learner, such as an RL loss, behavior-cloning loss, diffusion-policy training loss, or VLA fine-tuning objective. When $w_G$ is high, the contribution of the sample to planning-side updates is reduced, preventing grasp-induced failures from being incorrectly used to train the downstream policy.

Overall, GTP-FA iterates an execute--diagnose--update loop: execution produces trajectories and event summaries, diagnosis estimates failure responsibility, and the update stage assigns optimization pressure to either the grasp side or the planning side. Through this diagnosis-driven bidirectional update, the system reduces training contamination caused by misattribution and achieves more stable performance improvements in strongly coupled long-horizon manipulation tasks.

\section{Experiments}
\label{sec:exp_setup}

\subsection{Experimental Setup}

We evaluate GTP-FA in both simulated and real-robot environments. The simulation experiments are conducted on eight long-horizon manipulation tasks from ManiSkill3: StackCube, PokeCube, LiftPegUpright, PlaceSphere, PickCube, PushCube, PullCube, and PullCubeTool. These tasks cover diverse grasp--execution coupling scenarios, including stacking, pushing, pulling, placing, picking, pose adjustment, and tool use. The real-robot experiments are conducted on a Franka Research 3 platform to verify the feasibility of our method in physical environments. For VLA fine-tuning, we use 100 converted expert trajectories in simulation and 300 real expert trajectories per task (1,500 across five tasks) for the real-robot $\pi_{0.5}$ / GTP-FA-$\pi_{0.5}$ setting; the detailed task setup and execution protocol are provided in Sec.~\ref{sec:real_robot} and Appendix~\ref{app:experimental_protocols}.

For VLA-based experiments, we instantiate the VLA learner with $\pi_{0.5}$ using LoRA fine-tuning, and report the corresponding rows as $\pi_{0.5}$ variants in the main tables. For each learner, we evaluate the original policy, module-level ablation variants, and the corresponding full GTP-FA version. We denote the ablation setting by a two-bit suffix \texttt{GP}, where \texttt{G} indicates whether grasp-side optimization is enabled and \texttt{P} indicates whether planning-side or downstream-policy optimization is enabled. Thus, \texttt{00} denotes the original policy, \texttt{01} denotes planning-side-only optimization, \texttt{10} denotes grasp-side-only optimization, and \texttt{11} denotes naive joint optimization without failure attribution. Full GTP-FA further adds failure attribution and diagnostic routing on top of this setting.

In simulation, we report two metrics, \texttt{success\_once} and \texttt{success\_at\_end}. The former measures whether an episode ever reaches a successful state, while the latter measures whether the task remains successful at the end of the episode. Since VLA evaluation records only terminal task success, we report only \texttt{success\_at\_end} for VLA. Real-robot experiments are evaluated using task-level success rate.

For implementation, PPO, SAC, BC, DP, and their GTP-FA variants are mainly trained and evaluated on a workstation with $2\times$ RTX 4090 GPUs. VLA experiments are conducted on a $3\times$ A100 server due to the higher memory cost of fine-tuning and inference. Real-robot experiments are deployed on a Franka Research 3 platform equipped with a Robotiq gripper and base/wrist D435i cameras, with perception and policy inference running on an RTX 4090 workstation. Hardware differences only affect training and inference throughput; they do not change the task definitions, ablation settings, evaluation protocol, or success metrics. Full training budgets, random seeds, evaluation episodes, compute resources, and key hyperparameters are provided in Appendix~\ref{app:experimental_protocols}.

\begin{table*}[t]
\vspace{-2mm}
\caption{
Simulation results and module ablations on eight ManiSkill3 tasks.
Each entry reports \texttt{success\_once} / \texttt{success\_at\_end}. For VLA-based results, we instantiate the VLA learner with $\pi_{0.5}$ and report only \texttt{success\_at\_end}, since \texttt{success\_once} is unavailable in our VLA evaluation.
The suffix \texttt{GP} indicates whether grasp-side optimization ($G$) and planning-side/downstream optimization ($P$) are enabled: \texttt{00} denotes the original policy, \texttt{01} planning-side-only optimization, \texttt{10} grasp-side-only optimization, and \texttt{11} naive joint optimization without failure attribution.
Full GTP-FA further adds failure attribution and diagnostic routing.
Red numbers indicate the best value for each task and metric when available, and Green values highlight the average gains of full GTP-FA variants.
The last column reports the arithmetic mean of the eight displayed per-task changes relative to the corresponding \texttt{00} baseline, rounded to one decimal place in percentage points (pp); positive and negative values indicate improvement and degradation, respectively.
}
\label{tab:main_results}
\centering
\scriptsize
\setlength{\tabcolsep}{1.6pt}
\renewcommand{\arraystretch}{0.98}

\resizebox{\textwidth}{!}{
\begin{tabular}{@{}lccccccccc@{}}
\specialrule{1.2pt}{0pt}{3pt}

\multirow{2}{*}{\textbf{Method}}
& \multicolumn{8}{c}{\textbf{Task}} 
& \multirow{2}{*}{\textbf{Avg. \(\Delta\) (pp)}} \\
\cmidrule(lr){2-9}
& \textbf{Stack}
& \textbf{Poke}
& \textbf{LiftPeg}
& \textbf{Place}
& \textbf{Pick}
& \textbf{Push}
& \textbf{Pull}
& \textbf{PullTool}
& \\

\specialrule{1.2pt}{3pt}{3pt}

\multicolumn{10}{@{}l}{\textit{Original downstream policies (\texttt{00})}} \\
\textbf{PPO-00}
& 65.2\% / 29.1\%
& 75.0\% / 34.4\%
& \best{99.3\%} / 25.0\%
& 82.6\% / 82.6\%
& 99.2\% / 41.6\%
& \best{100\%} / 51.0\%
& \best{99.3\%} / 41.2\%
& 29.4\% / 29.3\%
& 0.0 / 0.0 \\

\textbf{BC-00}
& 0.0\% / 0.0\%
& 65.3\% / 46\%
& 41.3\% / 2.3\%
& 0.0\% / 0.0\%
& 0.1\% / 0.0\%
& 4.5\% / 4.3\%
& 71.3\% / 27.8\%
& 0.1\% / 0.1\%
& 0.0 / 0.0 \\

\textbf{DP-00}
& 0.0\% / 0.0\%
& 53.1\% / 40.5\%
& 6.9\% / 0.3\%
& 0.0\% / 0.0\%
& 0.2\% / 0.1\%
& 0.0\% / 0.0\%
& 0.0\% / 0.0\%
& 0.0\% / 0.0\%
& 0.0 / 0.0 \\

\textbf{SAC-00}
& 22.8\% / 14.3\%
& 64.3\% / 64.3\%
& 88.1\% / 75.3\%
& 86.7\% / 86.2\%
& 99\% / 92.4\%
& 99\% / 93\%
& 99.2\% / 93.1\%
& 65.3\% / 60.3\%
& 0.0 / 0.0 \\

\textbf{$\pi_{0.5}$-00}
& -- / 89.3\%
& -- / 46.0\%
& -- / 20.3\%
& -- / 73.3\%
& -- / 0.0\%
& -- / 93.3\%
& -- / 78.7\%
& -- / 19.0\%
& -- / 0.0 \\

\specialrule{0.8pt}{3pt}{3pt}

\multicolumn{10}{@{}l}{\textit{Planning-side-only optimization (\texttt{01})}} \\
\textbf{PPO-01}
& 52.1\% / 42.8\%
& 16.3\% / 9.9\%
& 3.1\% / 0.8\%
& 30.8\% / 28.5\%
& 35.8\% / 4.2\%
& 63.5\% / 14\%
& 35.8\% / 4.2\%
& 5.3\% / 5.3\%
& -50.9 / -28.1 \\

\textbf{BC-01}
& 0.0\% / 0.0\%
& 1.7\% / 0.6\%
& 0.8\% / 0.0\%
& 0.1\% / 0.1\%
& 0.0\% / 0.0\%
& 25.2\% / 4.5\%
& 0.2\% / 0.2\%
& 0.0\% / 0.0\%
& -19.3 / -9.4 \\

\textbf{DP-01}
& 7.2\% / 4.3\%
& 6.3\% / 3.8\%
& 1.1\% / 0.1\%
& 1.1\% / 0.0\%
& 26.6\% / 26.5\%
& 15.8\% / 14.7\%
& 63.1\% / 62.3\%
& 0.0\% / 0.0\%
& +7.6 / +8.9 \\

\textbf{SAC-01}
& 17.6\% / 9.8\%
& 27.4\% / 24.2\%
& 10.4\% / 6.2\%
& 25.7\% / 23.4\%
& 32.6\% / 29.2\%
& 26.3\% / 20.3\%
& 42.1\% / 21.9\%
& 8.3\% / 7.2\%
& -54.3 / -54.6 \\

\textbf{$\pi_{0.5}$-01}
& -- / 36.3\%
& -- / 13.3\%
& -- / 1.7\%
& -- / 21.3\%
& -- / 0.0\%
& -- / 71.0\%
& -- / 74.7\%
& -- / 0.0\%
& -- / -25.2 \\

\specialrule{0.8pt}{3pt}{3pt}

\multicolumn{10}{@{}l}{\textit{Grasp-side-only optimization (\texttt{10})}} \\
\textbf{PPO-10}
& 23.2\% / 9.4\%
& 1.7\% / 1.3\%
& 12.2\% / 3.7\%
& 0.0\% / 0.0\%
& 4.5\% / 0.6\%
& 9.3\% / 7.8\%
& 4.5\% / 0.6\%
& 1.6\% / 1.6\%
& -74.1 / -38.7 \\

\textbf{BC-10}
& 0.0\% / 0.0\%
& 0.2\% / 0.2\%
& 0.5\% / 0.1\%
& 0.0\% / 0.0\%
& 0.3\% / 0.2\%
& 0.0\% / 0.0\%
& 0.0\% / 0.0\%
& 1.3\% / 1.2\%
& -22.5 / -9.9 \\

\textbf{DP-10}
& 0.0\% / 0.0\%
& 0.0\% / 0.0\%
& 0.0\% / 0.0\%
& 0.0\% / 0.0\%
& 0.2\% / 0.1\%
& 0.0\% / 0.0\%
& 0.0\% / 0.0\%
& 3.5\% / 3.4\%
& -7.1 / -4.7 \\

\textbf{SAC-10}
& 18.4\% / 7.7\%
& 1.5\% / 1.2\%
& 7.9\% / 6.4\%
& 1.2\% / 0.3\%
& 51.7\% / 50.5\%
& 58.1\% / 52.4\%
& 21.5\% / 20.3\%
& 8.6\% / 4.1\%
& -56.9 / -54.5 \\

\textbf{$\pi_{0.5}$-10}
& -- / 0.0\%
& -- / 7.3\%
& -- / 0.0\%
& -- / 10.3\%
& -- / 0.0\%
& -- / 28.7\%
& -- / 5.3\%
& -- / 1.3\%
& -- / -45.9 \\

\specialrule{0.8pt}{3pt}{3pt}

\multicolumn{10}{@{}l}{\textit{Naive grasp--plan optimization without attribution (\texttt{11})}} \\
\textbf{PPO-11}
& 86.4\% / 68.7\%
& 34.8\% / 34.5\%
& 88.2\% / 41.3\%
& 77.8\% / 75.8\%
& 18.5\% / 6.5\%
& 0.3\% / 0.1\%
& 18.5\% / 6.5\%
& 53.4\% / 53.2\%
& -34.0 / -6.0 \\

\textbf{BC-11}
& 24.2\% / 14.2\%
& 41.2\% / 34.6\%
& 48.0\% / 22.0\%
& 36.1\% / 36.1\%
& 10.8\% / 8.1\%
& 68.7\% / 56.7\%
& 90.1\% / 66.1\%
& 34.4\% / 33.7\%
& +21.4 / +23.9 \\

\textbf{DP-11}
& 6.3\% / 3.4\%
& 32.7\% / 25.8\%
& 1.5\% / 0.3\%
& 0.0\% / 0.0\%
& 36.1\% / 32.8\%
& 11.7\% / 10.1\%
& 73.7\% / 65.3\%
& 49.1\% / 48.9\%
& +18.9 / +18.2 \\

\textbf{SAC-11}
& 26.6\% / 13.2\%
& 43.6\% / 41.3\%
& 75.5\% / 67.3\%
& 81.7\% / 71\%
& 79.2\% / 77.4\%
& 91.6\% / 90.4\%
& 93.4\% / 90.2\%
& 68.2\% / 63.1\%
& -8.1 / -8.1 \\

\textbf{$\pi_{0.5}$-11}
& -- / 92.3\%
& -- / 53.7\%
& -- / 31.0\%
& -- / 85.3\%
& -- / 5.3\%
& -- / 92.7\%
& -- / 79.3\%
& -- / 25.0\%
& -- / +5.6 \\

\specialrule{0.8pt}{3pt}{3pt}

\multicolumn{10}{@{}l}{\textit{Diagnosis-driven GTP-FA with failure attribution}} \\
\textbf{GTP-FA-PPO}
& \best{90.4\%} / 80.7\%
& \best{81.2\%} / 72.6\%
& 96.3\% / 48.7\%
& \best{93.1\%} / 89.5\%
& 91.2\% / 80.4\%
& 98.1\% / 68.3\%
& 96.1\% / 71.7\%
& 75.9\% / \best{74.3\%}
& \gain{+9.0 / +31.5} \\

\textbf{GTP-FA-BC}
& 84.5\% / 69.6\%
& 74.7\% / 63.8\%
& 72.4\% / 37.5\%
& 67.3\% / 67.3\%
& 82.7\% / 72\%
& 80.1\% / 69.9\%
& 94.6\% / 85.3\%
& 48.2\% / 48.2\%
& \gain{+52.7 / +54.1} \\

\textbf{GTP-FA-DP}
& 7.1\% / 5.5\%
& 63.1\% / 54.5\%
& 23.3\% / 7.7\%
& 0.5\% / 0.1\%
& 34.9\% / 34.8\%
& 10.5\% / 10.3\%
& 78.0\% / 73.5\%
& 60.1\% / 59.9\%
& \gain{+27.2 / +25.7} \\

\textbf{GTP-FA-SAC}
& 65.1\% / 53.6\%
& 68.0\% / 66.5\%
& 86.5\% / \best{77.2\%}
& 90.6\% / 87.3\%
& \best{99.5\%} / \best{95.2\%}
& 97.2\% / \best{95.6\%}
& 98.4\% / \best{96.8\%}
& \best{77.2\%} / 69.6\%
& \gain{+7.3 / +7.9} \\

\textbf{GTP-FA-$\pi_{0.5}$}
& -- / \best{95.3\%}
& -- / \best{76.7\%}
& -- / 71.3\%
& -- / \best{92.0\%}
& -- / 15.3\%
& -- / 95.3\%
& -- / 85.3\%
& -- / 45.0\%
& \gain{-- / +19.5} \\

\specialrule{1.2pt}{3pt}{0pt}
\end{tabular}
}
\vspace{-3mm}
\end{table*}

\subsection{Results}
\label{sec:results}

Table~\ref{tab:main_results} summarizes the simulation results of GTP-FA on eight ManiSkill3 tasks and five classes of downstream learners. This section focuses on the overall improvement of the full GTP-FA framework over the corresponding original base algorithm (\texttt{00}). Red numbers indicate the best result in each task column, and the last column reports the average change relative to the corresponding \texttt{00} baseline; the changes are reported for each available metric.

Overall, full GTP-FA improves the terminal success rate for PPO, BC, DP, SAC, and the VLA-based $\pi_{0.5}$ learner, indicating that the framework is not tied to a specific policy-learning paradigm.
For data-driven methods such as BC, DP, and $\pi_{0.5}$, the gains are more pronounced, with improvements of +54.1 pp, +25.7 pp, and +19.5 pp, respectively; the gain for SAC is relatively smaller because the original SAC baseline is already strong on several tasks. 

These results support our central hypothesis: in strongly coupled grasp--execution scenarios, simply strengthening the downstream policy is insufficient for stable task improvement. By distinguishing grasp-functional mismatch, grasp instability, and planning/policy insufficiency, and by routing optimization pressure to the corresponding module, GTP-FA achieves more reliable overall improvements across different base learners. More detailed final-performance comparisons, complete training curves, and VLA training diagnostics are provided in Appendix~\ref{app:supp_results}.

\subsection{Ablation Studies}
\label{sec:ablations}

Table~\ref{tab:main_results} further reports module-level ablations: \texttt{01} enables only planning-side optimization, \texttt{10} enables only grasp-side optimization, and \texttt{11} denotes naive grasp--plan joint optimization without failure attribution. Full GTP-FA further incorporates failure attribution and diagnostic routing.

First, planning-side-only optimization cannot reliably resolve grasp--plan coupling. Under \texttt{01}, only DP obtains a positive gain (+8.9 pp), whereas PPO, BC, SAC, and $\pi_{0.5}$ show signed changes of -28.1, -9.4, -54.6, and -25.2 pp, respectively. This indicates that directly updating the downstream policy can contaminate the learning signal when failed samples contain grasp-functional mismatch or grasp instability. Similarly, grasp-side-only optimization is also insufficient: \texttt{10} underperforms the corresponding \texttt{00} baseline for all base learners, with PPO, SAC, and $\pi_{0.5}$ decreasing by -38.7, -54.5, and -45.9 pp, respectively. This shows that improving grasp candidates alone cannot replace downstream policy adaptation.

Second, naive joint optimization \texttt{11} improves BC and DP by +23.9 and +18.2 pp, respectively, but still decreases PPO and SAC by 6.0 and 8.1 pp, while $\pi_{0.5}$ gains only +5.6 pp. This suggests that jointly optimizing grasping and planning is not necessarily stable; without failure attribution, optimization pressure can still be assigned to the wrong module.

In contrast, full GTP-FA achieves positive gains for all base learners. Compared with the \texttt{00} baseline, GTP-FA-PPO, GTP-FA-BC, GTP-FA-DP, GTP-FA-SAC, and GTP-FA-$\pi_{0.5}$ improve by +31.5, +54.1, +25.7, +7.9, and +19.5 pp, respectively. Together with the controlled experiments in Appendix~\ref{app:additional_validation}, these results support the role of failure attribution in coordinating grasp-side and planning-side optimization. 
Additional threshold-sensitivity analyses, closed-loop iteration curves, and complete training dynamics are provided in Appendices~\ref{app:fa_details} and~\ref{app:supp_results}.

\subsection{Real-Robot Experiments}
\label{sec:real_robot}

We further evaluate GTP-FA on a Franka Research 3 robot equipped with a Robotiq gripper and base/wrist Intel RealSense D435i cameras.
We consider five long-horizon real-robot manipulation tasks: placing an orange into a pink tray, stacking a blue cube onto an orange cube, pushing a yellow cube with a stick, pulling a yellow cube with a hook, and grasping the red handle of a gray cup to pour its contents into a blue-gray cup.
For the real-robot VLA setting, we instantiate the base policy with $\pi_{0.5}$ and compare the original $\pi_{0.5}$ baseline with GTP-FA-$\pi_{0.5}$.
Both methods use the same 300 real-robot expert trajectories per task (1,500 in total), so this comparison evaluates the combined grasp-selection and diagnosis-driven optimization pipeline.

\begin{table}[t]
\centering
\caption{
Real-robot evaluation protocol and results.
Each task is evaluated over 50 real-robot trials and reported using task-level success rate.
$\Delta$ denotes the absolute improvement of GTP-FA-$\pi_{0.5}$ over the original $\pi_{0.5}$ baseline in percentage points.
}
\label{tab:real_robot_protocol}
\scriptsize
\setlength{\tabcolsep}{3.5pt}
\renewcommand{\arraystretch}{1.12}
\begin{tabularx}{\linewidth}{Xcccc}
\toprule
\textbf{Task description}
& \textbf{$\pi_{0.5}$}
& \textbf{GTP-FA-$\pi_{0.5}$}
& \textbf{$\Delta$ (pp)}
& \textbf{Trials} \\
\midrule

Place the orange into the pink tray.
&  10\%
& \best{ 92\%}
& \gain{+82}
& 50 \\

Stack the blue cube onto the orange cube.
&  4\%
& \best{ 74\%}
& \gain{+70}
& 50 \\

Pick up the red end of the stick and push the yellow cube into the red target area.
&  24\%
& \best{ 86\%}
& \gain{+62}
& 50 \\

Pick up the red part of the hook and pull the yellow cube into the red target area.
&  16\%
& \best{ 78\%}
& \gain{+62}
& 50 \\

Pick up the red handle of the gray cup and pour the contents into the blue-gray cup.
&  2\%
& \best{ 54\%}
& \gain{+52}
& 50 \\

\midrule
\textbf{Average}
& \textbf{11.2\%}
& \best{76.8\%}
& \gain{+65.6}
& -- \\

\bottomrule
\end{tabularx}
\end{table}

Table~\ref{tab:real_robot_protocol} reports the real-robot evaluation results on all five tasks.
Each task is evaluated over 50 physical trials using task-level success rate.
Overall, the original $\pi_{0.5}$ baseline achieves only 11.2\% average success, whereas GTP-FA-$\pi_{0.5}$ improves the average success rate to 76.8\%, yielding an absolute gain of +65.6 pp.
Specifically, GTP-FA improves orange-to-tray, cube-stacking, \textsc{PokeCube}, \textsc{PullCubeTool}, and \textsc{PourWater} from 10\%, 4\%, 24\%, 16\%, and 2\% to 92\%, 74\%, 86\%, 78\%, and 54\%, respectively.

The low success rate of the original $\pi_{0.5}$ baseline in the real world is mainly due to the fact that these tasks require not only visual-language understanding, but also precise task-compatible grasping and stable post-grasp execution.
In physical environments, camera viewpoint, object scale, occlusion, hand-eye calibration error, contact uncertainty, and object slippage can amplify early grasping errors.
Since the $\pi_{0.5}$ policy directly predicts actions from visual observations and language instructions without explicit grasp-candidate filtering or grasp-conditioned diagnosis, an unsuitable or unstable grasp can easily make later placement, stacking, pushing, pulling, or pouring fail. 
These results suggest that the real-world bottleneck is not only policy expressiveness, but also the lack of an explicit interface for selecting task-compatible grasps before VLA execution.

In contrast, GTP-FA-$\pi_{0.5}$ first selects a grasp that is more compatible with the downstream objective by considering task semantics, grasp candidates, and diagnostic risk, and then hands off a more reliable post-grasp state to the downstream $\pi_{0.5}$ policy.
This reduces the propagation of grasping errors to later stages and is especially beneficial for cube stacking, tool use, and pouring, where the contact region, functional tool end, and object pose are critical.
Thus, the real-robot gains of GTP-FA arise not merely from stronger VLA fine-tuning, but from the coordination between task-aware grasping, failure attribution, and downstream policy optimization.
More real-robot success/failure videos, execution-interface screenshots, and grasp-pose visualizations are provided on the project page:
\href{https://sites.google.com/view/gtp-fa/}{https://sites.google.com/view/gtp-fa/}.

\section{Conclusion}

We presented GTP-FA, a closed-loop Grasp-Then-Plan framework for long-horizon robotic manipulation that addresses failure-attribution ambiguity caused by the strong coupling between grasping and downstream planning/control. By distinguishing grasp-functional mismatch, grasp instability, and planning/policy insufficiency, GTP-FA routes optimization pressure to the appropriate module and enables coordinated grasp-side optimization and planning-side policy learning. Experiments in simulation and on real robots show that GTP-FA improves task success across diverse base learners, including PPO, BC, DP, SAC, and VLA, demonstrating that the framework is not tied to a specific policy-learning paradigm. Although our VLA experiments instantiate the learner with $\pi_{0.5}$, the proposed interface only requires rollout data and downstream fine-tuning samples, and is not specific to this backbone. Ablation studies further show that optimizing only one side or performing naive joint optimization without attribution cannot reliably resolve strongly coupled failures. Future work will extend GTP-FA to more objects, robot platforms, and complex multi-stage tasks, while incorporating finer-grained online feedback to further improve robustness in physical environments.  \begin{ack} This work was partially supported by the National Natural Science Foundation of China (No. 62676094), partially by the Big Data Computing Center of Southeast University, and partially by SEU Kunpeng\&Ascend Center of Cultivation. \end{ack}

{\small
\bibliographystyle{plainnat}
\bibliography{ICML2026}
}

\clearpage

\appendix

\section{Technical appendices and supplementary material}

\subsection{Additional Details for Failure Attribution}
\label{app:fa_details}

\subsubsection{Weak Failure-Attribution Label Construction}
\label{app:weak_fa_labels}

We run $K$ controlled repeated trials under a fixed grasp condition $g$ and record the following grasp-level statistics.

\paragraph{End success rate.}
\begin{equation}
\label{eq:qhat_end_app}
\hat{q}_{\mathrm{end}}(g)
= \frac{1}{K}\sum_{k=1}^{K}\mathbb{I}\{y_{\mathrm{end}}^{(k)}=1\}.
\end{equation}

\paragraph{Success-once rate.}
To capture non-stationary outcomes in which the task succeeds at some intermediate time step but fails at termination, we also compute
\begin{equation}
\label{eq:qhat_once_app}
\hat{q}_{\mathrm{once}}(g)
= \frac{1}{K}\sum_{k=1}^{K}\mathbb{I}\{y_{\mathrm{once}}^{(k)}=1\}.
\end{equation}

\paragraph{Non-stationary fraction.}
\begin{equation}
\label{eq:frac_unstable_app}
\mathrm{frac}_{\mathrm{unstable}}(g)
= \frac{1}{K}\sum_{k=1}^{K}
\mathbb{I}\{y_{\mathrm{once}}^{(k)}=1 \wedge y_{\mathrm{end}}^{(k)}=0\}.
\end{equation}

In addition, we record an execution event summary $\xi$, including slip/drop indicators, grasp-holding duration, terminal-state deviation, and other task-specific execution events. We construct pseudo-labels following the priority rule \emph{``separate G1 first, then distinguish P from G2''}, and only use high-confidence samples for supervised training.

\paragraph{(1) FM-G1: functional/contact-region mismatch.}
If an execution provides clear evidence of a wrong contact region or task-functional mismatch, we directly label the sample as \textbf{FM-G1}. Typical cases include grasps that make a key sub-operation structurally infeasible, grasps in which the gripper blocks a critical workspace, or grasp poses that lack the degrees of freedom required by the downstream task. This rule is applied first so that functional grasp failures are isolated before considering planning-side or stability-related failures.

\paragraph{(2) Distinguishing FM-P and FM-G2 for non-G1 samples.}
For the remaining samples, we use $\hat{q}_{\mathrm{end}}(g)$ together with instability events to distinguish planning-side failures from grasp-instability failures. Given two thresholds $\theta_{\mathrm{low}} \le \theta_{\mathrm{high}}$, we use the following rule:
\begin{itemize}[leftmargin=1.5em]
    \item If $\hat{q}_{\mathrm{end}}(g) \ge \theta_{\mathrm{high}}$, the task is usually realizable under grasp condition $g$. Failures under this grasp are therefore more likely to be caused by planning/policy insufficiency or execution stochasticity, and are labeled as \textbf{FM-P}.
    \item If $\hat{q}_{\mathrm{end}}(g) \le \theta_{\mathrm{low}}$, the task is rarely completed under grasp condition $g$. If we also observe clear instability events, such as dropping or a high non-stationary fraction, the sample is labeled as \textbf{FM-G2}.
    \item If $\hat{q}_{\mathrm{end}}(g) \in (\theta_{\mathrm{low}},\theta_{\mathrm{high}})$, the sample lies in an ambiguous middle range where FM-G2 and FM-P are difficult to separate reliably. We treat such samples as uncertain and either discard them or handle them with soft labels to reduce pseudo-label noise.
\end{itemize}

\paragraph{(3) Successful samples and filtering.}
When $\hat{q}_{\mathrm{end}}(g)$ is extremely high, the corresponding grasp condition can be treated as an approximately successful grasp condition and used for class balancing or contrastive construction. When $\mathrm{frac}_{\mathrm{unstable}}(g)$ is high, it provides an auxiliary stability cue that increases the confidence of FM-G2 supervision. Overall, we retain only high-confidence samples for supervised attribution learning, which reduces pseudo-label noise and mitigates confusion between FM-G2 and FM-P. These labels are not intended as ground-truth causal annotations; rather, they provide high-confidence diagnostic supervision for routing optimization signals between the grasp side and the downstream policy side.

\begin{figure*}[t]
    \centering
    \begin{subfigure}[t]{0.48\linewidth}
        \centering
        \includegraphics[width=\linewidth]{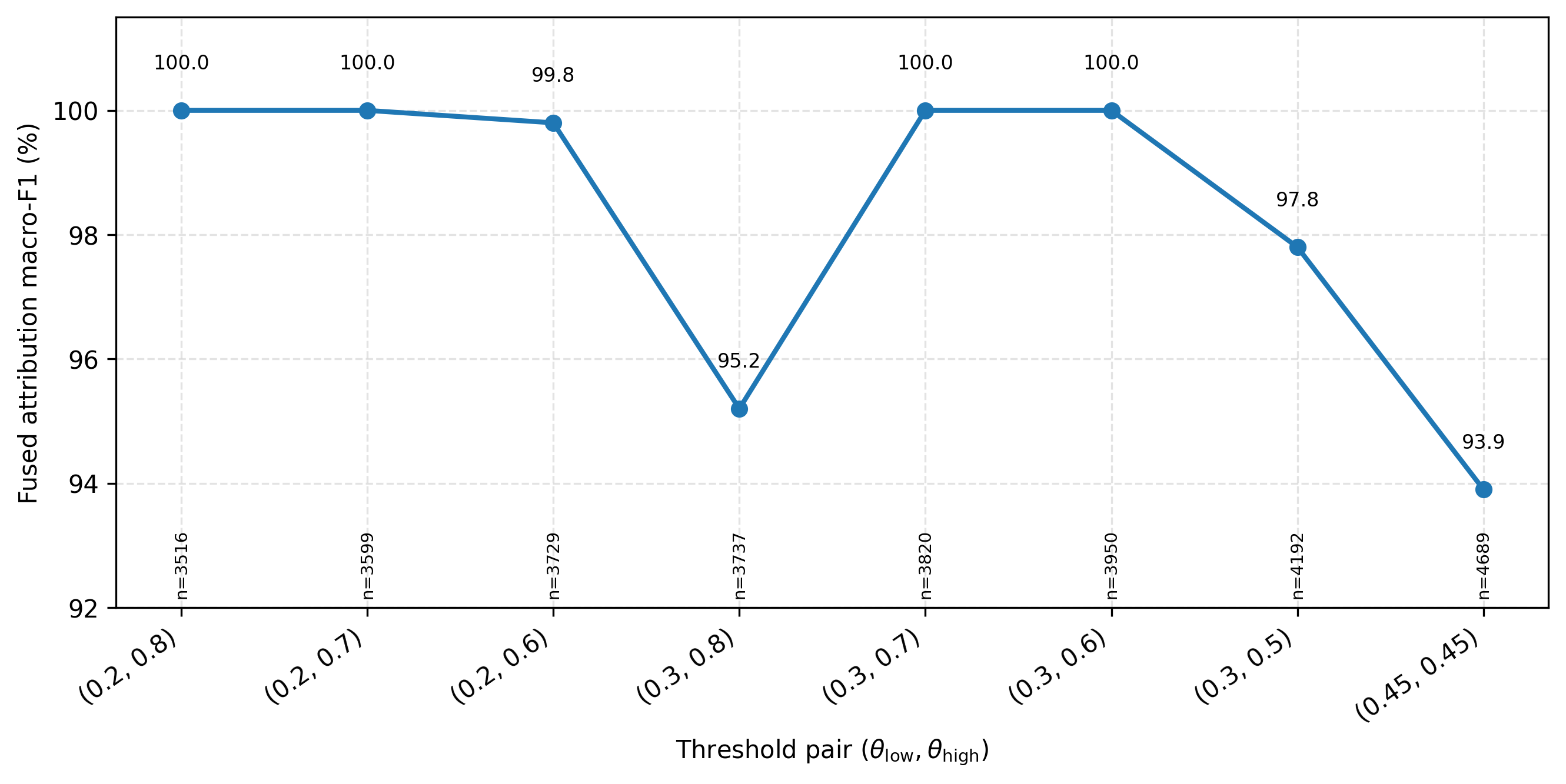}
        \caption{Fused attribution macro-F1 under different weak-label thresholds.}
        \label{fig:threshold_fused_f1}
    \end{subfigure}
    \hfill
    \begin{subfigure}[t]{0.48\linewidth}
        \centering
        \includegraphics[width=\linewidth]{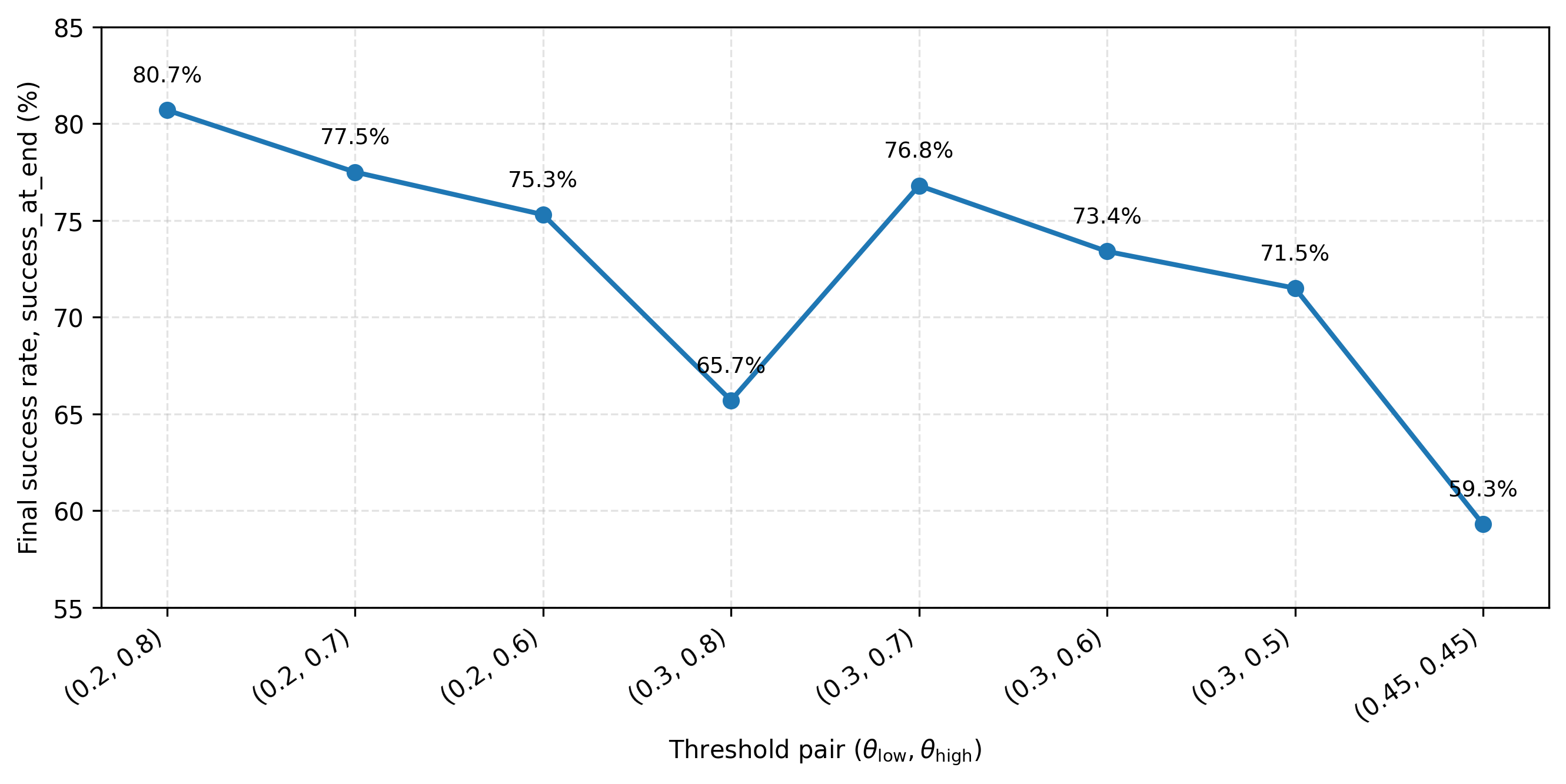}
        \caption{Final GTP-FA-PPO success rate under different weak-label thresholds.}
        \label{fig:threshold_success}
    \end{subfigure}
    \caption{
    Sensitivity analysis of weak failure-attribution thresholds. 
    The left figure reports the macro-F1 of the fused attribution model, which measures how well the fused diagnostic prediction matches the constructed pseudo-labels across failure modes. 
    The right figure reports the final \texttt{success\_at\_end} of the complete GTP-FA-PPO pipeline. 
    The results show that conservative high-confidence thresholds, especially $(\theta_{\mathrm{low}},\theta_{\mathrm{high}})=(0.2,0.8)$, provide a better balance between attribution quality and downstream task performance.
    }
    \label{fig:threshold_sensitivity}
\end{figure*}

\paragraph{Threshold sensitivity analysis.}
The thresholds $\theta_{\mathrm{low}}$ and $\theta_{\mathrm{high}}$ control the trade-off between label coverage and label purity. 
A narrow ambiguous region produces more pseudo-labeled samples but may force uncertain cases into hard labels, whereas a wider high-confidence band discards more ambiguous samples but yields cleaner supervision.
To study this effect, we sweep eight threshold pairs, ranging from conservative high-confidence bands such as $(0.2,0.8)$ to the permissive no-ambiguity setting $(0.45,0.45)$, as shown in Fig.~\ref{fig:threshold_sensitivity}.
For each pair, we re-generate weak labels, train the failure attribution discriminator $D$, train the grasp-conditioned embedding model $E$, evaluate the fused attribution model, and then run the complete GTP-FA-PPO closed-loop pipeline to measure the final task success rate.
We measure attribution quality using macro-F1 over the three failure modes, computed between the fused diagnostic prediction and the constructed high-confidence pseudo-labels. 
For each failure mode $m\in\mathcal{M}$, we compute its F1 score and report the macro-average:
\begin{equation}
\label{eq:macro_f1_app}
\begin{aligned}
\mathrm{F1}_m
&=
\frac{
2\,\mathrm{Precision}_m\,\mathrm{Recall}_m
}{
\mathrm{Precision}_m+\mathrm{Recall}_m
},
\qquad m\in\mathcal{M},\\
\mathrm{Macro\text{-}F1}
&=
\frac{1}{|\mathcal{M}|}
\sum_{m\in\mathcal{M}}
\mathrm{F1}_m .
\end{aligned}
\end{equation}

As shown in Fig.~\ref{fig:threshold_fused_f1}, the fused attribution macro-F1 remains close to saturation for several reasonable threshold pairs, including $(0.2,0.8)$, $(0.2,0.7)$, $(0.3,0.7)$, and $(0.3,0.6)$. This indicates that the weak-label construction rule is robust within a meaningful range of high-confidence thresholds. In contrast, when $\theta_{\mathrm{low}}$ and $\theta_{\mathrm{high}}$ are too close, e.g., $(0.45,0.45)$, the ambiguous region is effectively removed and more uncertain samples are forced into hard pseudo-labels, leading to a drop in attribution quality.

Fig.~\ref{fig:threshold_success} further shows that attribution quality is not the only factor that matters: different thresholds also lead to different closed-loop task success rates. The pair $(0.2,0.8)$ achieves the highest final \texttt{success\_at\_end} of $80.7\%$, while the overly permissive setting $(0.45,0.45)$ drops to $59.3\%$. Figure~\ref{fig:threshold_success} reports threshold sensitivity under a representative GTP-FA-PPO closed-loop setting. Other base algorithms and tasks follow the same weak-label construction, $D/E$ fusion, and hard-P screening procedure with the same default threshold setting; we therefore do not show them individually here. These results suggest that a conservative threshold band improves pseudo-label purity and produces more stable diagnosis-driven optimization. Therefore, we use $(\theta_{\mathrm{low}},\theta_{\mathrm{high}})=(0.2,0.8)$ as the default setting in our main experiments.

\subsubsection{Discriminator Inputs and Training}
\label{app:disc_inputs_train}

The failure attribution discriminator $D$ takes multi-source evidence from a single execution $(\tau,o,g,\xi)$.
Here, $\tau$ denotes the task instruction, $o$ denotes key visual/state observations such as the post-grasp initial state, terminal state, and their deviations, $g$ denotes the selected grasp condition, and $\xi$ denotes execution-summary features such as slip/drop indicators, grasp-holding duration, collision events, and terminal deviation.
These inputs provide complementary evidence for distinguishing functional grasp mismatch, grasp instability, and planning/policy insufficiency.

We implement $D$ with a lightweight encode--fuse--classify architecture.
Task, observation, grasp, and execution-summary features are first encoded by separate small encoders and then concatenated into a fused representation.
A classification head outputs a categorical distribution over the three failure modes:
\begin{equation}
\label{eq:d_output_app}
p_D(m\mid \tau,o,g,\xi), \qquad
m\in\mathcal{M}=\{\mathrm{FM\text{-}G1},\mathrm{FM\text{-}G2},\mathrm{FM\text{-}P}\}.
\end{equation}

The discriminator is trained on the high-confidence pseudo-labeled dataset constructed in Appendix~\ref{app:weak_fa_labels}.
Given a pseudo-label $\tilde{m}$, we minimize the cross-entropy loss:
\begin{equation}
\label{eq:ld_app}
\mathcal{L}_{D}
=
-\mathbb{E}_{(\tau,o,g,\xi,\tilde{m})}
\big[
\log p_D(\tilde{m}\mid \tau,o,g,\xi)
\big].
\end{equation}

After training, $D$ provides both a failure-mode prediction and a confidence score.
The predicted distribution is used as the failure explanation and diagnostic routing signal in the main text.
When combined with the grasp-conditioned neighborhood prior from $E$, the confidence of $D$ also determines how much the final fused attribution should rely on the discriminator prediction versus the embedding-based prior.

\subsubsection{Grasp-Conditioned Diagnostic Embedding and $D/E$ Fusion}
\label{app:de_fusion}

To improve failure attribution generalization to unseen grasps, we introduce a Grasp-Conditioned Diagnostic Representation Space $E$ in addition to the failure attribution discriminator $D$. The key assumption is that grasp conditions inducing similar post-grasp initialization states should exhibit similar diagnostic outcomes. Therefore, $E$ does not directly predict actions; instead, it learns a representation space for nearest-neighbor retrieval and diagnostic prior estimation.

\paragraph{Grasp-conditioned embedding.}
For each grasp condition $g$, we extract a grasp-conditioned initialization feature $x_g$ from the environment or rollout records. This feature describes the post-grasp state induced by the selected grasp, such as object pose, relative gripper pose, contact region, or other observable state features. We learn an embedding function
\begin{equation}
e = f_{\theta}(x_g) \in \mathbb{R}^{d}.
\end{equation}
Given a rollout dataset
\begin{equation}
\mathcal{D}_{E}
=
\{(x_{g,i},y_i,m_i)\}_{i=1}^{N},
\end{equation}
where $y_i\in\{0,1\}$ denotes the success or failure outcome and $m_i\in\mathcal{M}$ denotes the high-confidence failure-mode label constructed by the weak-labeling rule, successful samples can also be used as auxiliary supervision to improve the separability of success and failure structures in the embedding space.

\paragraph{Local diagnostic-consistency objective.}
We encourage grasp conditions with similar diagnostic outcomes to be close in the embedding space. For each anchor sample $i$, we construct a positive set $\mathcal{P}(i)$ consisting of samples that share the same failure mode or the same success/failure attribute. Let $e_i=f_\theta(x_{g,i})$. We use a supervised contrastive objective:
\begin{equation}
\mathcal{L}_{\mathrm{con}}
=
-\sum_i
\frac{1}{|\mathcal{P}(i)|}
\sum_{p\in\mathcal{P}(i)}
\log
\frac{
\exp(\mathrm{sim}(e_i,e_p)/t)
}{
\sum_{a\neq i}\exp(\mathrm{sim}(e_i,e_a)/t)
},
\end{equation}
where $\mathrm{sim}(\cdot,\cdot)$ is a similarity function and $t$ is a temperature parameter. This objective encourages grasp conditions with similar diagnostic attributes to form local clusters in the representation space.

We further attach lightweight prediction heads to the embedding representation to predict the success/failure label and the failure-mode label:
\begin{equation}
\mathcal{L}_{\mathrm{aux}}
=
\mathrm{CE}(\hat{y}_i,y_i)
+
\mathrm{CE}(\hat{m}_i,m_i).
\end{equation}
The final training objective for $E$ is
\begin{equation}
\min_{\theta}
\mathcal{L}_{E}
=
\mathcal{L}_{\mathrm{con}}
+
\lambda \mathcal{L}_{\mathrm{aux}},
\end{equation}
where $\lambda$ controls the weight of the auxiliary classification losses. This design enables $E$ to capture both local diagnostic consistency and basic discriminative structure.

\paragraph{Neighborhood diagnostic prior.}
After training, we maintain an embedding bank constructed from training or rollout samples:
\begin{equation}
\mathcal{B}
=
\{(e_j,m_j)\}_{j=1}^{M}.
\end{equation}
For an unseen grasp, its grasp-conditioned feature is mapped to an embedding $e$. We retrieve its $k$ nearest neighbors $\mathcal{N}_k(e)$ from the embedding bank and construct a diagnostic prior by weighted voting over their failure-mode labels:
\begin{equation}
\tilde{p}_{E}(m\mid e)
=
\sum_{j\in\mathcal{N}_k(e)}
w_j \mathbb{I}\{m_j=m\},
\end{equation}
where the neighbor weight is defined as
\begin{equation}
w_j
=
\frac{
\exp(\mathrm{sim}(e,e_j)/t)
}{
\sum_{\ell\in\mathcal{N}_k(e)}
\exp(\mathrm{sim}(e,e_{\ell})/t)
}.
\end{equation}
This prior can be interpreted as a local vote over diagnostically similar grasps. When $D$ becomes unreliable on unseen grasps or distribution-shifted samples, $\tilde{p}_{E}(m\mid e)$ provides a smoother neighborhood constraint.

\paragraph{Fusion with discriminator $D$.}
Finally, we fuse the single-execution diagnostic prediction from $D$ with the neighborhood diagnostic prior from $E$:

\begin{equation}
p_{\mathrm{fuse}}(m\mid\tau,o,g,\xi,e)
=
(1-\alpha)p_D(m\mid\tau,o,g,\xi)
+
\alpha \tilde{p}_E(m\mid e).
\end{equation}

For simplicity, we write it as $p_{\mathrm{fuse}}(m)$ when the conditioning variables are clear. Here, $\alpha\in[0,1]$ is determined by the confidence of $D$. If $D$ is confident on the current sample, we use a smaller $\alpha$ so that the fused result relies more on the discriminator prediction. If $D$ is uncertain, we use a larger $\alpha$ so that the fused result relies more on the neighborhood diagnostic prior.

This $D/E$ fusion mechanism reduces attribution noise on unseen grasps and stabilizes failure-mode prediction under mild grasp distribution shifts. The fused attribution $p_{\mathrm{fuse}}$ is then used for responsibility assignment, hard-P sample screening, grasp-side risk suppression, and planning-side policy updates.

\subsection{Formulaic Supplements for Diagnosis-driven Bidirectional Optimization}
\label{app:diag_bidir_formulas}

\subsubsection{Definition of the hard-P subset}
\label{app:hardP_def}

Given a candidate grasp condition $g$, we perform conditional aggregation of the fused diagnosis outputs over the subset of rollouts that
use $g$, obtaining expectation estimates of planning-side and grasp-side risks:
\begin{equation}
\label{eq:rbar_defs}
\begin{aligned}
\bar{p}_{P}(g)
&=
\mathbb{E}\!\left[p_{\mathrm{fuse}}(\mathrm{FM\text{-}P}\mid\cdot)\mid g\right],\\
\bar{p}_{G1}(g)
&=
\mathbb{E}\!\left[p_{\mathrm{fuse}}(\mathrm{FM\text{-}G1}\mid\cdot)\mid g\right],\\
\bar{p}_{G2}(g)
&=
\mathbb{E}\!\left[p_{\mathrm{fuse}}(\mathrm{FM\text{-}G2}\mid\cdot)\mid g\right].
\end{aligned}
\end{equation}

We then define the planning-dominant hard-P subset as
\begin{equation}
\label{eq:hardp_set_def}
\mathcal{G}_{\mathrm{hardP}}
=
\left\{
g\in\mathcal{G}_{\mathrm{full}}
\;\middle|\;
\bar{p}_{P}(g)\ge \delta_P,\;
\bar{p}_{G1}(g)\le \delta_{G1},\;
\bar{p}_{G2}(g)\le \delta_{G2},\;
n(g)\ge n_{\min}
\right\},
\end{equation}
where $n(g)$ denotes the number of diagnostic samples collected for grasp condition $g$.
$(\delta_{P},\delta_{G1},\delta_{G2})$ control the confidence-based screening for ``planning-dominant'' starts:
$\delta_{P}$ enforces sufficiently high planning-failure risk, while $\delta_{G1},\delta_{G2}$ suppress grasp-side (G1/G2)
contamination so that the subset better matches start conditions that are relatively feasible in grasping but challenging for planning.

In our implementation, the hard-P screening procedure is shared across different base algorithms. 
The system first uses the current downstream policy to generate rollouts under candidate grasp conditions, and then estimates the planning-side and grasp-side risks of each grasp condition using the fused $D/E$ diagnostic outputs. 
To avoid unstable estimates caused by too few rollouts, we require each candidate grasp condition to have a minimum number of diagnostic samples, and select at most a fixed number of hard-P starts from the candidates that satisfy the screening criteria. 
In the main experiments, we set the minimum aggregation count to 5 and select at most 200 hard-P starts. 
The screening thresholds are set to $\delta_P=0.65$, $\delta_{G2}=0.25$, and $\delta_{G1}=0.20$. 
This requires a candidate start to have sufficiently high planning-failure risk while keeping the risks of grasp-functional mismatch and grasp instability low. 
As a result, $\mathcal{G}_{\mathrm{hardP}}$ better approximates a set of starts where the grasp side is relatively feasible but downstream planning or policy execution remains challenging.

Note that the screening thresholds $(\delta_P,\delta_{G1},\delta_{G2})$ are different from the weak-label thresholds $(\theta_{\mathrm{low}},\theta_{\mathrm{high}})$ in Appendix~\ref{app:weak_fa_labels}. 
The latter are applied to the repeated-trial success rate $\hat{q}_{\mathrm{end}}(g)$ under a fixed grasp condition and are used to construct high-confidence pseudo-labels, whereas the former are applied to the averaged failure-mode probabilities after $D/E$ fusion and are used to select planning-dominant hard-P starts. 
The two sets of thresholds are sequentially related in the pipeline but are not the same hyperparameters.

\subsubsection{Distribution reshaping as an importance-weighting view}
\label{app:reshape_importance}

Let $\mu_{\mathrm{full}}(g)$ denote the baseline sampling distribution over the full pool $\mathcal{G}_{\mathrm{full}}$, and
$\mu_{\mathrm{hardP}}(g)$ denote the sampling distribution over $\mathcal{G}_{\mathrm{hardP}}$. In training, we adopt the mixed
reset distribution
\begin{equation}
\label{eq:mu_rho}
\mu_{\rho}(g) = (1-\rho)\,\mu_{\mathrm{full}}(g) + \rho\,\mu_{\mathrm{hardP}}(g),
\end{equation}
where $\rho\in[0,1]$ controls the focus strength on planning-bottleneck regions. This reshaping can be interpreted as explicit
importance reweighting over the baseline distribution: for any expectation term $f(g)$,
\begin{equation}
\label{eq:importance_view}
\mathbb{E}_{g\sim \mu_{\rho}}[f(g)]
=
\mathbb{E}_{g\sim \mu_{\mathrm{full}}}
\!\left[
\frac{\mu_{\rho}(g)}{\mu_{\mathrm{full}}(g)}\, f(g)
\right],
\end{equation}
where
\begin{equation}
\label{eq:omega_rho}
\omega_{\rho}(g)=\frac{\mu_{\rho}(g)}{\mu_{\mathrm{full}}(g)}
\end{equation}
acts as an explicit upsampling weight for hard planning starts. This equivalence highlights that ``focusing on
$\mathcal{G}_{\mathrm{hardP}}$ at the reset-distribution level'' matches ``importance weighting at the sample level'' in objective effect.

During closed-loop training, we do not inject hard-P samples in the initial iteration, i.e., $\rho=0$. 
Starting from later iterations, the system uses the hard-P subset mined from the previous iteration to reshape the training distribution. 
For interactive reinforcement learning methods such as PPO and SAC, this is implemented as start-state resampling during training, where planning-bottleneck starts are sampled from $\mathcal{G}_{\mathrm{hardP}}$ with a certain probability. 
In the main experiments, this ratio is set to $\rho=0.2$. 
For data-driven methods such as BC, DP, and $\pi_{0.5}$, this process does not rely on online reset interaction. 
Instead, it corresponds to restructuring and resampling the training data, rollout data, or fine-tuning data induced by grasp conditions, so that the model is exposed more often to samples that are genuinely difficult for the planning side but relatively feasible for the grasp side.

During standard evaluation, we always set $\rho=0$, i.e., no hard-P starts are injected, so that different methods are compared under the same evaluation distribution. 
Therefore, distribution reshaping only affects sample coverage during training or data construction, while leaving the final evaluation protocol unchanged.

\subsubsection{Grasp-side Risk Calibration and Task-prior Scoring}
\label{app:grasp_side_scoring}

Sec.~\ref{sec:bidirectional_optimization} gives the unified scoring form for grasp-side updates. Here we further explain the concrete meaning and implementation of the task-prior score $s_{\mathrm{prior}}$, the risk predictor $r_\phi$, and the final grasp score.

First, the base grasping model assigns an original score $s_{\mathrm{base}}(g;o)$ to each candidate grasp $g\in\mathcal{G}(o)$. This score mainly reflects geometric feasibility, grasp stability, or the internal quality estimate of the base grasping model. However, in task-oriented manipulation, relying only on $s_{\mathrm{base}}$ may select grasps that are geometrically stable but incompatible with the downstream task. We therefore introduce a task-prior score $s_{\mathrm{prior}}(g;\pi_{\mathrm{VLM}}(\tau))$.

Here, $\pi_{\mathrm{VLM}}(\tau)$ denotes a structured task prior obtained from a VLM using the task instruction and visualized candidate grasps. This prior does not directly predict a continuous grasp pose. Instead, it specifies preferred grasp regions, forbidden grasp regions, buffer regions, and geometric constraint rules. These rules are then converted into candidate-grasp scores or filtering masks. For example, if a candidate grasp lies in a functional contact region, a release-blocking region, or an obviously unstable region, its prior score is reduced; if it lies in a task-preferred region, its prior score is increased. The concrete VLM prompts and task-specific filtering examples are provided in Appendix~\ref{app:vlm_task_prior}.

Second, we train a lightweight risk predictor $r_\phi(z(g))$ using the fused failure-attribution signal. Here, $z(g)$ can be the candidate-grasp feature from the base grasp scorer, the grasp-pose encoding, or an intermediate representation from the grasp-scoring network. This predictor estimates the risk that the current candidate grasp will induce an FM-G2 failure, i.e., grasp instability, slip, or drop:
\begin{equation}
r_\phi(z(g))
\approx
p_{\mathrm{fuse}}(\mathrm{FM\text{-}G2}\mid \tau,o,g,\xi, e).
\end{equation}
In implementation, $r_\phi$ can be used as a calibration head on top of the base grasp scorer, or as a lightweight fine-tuning signal for the grasping model. The final GTP-FA grasp-side score is written as
\begin{equation}
S_{\mathrm{GTP\text{-}FA}}(g)
=
s_{\mathrm{base}}(g;o)
+
\lambda s_{\mathrm{prior}}(g;\pi_{\mathrm{VLM}}(\tau))
-
\beta r_\phi(z(g)).
\end{equation}
Here, $\lambda$ controls the strength of the task prior, and $\beta$ controls the strength of instability-risk suppression. The final grasp is selected by
\begin{equation}
g^*
=
\arg\max_{g\in\mathcal{G}(o)}
S_{\mathrm{GTP\text{-}FA}}(g).
\end{equation}

This design injects both task-semantic constraints and diagnostic risk signals into grasp-side optimization. The task prior mainly reduces FM-G1 failures caused by functional mismatch, while risk calibration mainly suppresses FM-G2 failures caused by unstable grasps. Together, they encourage the grasp module to output candidates that are not only geometrically feasible, but also better suited for downstream planning and policy execution.

\subsection{Coarse-grained: VLM-guided Task-prior Constraints}
\label{app:vlm_task_prior}

\subsubsection{Task-prior prompting and candidate grasp filtering}

To incorporate high-level task semantics into low-level grasp selection while avoiding direct continuous pose prediction by VLMs/LLMs, we formulate task-prior construction as a candidate grasp filtering problem.
The system first uses GraspNet to generate a set of geometrically feasible grasp candidates, and then provides the task instruction, scene observation, and visualized grasp candidates to a VLM/LLM.
The goal of the VLM is not to predict the final grasp pose, but to identify which candidate grasps are more consistent with the downstream task and which candidates should be avoided because they occupy functional regions, destabilize the object, or make subsequent placement/release difficult.

We use a structured prompt to elicit coarse task-level grasp constraints.
The prompt asks the model to identify: (i) which object or object part should interact with the environment during task execution; (ii) which region should be preferred for grasping; (iii) which region should be avoided because grasping it may block the functional part, destabilize manipulation, or hinder later placement/release; and (iv) when no explicit functional part exists, which geometric constraints favor stable execution, such as avoiding bottom, corner, or release-unfriendly grasps.
The model returns a structured response including \texttt{preferred\_grasp\_region}, \texttt{forbidden\_grasp\_region}, \texttt{optional\_buffer\_region}, \texttt{geometric\_rule}, and \texttt{explanation}, which is then converted into a pruning rule over the candidate grasp set.

This design uses VLM reasoning only as a coarse task-level prior, while leaving low-level grasp geometry to GraspNet, TaskScore, and subsequent optimization.
It therefore exploits the semantic reasoning capability of VLMs/LLMs without relying on them for high-precision continuous action prediction.

\begin{tcolorbox}[
  enhanced,
  breakable,
  colback=lightgray!25,
  colframe=black!45,
  boxrule=0.7pt,
  arc=1mm,
  left=5pt,
  right=5pt,
  top=5pt,
  bottom=5pt,
  title={General VLM Prompt for Task-prior Grasp Filtering}
]
\small
\textbf{Prompt:}

You are given a robot manipulation scene and a task instruction. 
Your goal is not to predict an exact grasp pose, but to identify task-prior grasping constraints.

Please analyze:
\begin{enumerate}[leftmargin=1.5em, itemsep=0.15em]
    \item Which object or object part should interact with the environment during task execution?
    \item Which region should be grasped by the robot gripper?
    \item Which region should be avoided because grasping it would block the functional part, destabilize the object, or make the subsequent placement/release difficult?
    \item If the task does not have an explicit functional part, infer geometric constraints that favor stable execution, such as avoiding bottom, corner, or release-unfriendly grasps.
\end{enumerate}

Return the result as:
\begin{lstlisting}[style=pddlStyle]
preferred_grasp_region:
forbidden_grasp_region:
optional_buffer_region:
geometric_rule:
explanation:
\end{lstlisting}
\end{tcolorbox}

\subsubsection{Task-specific prior rules}

For tool-use tasks with explicit functional parts, the VLM prior mainly prevents the gripper from occupying the region that must interact with the environment.
In \textsc{PullCubeTool}, the hook tip and hook-root region are functional parts used to engage and pull the cube.
Therefore, the prior keeps grasps on the elongated straight handle, especially the middle-to-rear handle region, while removing grasps on the hook tip, the curved hook, and the hook-root transition, which may block the hook from engaging the cube.

In \textsc{PokeCube}, the front end of the peg serves as the functional contact point for pushing the cube.
Grasping near this end may block or weaken the poking interaction.
The prior therefore preserves grasps on the rear half of the peg and removes candidates near the front functional end.

In \textsc{LiftPegUpright}, the peg can in principle be grasped at multiple locations, but grasps close to an end tend to induce off-center torque and unstable rotation during lifting.
We therefore prune grasps near the red end and retain grasps around the middle region or the opposite colored side, which are more favorable for stable lifting and posture adjustment.

For tasks without explicit functional parts, the VLM prior mainly encodes execution-feasibility constraints.
In \textsc{PlaceSphere}, the sphere has no semantic functional end, so the prior favors top-down or upper-side grasps that support stable lifting, placement, and release.
Candidates approaching from below, near the table-contact region, or from release-unfriendly directions are removed.

In \textsc{StackCube}, the target cube can be grasped from several valid directions.
We therefore apply only a weak prior: clearly unfavorable candidates, such as bottom grasps, corner grasps, or grasps whose centers lie outside the main body of the red cube, are pruned, while most stable side and top-down grasps are retained.
For \textsc{PickCube}, \textsc{PushCube}, and \textsc{PullCube}, which are also single-cube manipulation tasks, we use similar weak filtering rules and omit separate visualizations for space.

\subsubsection{Visualization of task-prior grasp filtering}

\begin{figure*}[ht]
    \centering
    \setlength{\tabcolsep}{2pt}
    \renewcommand{\arraystretch}{0.75}
    \small
    \begin{tabular}{@{}lccc@{}}
        \toprule
        \textbf{Task} & \textbf{Raw candidates} & \textbf{Kept candidates} & \textbf{Dropped candidates} \\
        \midrule

        \textsc{StackCube}
        &
        \includegraphics[width=0.215\linewidth]{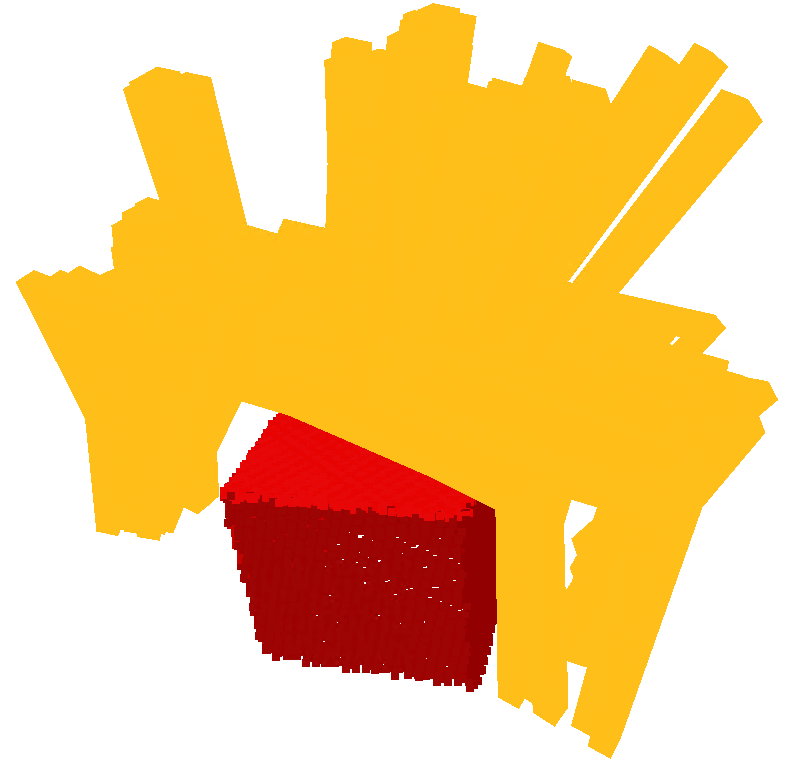}
        &
        \includegraphics[width=0.215\linewidth]{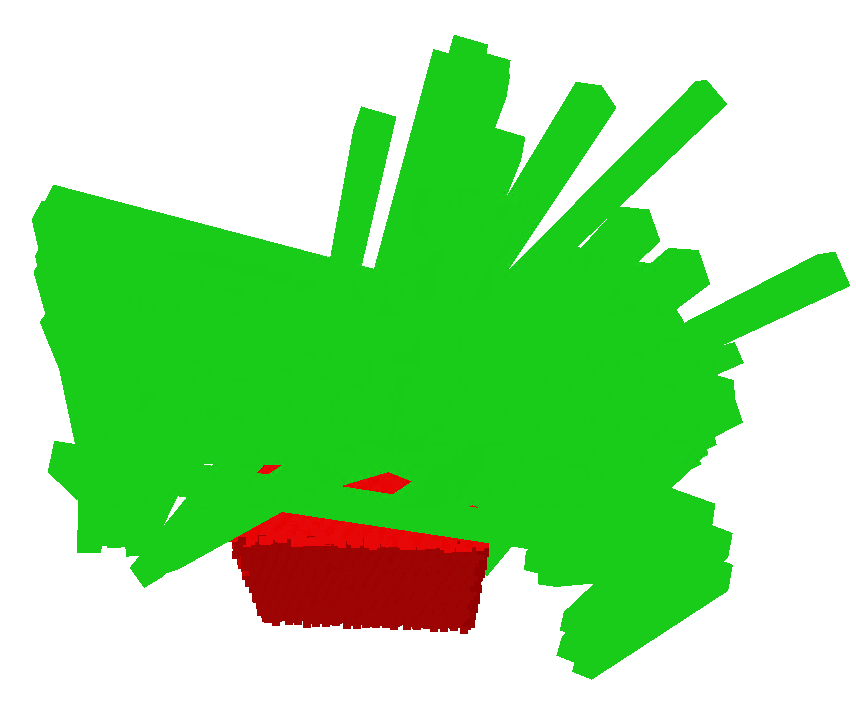}
        &
        \includegraphics[width=0.215\linewidth]{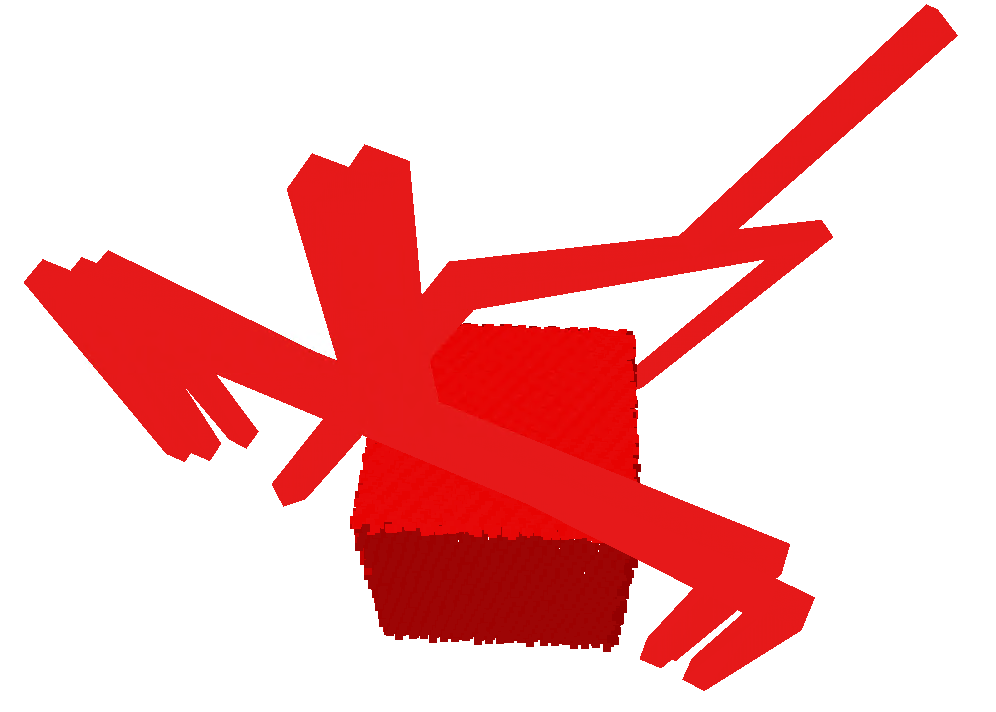}
        \\[-0.2em]

        \textsc{PlaceSphere}
        &
        \includegraphics[width=0.215\linewidth]{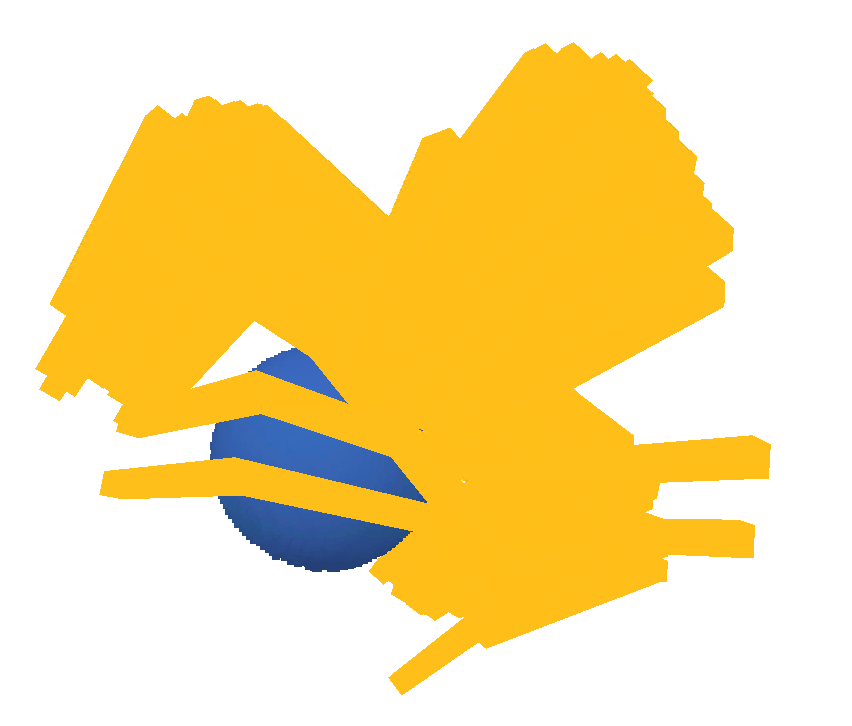}
        &
        \includegraphics[width=0.215\linewidth]{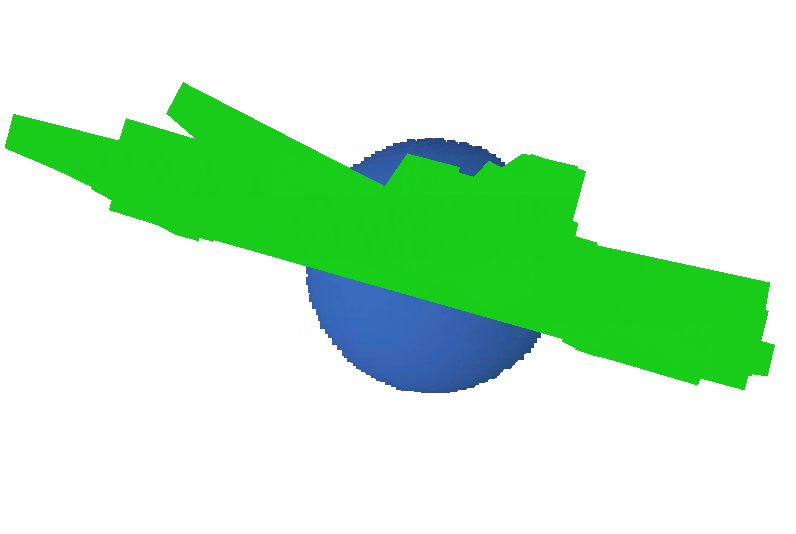}
        &
        \includegraphics[width=0.215\linewidth]{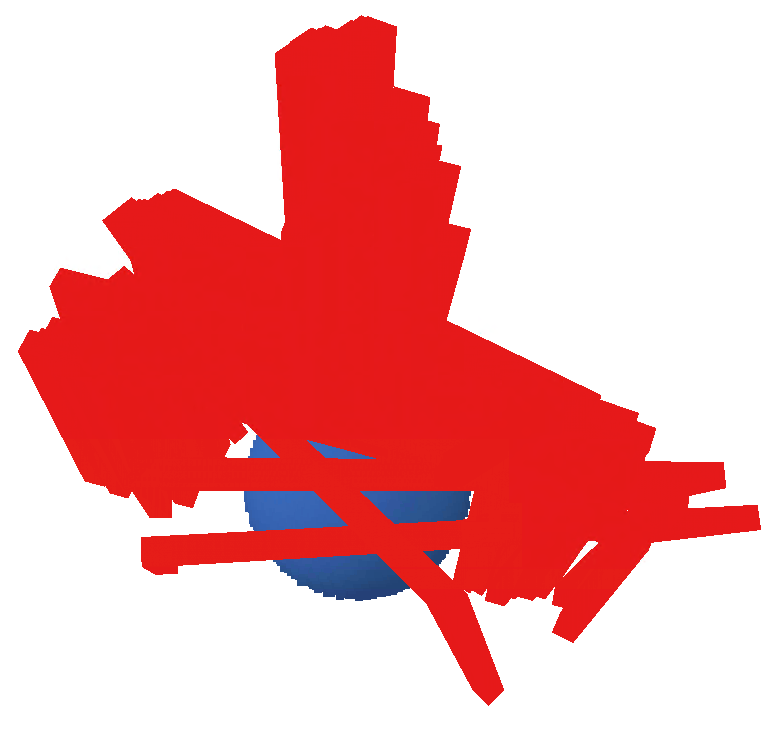}
        \\[-0.2em]

        \textsc{LiftPegUpright}
        &
        \includegraphics[width=0.215\linewidth]{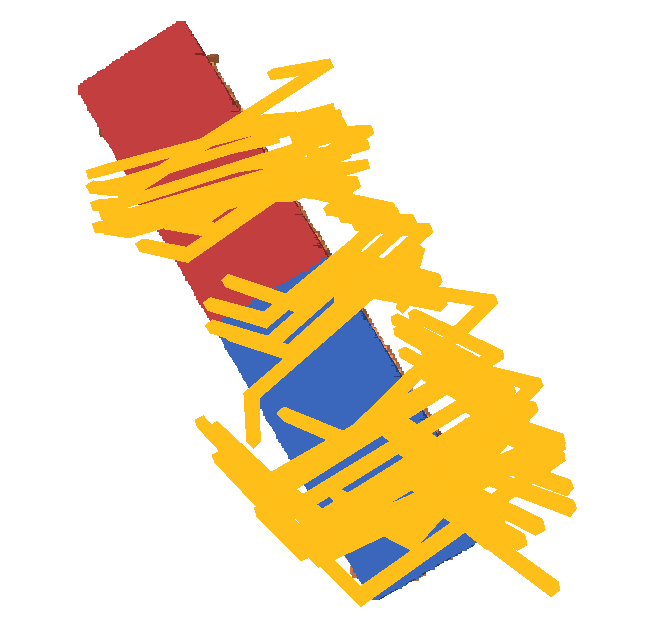}
        &
        \includegraphics[width=0.215\linewidth]{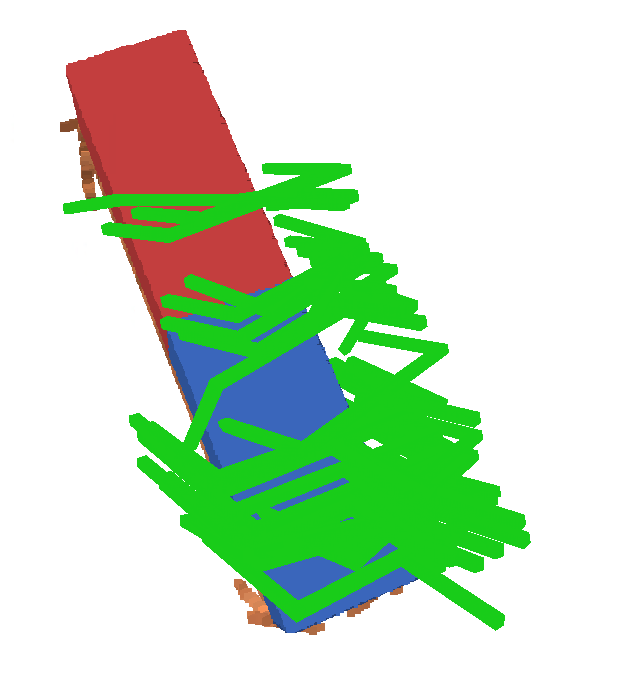}
        &
        \includegraphics[width=0.215\linewidth]{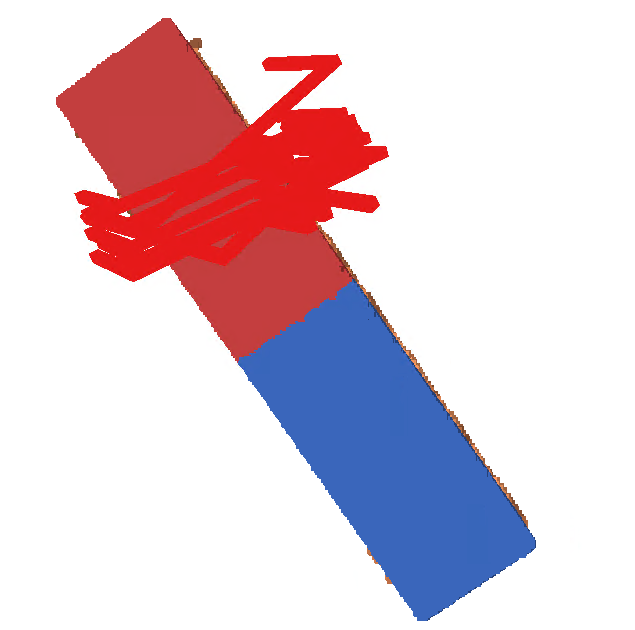}
        \\[-0.2em]

        \textsc{PokeCube}
        &
        \includegraphics[width=0.215\linewidth]{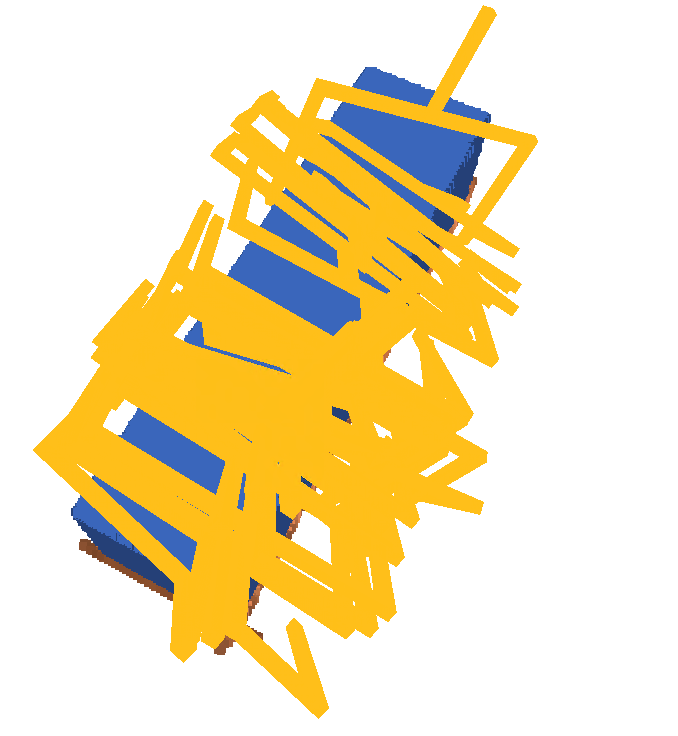}
        &
        \includegraphics[width=0.215\linewidth]{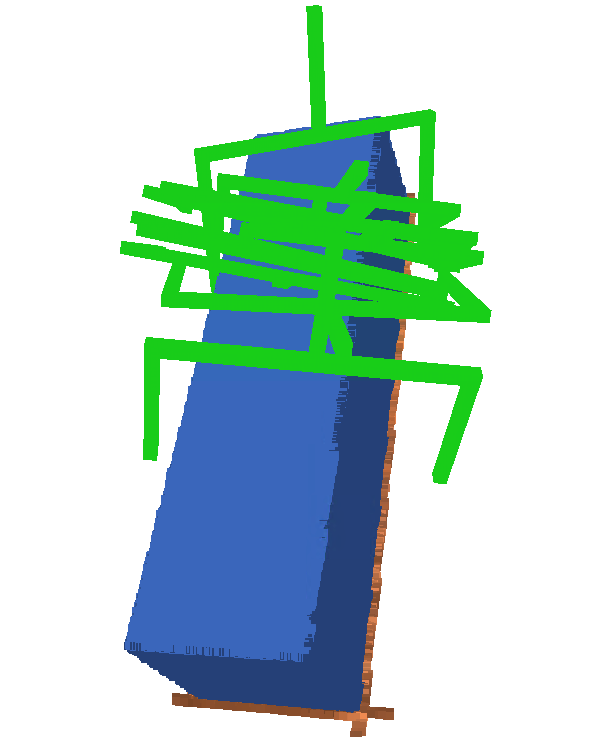}
        &
        \includegraphics[width=0.215\linewidth]{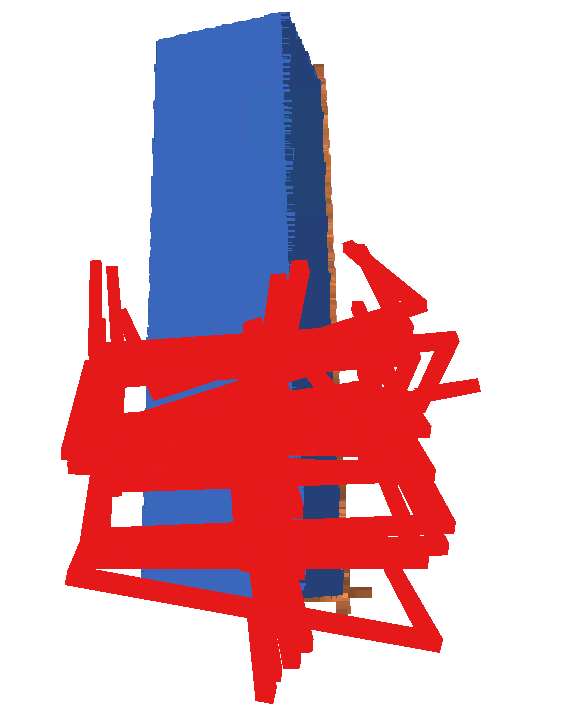}
        \\[-0.2em]

        \textsc{PullCubeTool}
        &
        \includegraphics[width=0.215\linewidth]{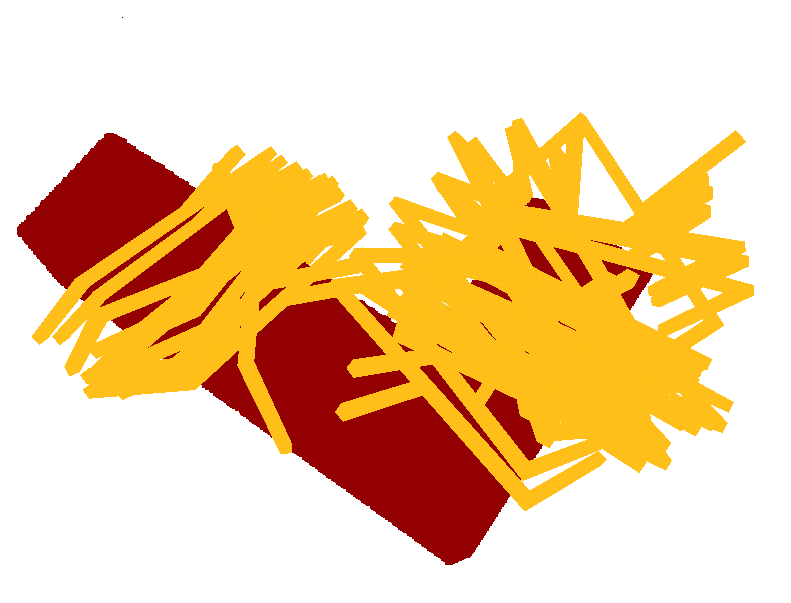}
        &
        \includegraphics[width=0.215\linewidth]{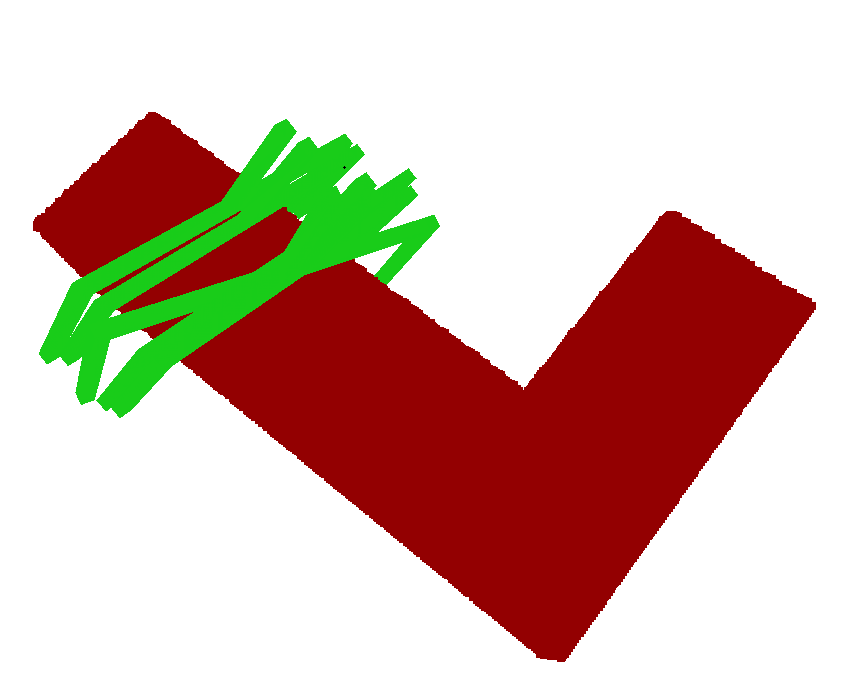}
        &
        \includegraphics[width=0.215\linewidth]{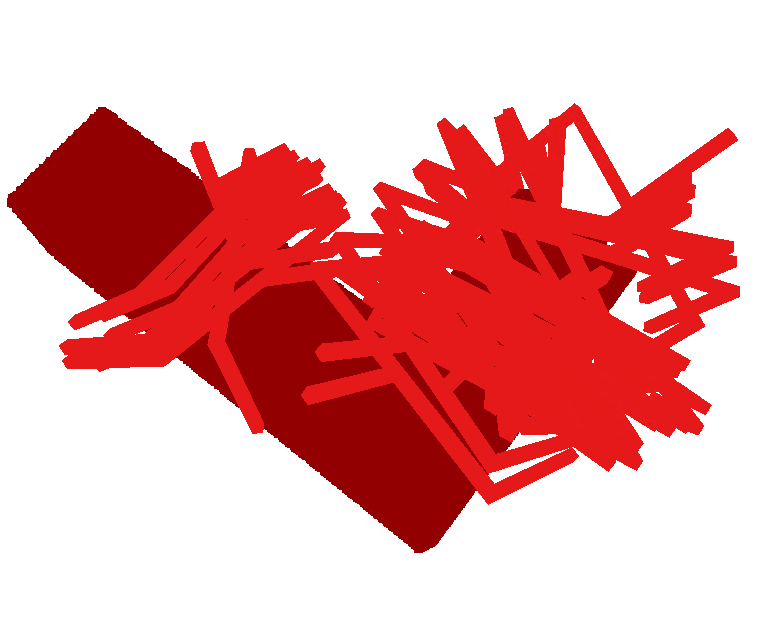}
        \\

        \bottomrule
    \end{tabular}
    \caption{
    Visualization of VLM/LLM-guided task-prior grasp filtering.
    Each row shows one task, and the three columns show the raw GraspNet candidates, the retained task-consistent candidates, and the pruned candidates, respectively.
    Yellow grasps denote the original candidates, green grasps denote retained candidates, and red grasps denote dropped candidates.
    The task prior removes grasps that occupy functional regions, induce unstable manipulation, or make subsequent placement/release difficult, while preserving candidates that are more consistent with downstream task execution.
    }
    \label{fig:vlm_grasp_filtering}
\end{figure*}

Figure~\ref{fig:vlm_grasp_filtering} visualizes the effect of VLM/LLM-guided task-prior filtering.
For each task, we show the raw GraspNet candidates, the task-consistent candidates retained after filtering, and the pruned candidates.
The VLM prior does not merely reduce the number of candidates; instead, it removes grasps that may be geometrically feasible but inconsistent with the downstream task.
For example, in \textsc{PullCubeTool} and \textsc{PokeCube}, many removed grasps occupy the functional end of the tool; in \textsc{LiftPegUpright}, removed grasps concentrate near the unstable end region; in \textsc{PlaceSphere}, the prior removes candidates that are unfavorable for lifting or release; and in \textsc{StackCube}, the prior is conservative and mainly removes bottom, corner, or off-object grasps.

This coarse task prior provides a more appropriate search space for the subsequent grasp-side optimization.
It prunes task-inconsistent candidates before fine-grained scoring and continuous refinement, reducing functional grasp failures of type FM-G1 and lowering the chance that downstream policy learning is contaminated by poor grasp-induced initial conditions.

\subsubsection{Real-world SoM-guided Task-prior Construction}
\label{app:som_real_grasp_selection}

To further illustrate how task priors are constructed in real-robot settings, we visualize the VLM-guided task-prior grasp selection process on three real-robot tasks: \textsc{PokeCube}, \textsc{PullCubeTool}, and \textsc{PourWater}.
Consistent with the formulation in the main text and Appendix~A.3, the VLM does not directly output a continuous 6-DoF grasp pose.
Instead, given the task instruction and visual observation, it identifies task-relevant object regions, functional parts, preferred grasp regions, and regions that should be avoided.
These coarse semantic constraints are then converted into grasp-candidate filtering or scoring rules, and are combined with GraspNet candidates, task-prior scoring, and diagnostic risk calibration to select the final execution grasp.

Specifically, we use Set-of-Mark (SoM) prompting as a visual grounding interface for real-world images~\citep{yang2023set}.
This step explicitly visualizes candidate regions, target objects, and functional parts with readable indices, masks, or bounding marks, enabling the VLM to more reliably associate language instructions with image regions.
SoM marks guide the VLM in defining preferred, forbidden, and buffer regions and their geometric constraints for grasp-candidate filtering.
These priors are then converted into filtering rules or scoring terms for grasp candidates, corresponding to the task-prior score $s_{\mathrm{prior}}(g;\pi_{\mathrm{VLM}}(\tau))$ in the grasp-side objective.

\begin{figure*}[t]
    \centering
    \setlength{\tabcolsep}{2pt}

    \begin{subfigure}[t]{0.32\linewidth}
        \centering
        \includegraphics[width=\linewidth]{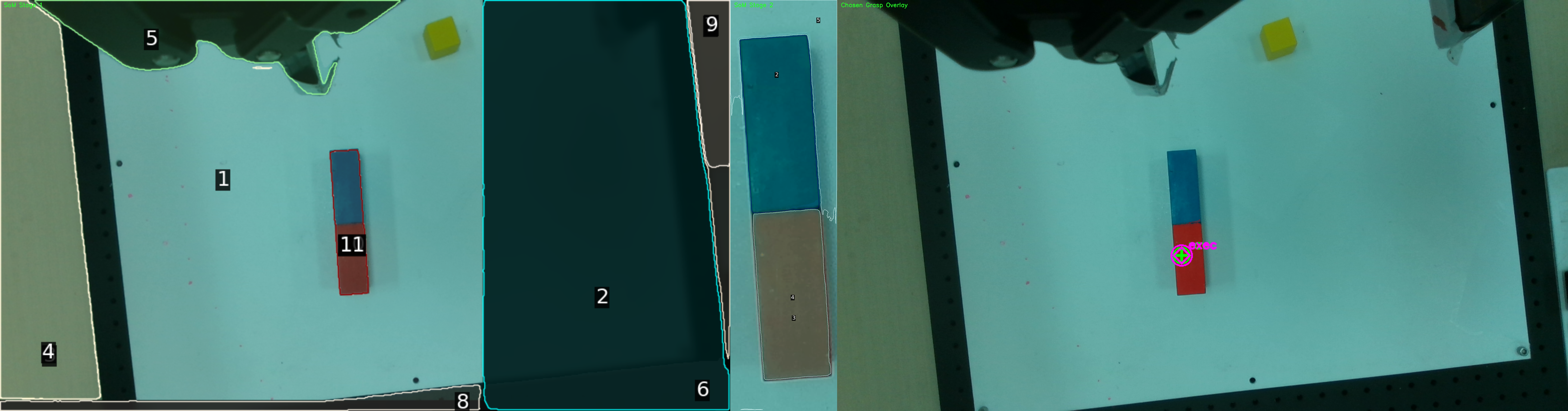}
        \caption{\textsc{PokeCube}: grasp the red end of the stick for pushing.}
        \label{fig:som_pokecube}
    \end{subfigure}
    \hfill
    \begin{subfigure}[t]{0.32\linewidth}
        \centering
        \includegraphics[width=\linewidth]{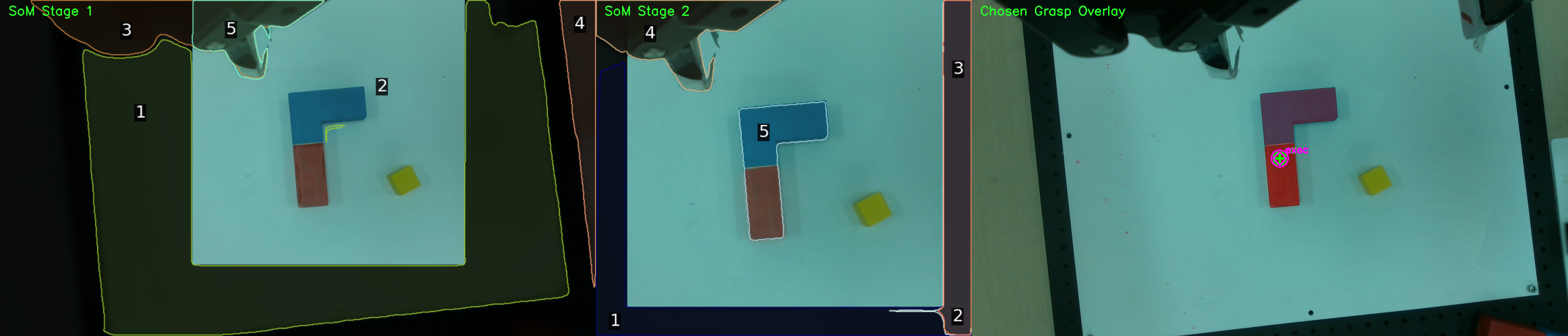}
        \caption{\textsc{PullCubeTool}: grasp the red part of the hook for pulling.}
        \label{fig:som_pullcubetool}
    \end{subfigure}
    \hfill
    \begin{subfigure}[t]{0.32\linewidth}
        \centering
        \includegraphics[width=\linewidth]{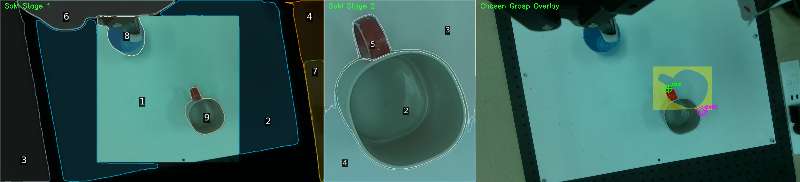}
        \caption{\textsc{PourWater}: grasp the red handle of the gray cup for pouring.}
        \label{fig:som_pourwater}
    \end{subfigure}

    \caption{
    Real-world SoM-guided task-prior grasp selection.
    For each task, we show SoM Stage 1 region marking, SoM Stage 2 target/function-part selection, and the final chosen grasp overlay.
    SoM marks help the VLM ground task semantics to concrete image regions, thereby generating task-compatible grasp priors for \textsc{PokeCube}, \textsc{PullCubeTool}, and \textsc{PourWater}.
    }
    \label{fig:som_real_task_prior}
\end{figure*}

Figure~\ref{fig:som_real_task_prior} shows the SoM-guided task-prior grasp selection process for the three real-robot tasks.
Each task proceeds through three stages.
SoM Stage 1 marks candidate regions in the real image, helping the model identify objects, tools, and target areas in the scene.
SoM Stage 2 further selects the target object or functional part according to the task instruction.
The Chosen Grasp Overlay then visualizes the final task-compatible grasp location on the real image for downstream execution.

In \textsc{PokeCube}, the robot needs to grasp the red end of the stick and use the stick to push the yellow cube into the red target area.
The key challenge is to distinguish the graspable part of the stick from the functional contact geometry required for pushing the cube.
The SoM marks help the VLM locate the stick, the yellow cube, and the red target area, and further generate a task-compatible grasp prior.
The final grasp should allow the robot to stably control the stick while preserving an effective contact geometry for pushing the yellow cube.

In \textsc{PullCubeTool}, the robot needs to grasp the red part of the hook and use the hook to pull the yellow cube into the red target area.
This task is particularly sensitive to the functional part of the tool: if the grasp occupies the hook tip or the curved contact region, the tool may fail to hook or pull the cube effectively.
Therefore, the SoM-guided task-prior selection first localizes the hook, the yellow cube, and the target area, and then selects a grasp region that does not interfere with the hook's functional end.
The final chosen grasp overlay shows that the system prefers a grasp location that supports stable tool control while preserving the hook interaction function.

In \textsc{PourWater}, the robot needs to grasp the red handle of the gray cup and pour its contents into the blue-gray cup.
Here, the grasp region has a clear semantic meaning: the robot should grasp the red handle rather than the cup body or rim.
Grasping the cup body may obstruct the pouring motion or induce an object pose that is unsuitable for transferring the contents into the target cup.
The SoM marks help the VLM distinguish the cup body, handle, target cup, and surrounding background regions, and identify the red handle as the preferred task-relevant grasp region.
The final selected grasp provides a more suitable initial object pose for the downstream pouring motion.

These real-robot visualizations show that SoM-guided task-prior grasp selection can translate functional semantics in language instructions into actionable image-region constraints.
It does not replace the low-level grasp detector or directly predict continuous grasp poses.
Instead, it provides task-level semantic priors for grasp-candidate filtering and grasp scoring.
In this way, GTP-FA can suppress grasps that occupy functional regions, disrupt tool-object interaction, or hinder later release or pouring motions, thereby providing a more reliable grasp-conditioned state for downstream VLA execution.

\clearpage
\subsection{Virtual environment experiment results}
\label{app:supp_results}
\FloatBarrier

This section provides additional simulation results to complement the main results in Sec.~\ref{sec:results} and Sec.~\ref{sec:ablations}. 
We include final \texttt{success\_at\_end} comparisons, full learning curves for both \texttt{success\_at\_end} and \texttt{success\_once}, closed-loop iteration curves for GTP-FA, and VLA training-loss diagnostics. 
These figures provide a more detailed view of the final performance, learning dynamics, and diagnosis-driven closed-loop improvement process. 
All experiments follow the setup described in Sec.~\ref{sec:exp_setup}; detailed protocols and hyperparameters are provided in Appendix~\ref{app:experimental_protocols}.

\subsubsection{Final success-at-end comparisons}
\label{app:final_success_bars}

\begin{figure}[H]
    \centering
    \includegraphics[width=0.82\linewidth,height=0.52\textheight,keepaspectratio]{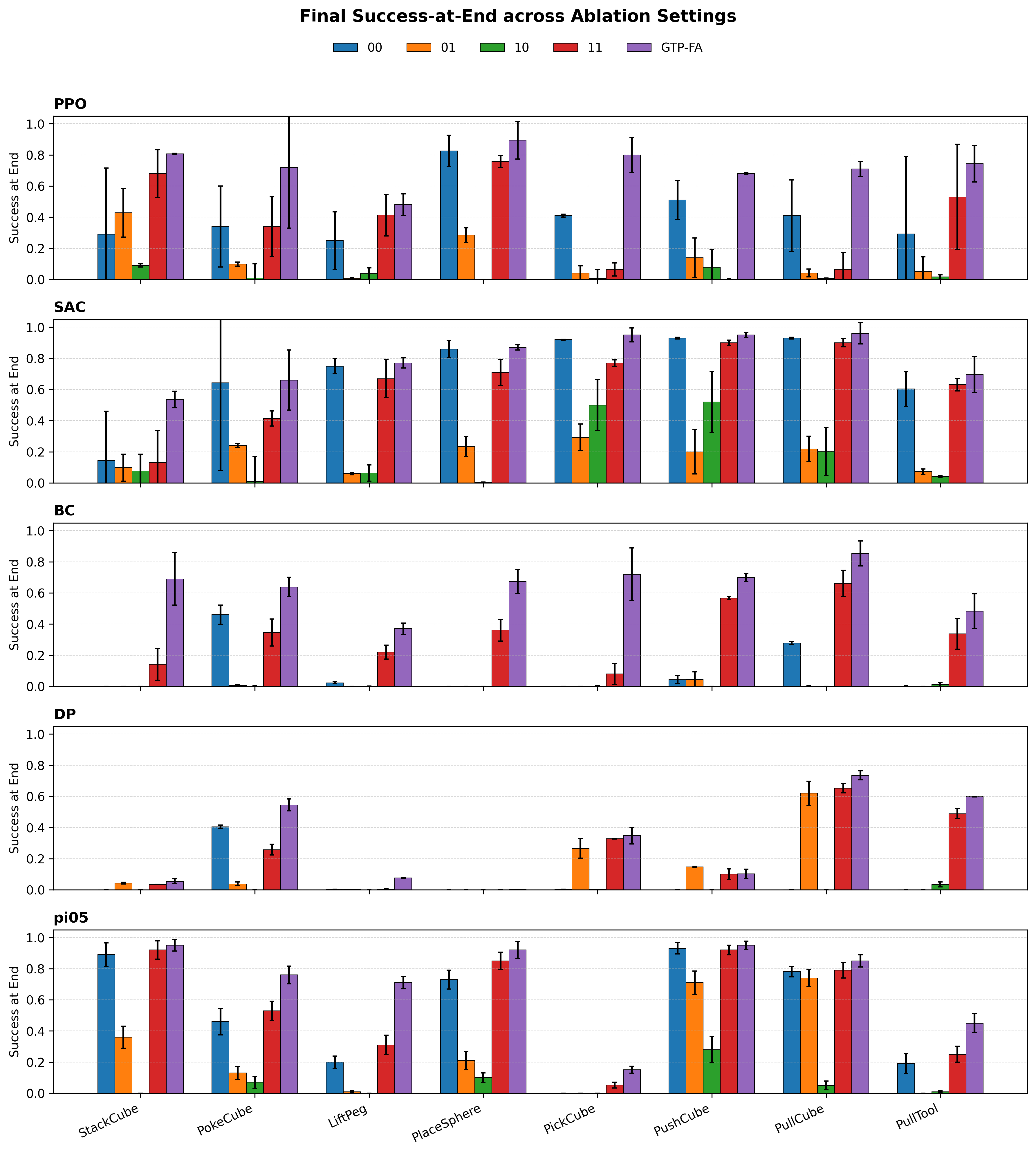}
    \caption{
    Final \texttt{success\_at\_end} across ablation settings and downstream learners.
    Each group compares the original downstream policy (\texttt{00}), planning-side-only optimization (\texttt{01}), grasp-side-only optimization (\texttt{10}), naive grasp--plan optimization without attribution (\texttt{11}), and the full GTP-FA variant.
    Bars show the mean terminal success rate across all runs for each task. Error bars indicate standard deviations. 
    GTP-FA achieves stronger and more consistent final performance across diverse downstream learners and manipulation tasks.
    }
    \label{fig:a4_final_success_bars}
\end{figure}

Figure~\ref{fig:a4_final_success_bars} summarizes the final \texttt{success\_at\_end} performance across the five downstream policy learners and eight ManiSkill3 tasks. 
The grouped bars compare the original policies, module-level ablations, and the corresponding GTP-FA variants. 
Overall, GTP-FA achieves the best or competitive final performance in most task--algorithm combinations. 
The gains are particularly pronounced for data-driven learners such as BC and DP, where the original policies are more sensitive to grasp-induced distribution shift and error attribution. 
For stronger baselines such as SAC and VLA, GTP-FA still improves several challenging tasks, indicating that diagnosis-driven routing can enhance both weak and strong downstream learners.

\begin{figure*}[t]
    \centering
    
    \begin{subfigure}[t]{0.49\linewidth}
        \centering
        \includegraphics[width=\linewidth]{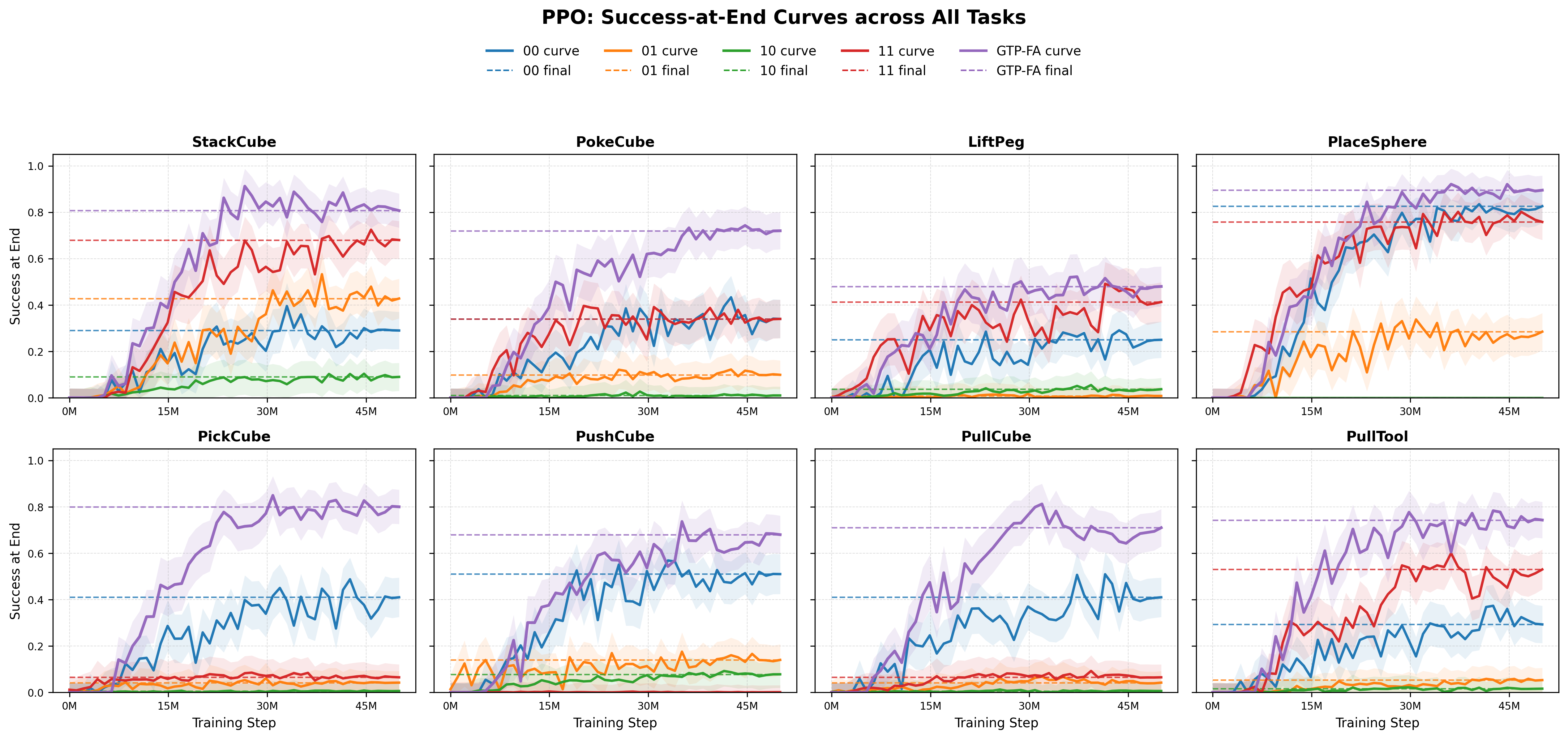}
        \caption{PPO}
        \label{fig:a4_success_end_ppo}
    \end{subfigure}
    \hfill
    \begin{subfigure}[t]{0.49\linewidth}
        \centering
        \includegraphics[width=\linewidth]{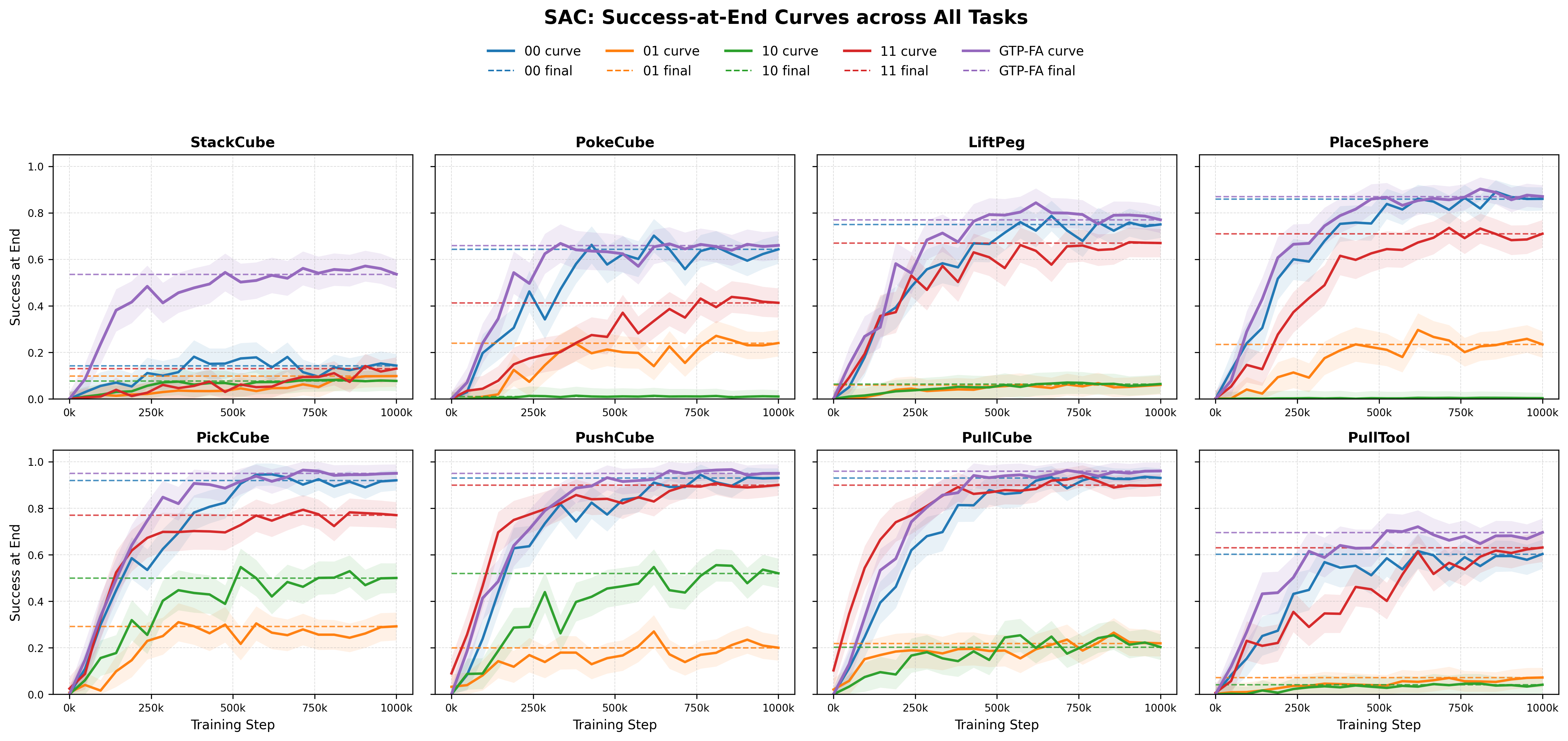}
        \caption{SAC}
        \label{fig:a4_success_end_sac}
    \end{subfigure}

    \vspace{0.5em}

    \begin{subfigure}[t]{0.49\linewidth}
        \centering
        \includegraphics[width=\linewidth]{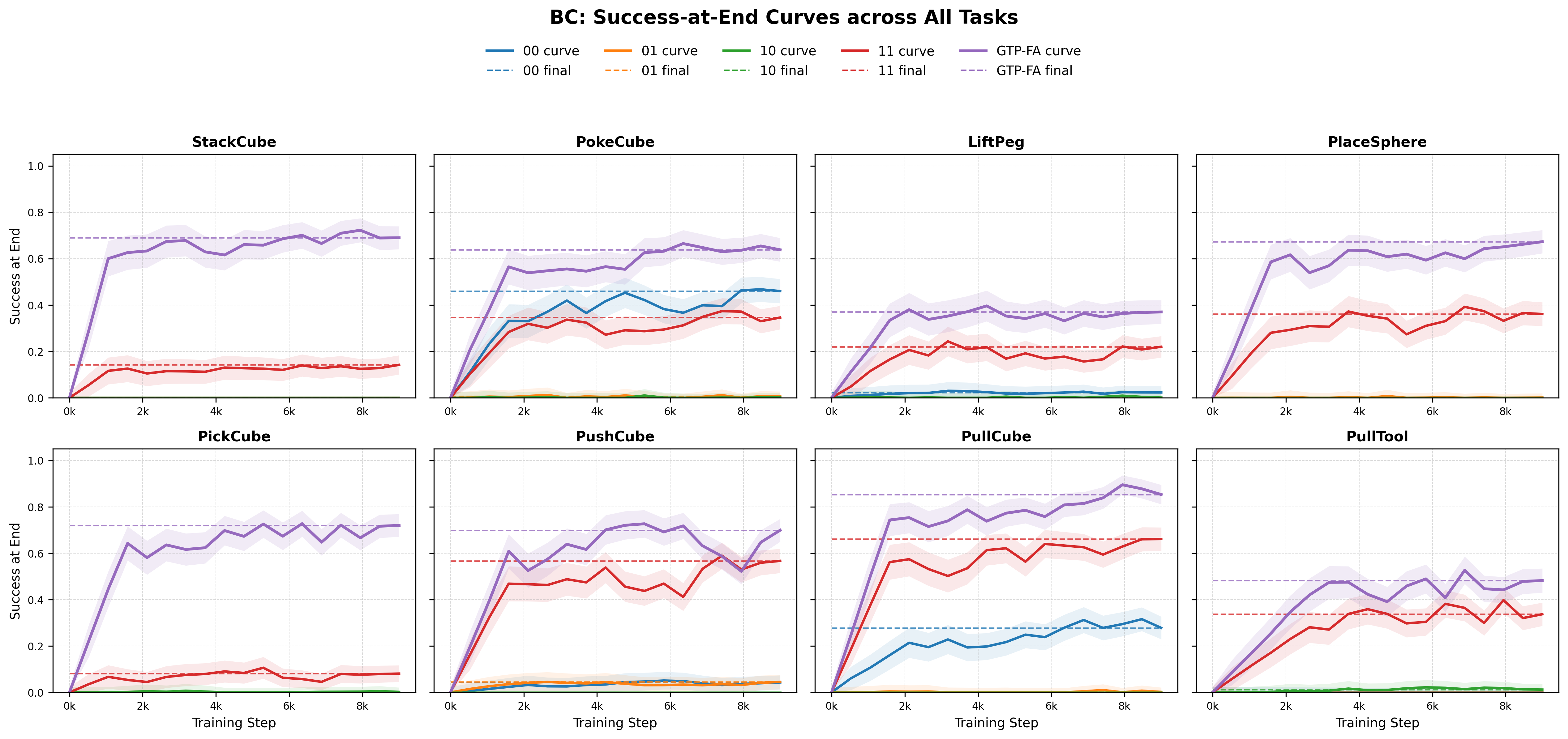}
        \caption{BC}
        \label{fig:a4_success_end_bc}
    \end{subfigure}
    \hfill
    \begin{subfigure}[t]{0.49\linewidth}
        \centering
        \includegraphics[width=\linewidth]{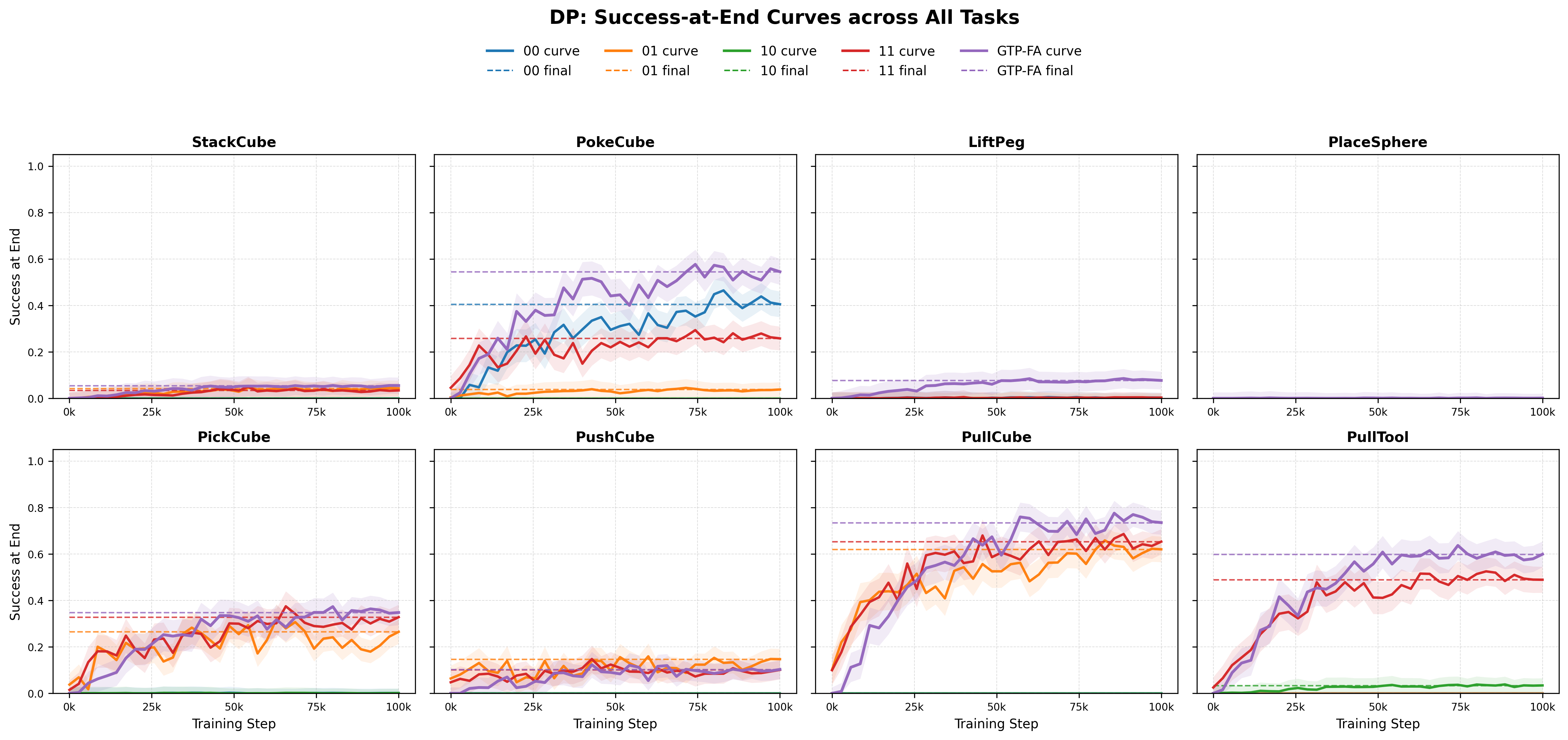}
        \caption{DP}
        \label{fig:a4_success_end_dp}
    \end{subfigure}

    \caption{
    Full \texttt{success\_at\_end} learning curves across all simulation tasks.
    Each subfigure corresponds to one downstream learner and reports the performance of five settings: the original policy (\texttt{00}), planning-side-only optimization (\texttt{01}), grasp-side-only optimization (\texttt{10}), naive grasp--plan optimization without attribution (\texttt{11}), and the full GTP-FA variant.
    Overall, GTP-FA consistently improves terminal success, convergence behavior, or training stability across diverse tasks and learning paradigms.
    }
    \label{fig:a4_success_end_all}
\end{figure*}

\begin{figure*}[t]
    \centering
    
    \begin{subfigure}[t]{0.49\linewidth}
        \centering
        \includegraphics[width=\linewidth]{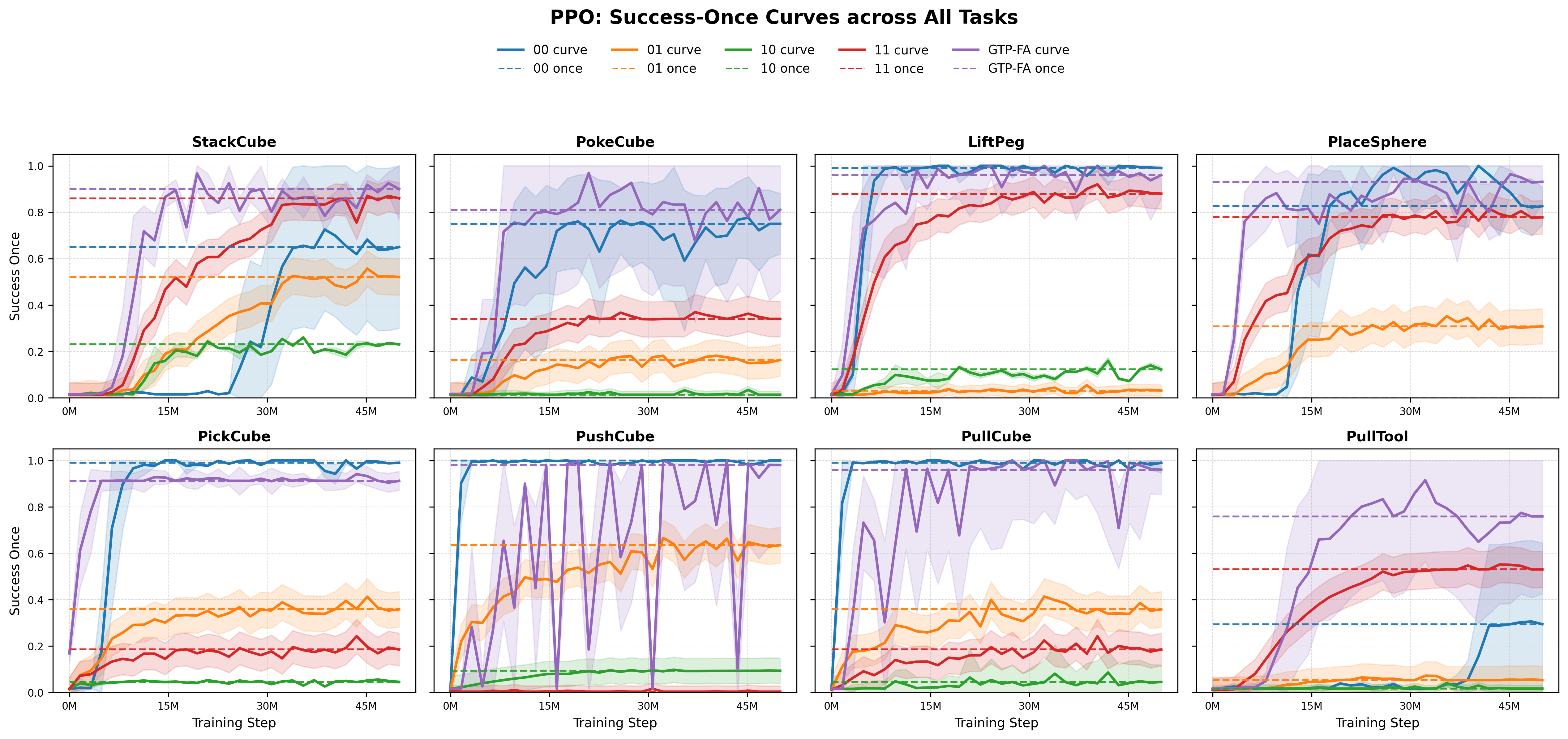}
        \caption{PPO}
        \label{fig:a4_success_once_ppo}
    \end{subfigure}
    \hfill
    \begin{subfigure}[t]{0.49\linewidth}
        \centering
        \includegraphics[width=\linewidth]{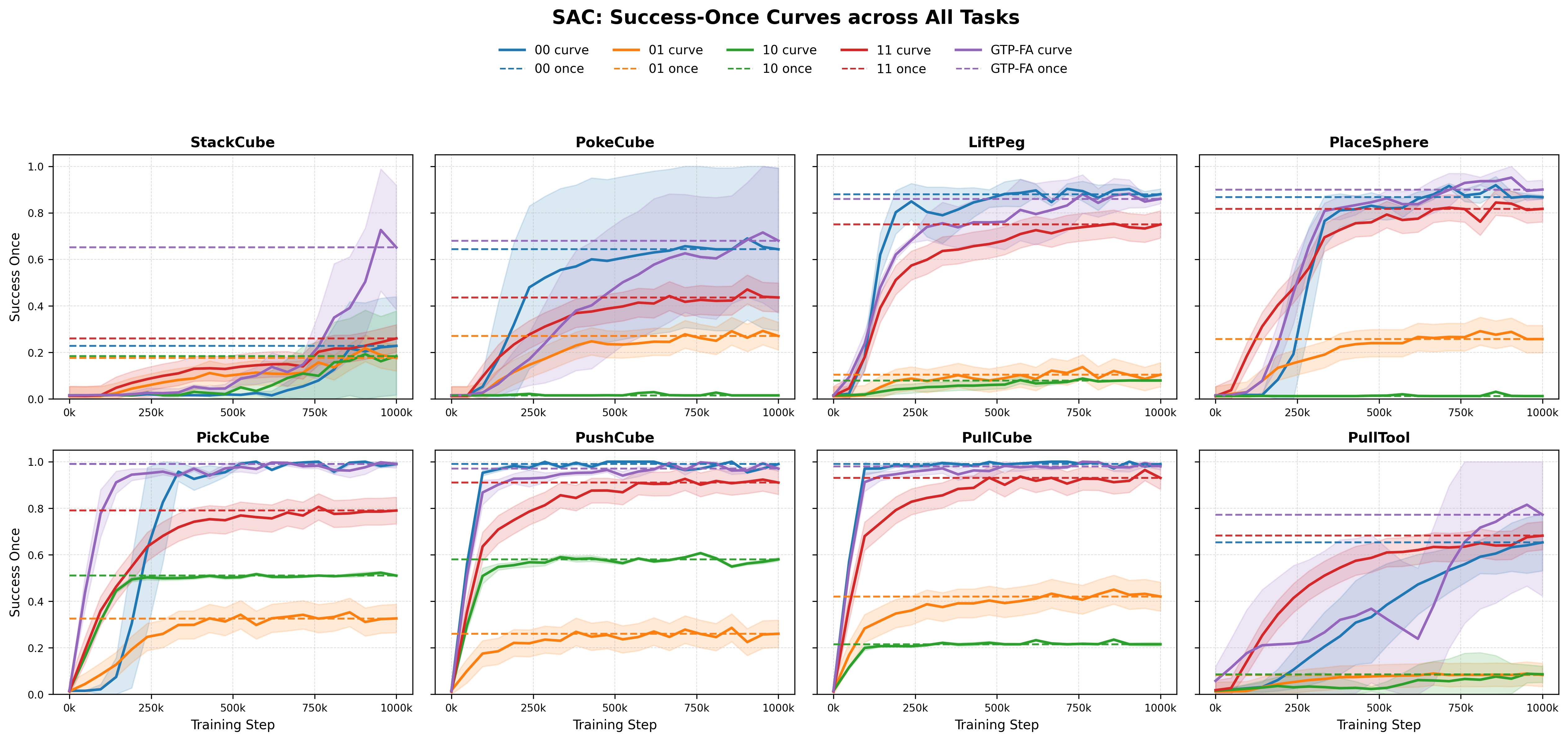}
        \caption{SAC}
        \label{fig:a4_success_once_sac}
    \end{subfigure}

    \vspace{0.5em}

    \begin{subfigure}[t]{0.49\linewidth}
        \centering
        \includegraphics[width=\linewidth]{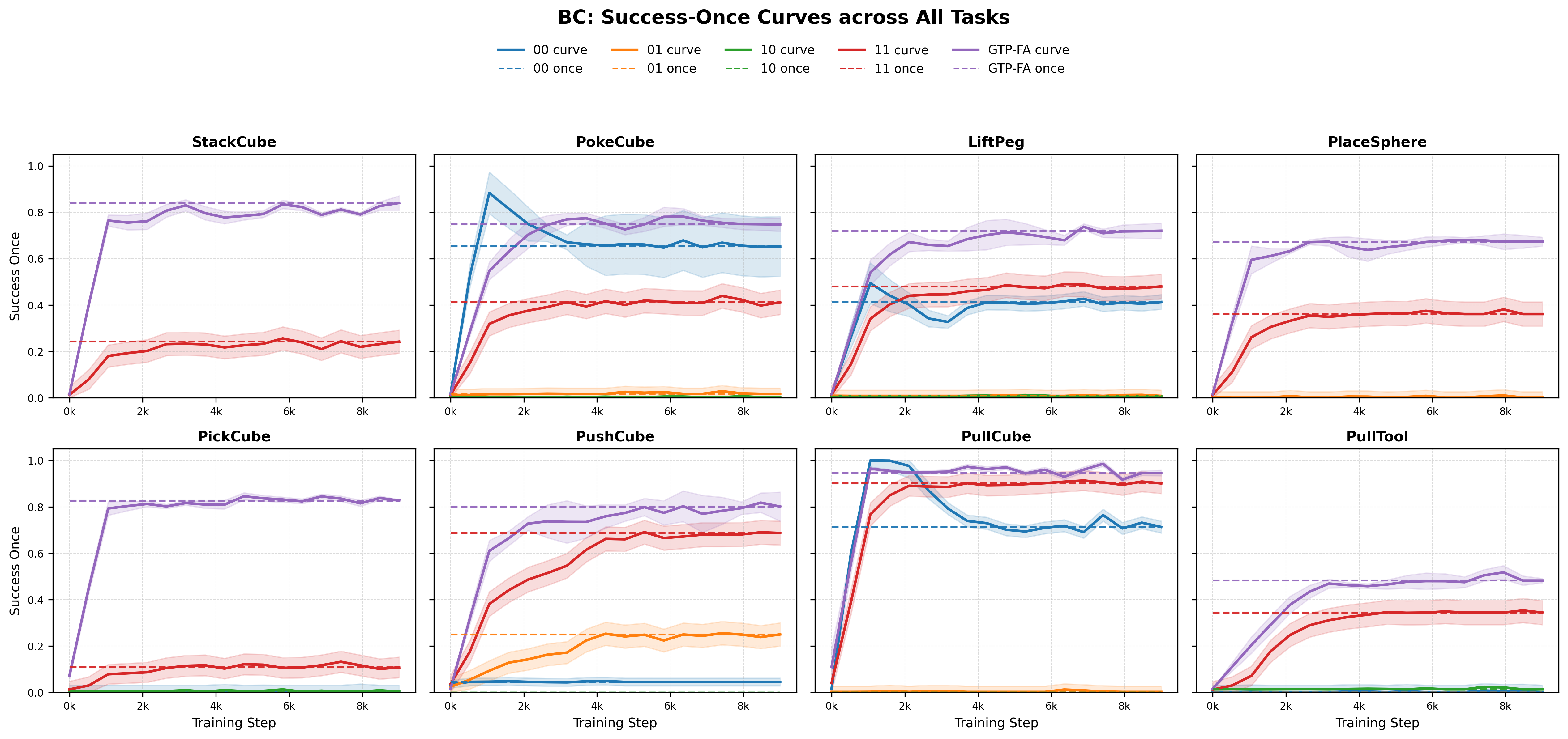}
        \caption{BC}
        \label{fig:a4_success_once_bc}
    \end{subfigure}
    \hfill
    \begin{subfigure}[t]{0.49\linewidth}
        \centering
        \includegraphics[width=\linewidth]{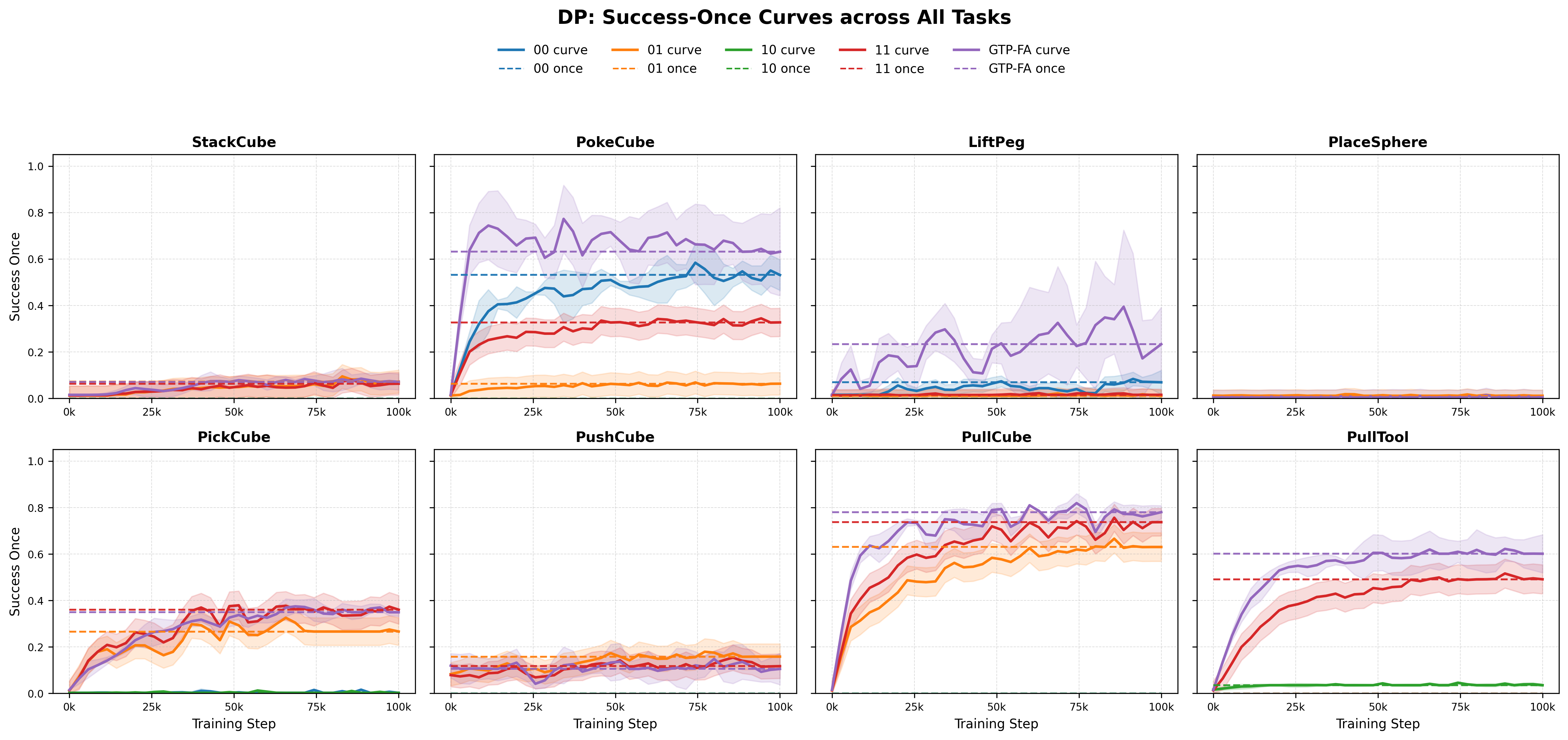}
        \caption{DP}
        \label{fig:a4_success_once_dp}
    \end{subfigure}

    \caption{
    Full \texttt{success\_once} learning curves across all simulation tasks.
    This metric measures whether the policy reaches a successful state at least once during an episode.
    Across PPO, SAC, BC, and DP, GTP-FA generally improves or stabilizes success-reaching behavior, and in conjunction with Figure~\ref{fig:a4_success_end_all}, shows that the proposed attribution-guided framework improves both reaching success and maintaining success until episode termination.
    }
    \label{fig:a4_success_once_all}
\end{figure*}

\clearpage

\subsubsection{Full success-at-end learning curves}
\label{app:success_end_curves}
\FloatBarrier

Figure~\ref{fig:a4_success_end_all} summarizes the full \texttt{success\_at\_end} learning curves across all simulation tasks for four downstream learners: PPO, SAC, BC, and DP. 
Each subfigure corresponds to one algorithm and contains the results for all eight tasks under the five settings (\texttt{00}, \texttt{01}, \texttt{10}, \texttt{11}, and GTP-FA). 
Since \texttt{success\_at\_end} requires the task to remain successful at episode termination, it provides a strict measure of execution stability in long-horizon manipulation.

Overall, GTP-FA consistently achieves stronger or more stable terminal performance across a wide range of tasks and algorithms. 
The improvements are particularly clear for PPO and BC, where diagnosis-driven routing and distribution reshaping substantially improve convergence and final success rates. 
For SAC, whose original baseline is already strong on several tasks, GTP-FA still yields gains or preserves strong performance on challenging settings. 
For DP, GTP-FA improves several tasks such as \textsc{PokeCube}, \textsc{PullCube}, and \textsc{PullTool}, although the magnitude of the gains remains task-dependent. 
These results show that the benefits of failure attribution are reflected not only in final numbers, but also in the full optimization dynamics.

\subsubsection{Full success-once learning curves}
\label{app:success_once_curves}
\FloatBarrier

Figure~\ref{fig:a4_success_once_all} shows the full \texttt{success\_once} learning curves across all simulation tasks for PPO, SAC, BC, and DP. 
Unlike \texttt{success\_at\_end}, this metric only requires the policy to reach a successful state at least once during an episode, and therefore reflects whether the policy can find successful behaviors even if it cannot always maintain them until termination.

Across most tasks and algorithms, GTP-FA improves or stabilizes \texttt{success\_once}, indicating that the proposed framework enhances both the ability to discover successful states and the robustness of subsequent policy updates. 
Comparing Figure~\ref{fig:a4_success_once_all} with Figure~\ref{fig:a4_success_end_all} further reveals that some baselines can occasionally reach success but fail to preserve it, whereas GTP-FA more consistently improves both metrics. 
This supports the central claim of the paper: attribution-guided optimization helps distinguish grasp-side and planning-side bottlenecks, leading to more effective downstream refinement.

\subsubsection{Closed-loop improvement across GTP-FA iterations}
\label{app:gtpfa_iteration_curves}
\FloatBarrier

\begin{figure*}[p]
    \centering

    \begin{subfigure}[t]{0.49\linewidth}
        \centering
        \includegraphics[width=\linewidth]{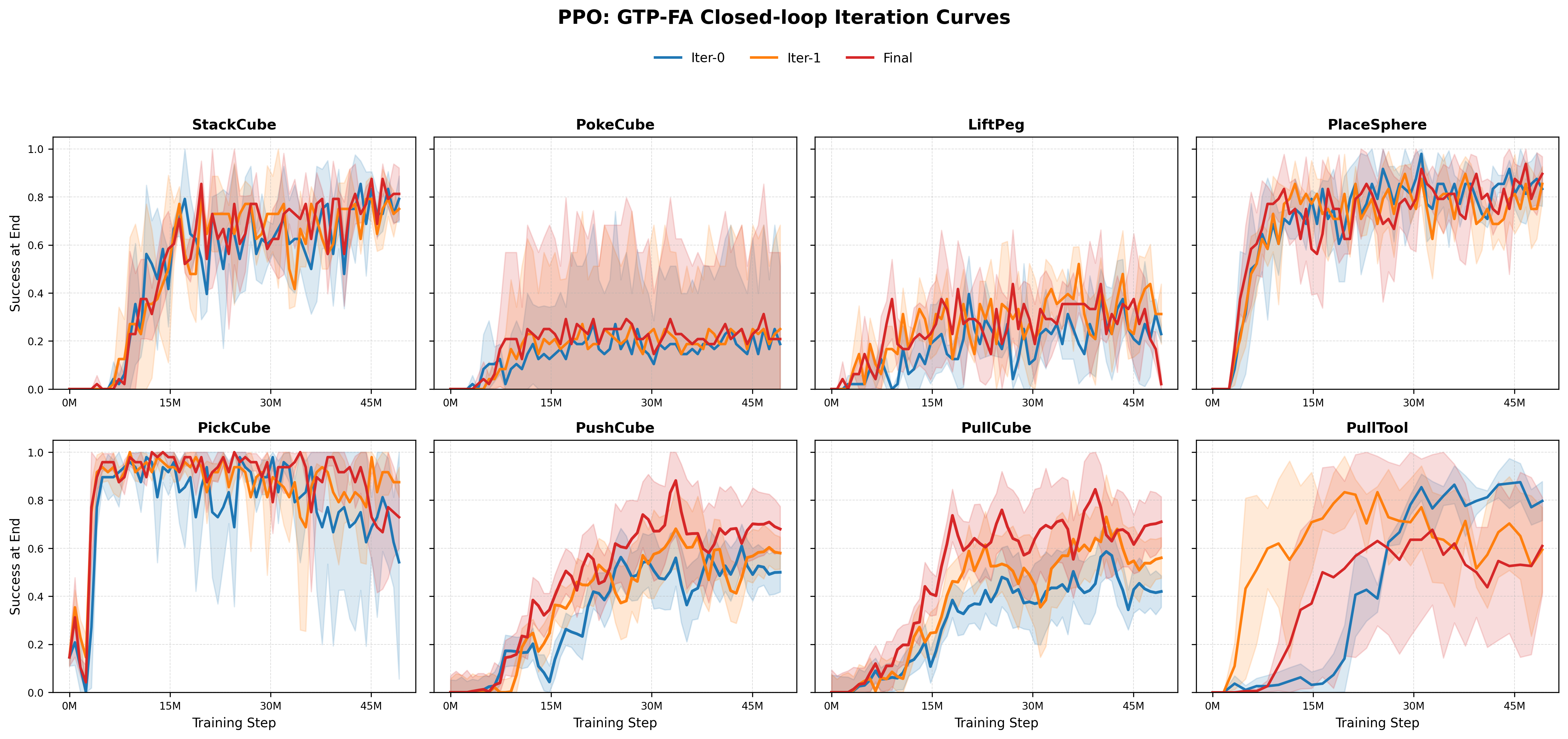}
        \caption{PPO}
        \label{fig:a4_iter_ppo}
    \end{subfigure}
    \hfill
    \begin{subfigure}[t]{0.49\linewidth}
        \centering
        \includegraphics[width=\linewidth]{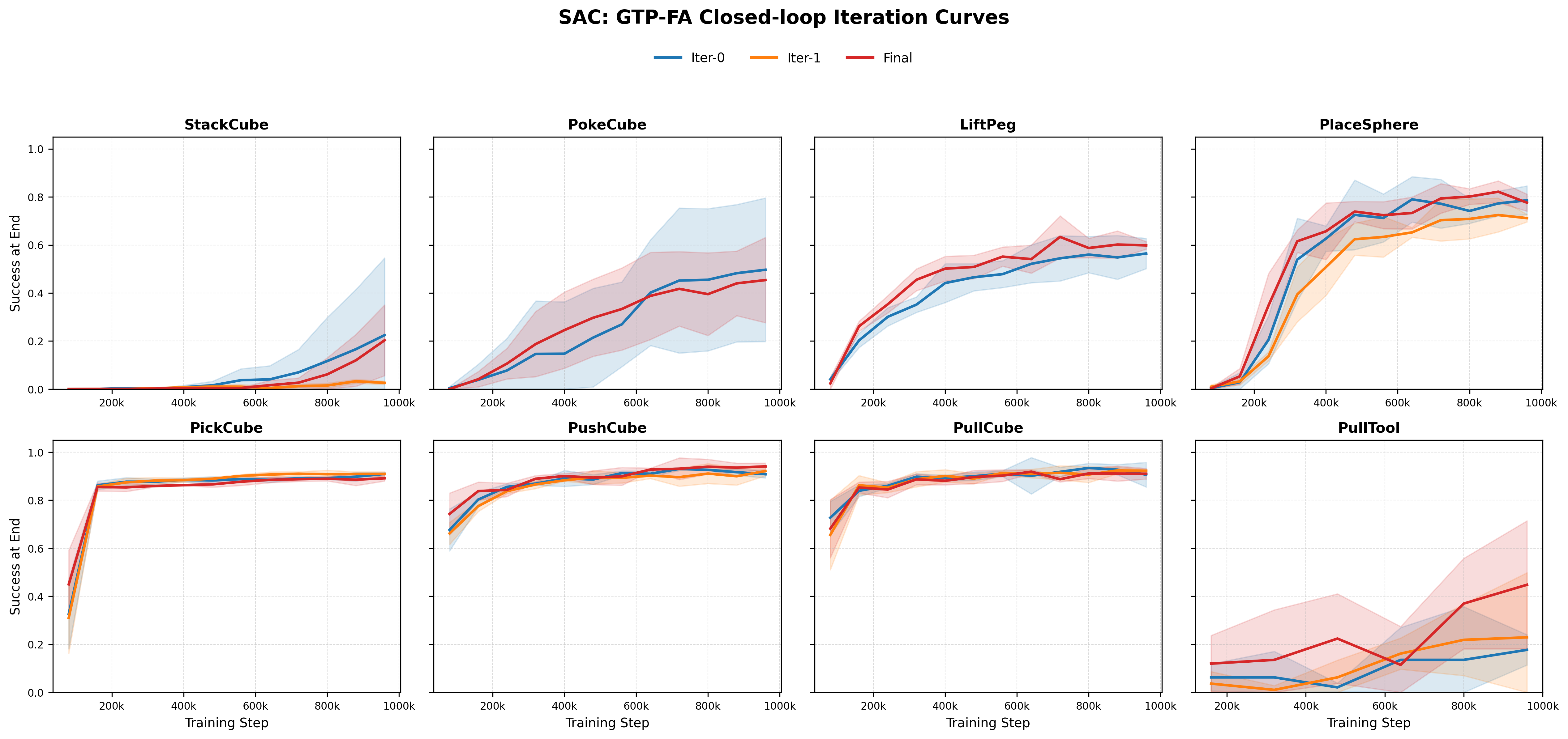}
        \caption{SAC}
        \label{fig:a4_iter_sac}
    \end{subfigure}

    \vspace{0.6em}

    \begin{subfigure}[t]{0.49\linewidth}
        \centering
        \includegraphics[width=\linewidth]{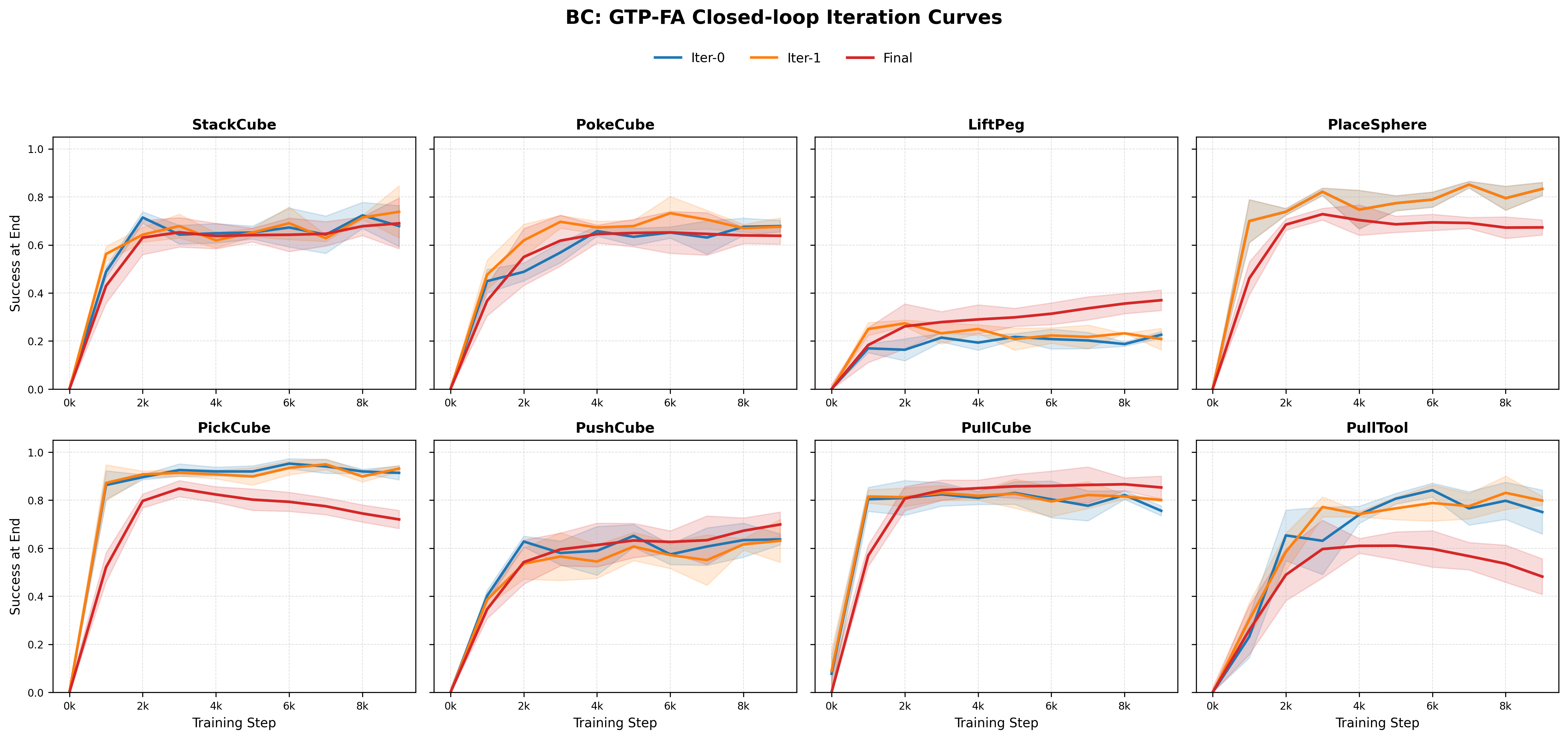}
        \caption{BC}
        \label{fig:a4_iter_bc}
    \end{subfigure}
    \hfill
    \begin{subfigure}[t]{0.49\linewidth}
        \centering
        \includegraphics[width=\linewidth]{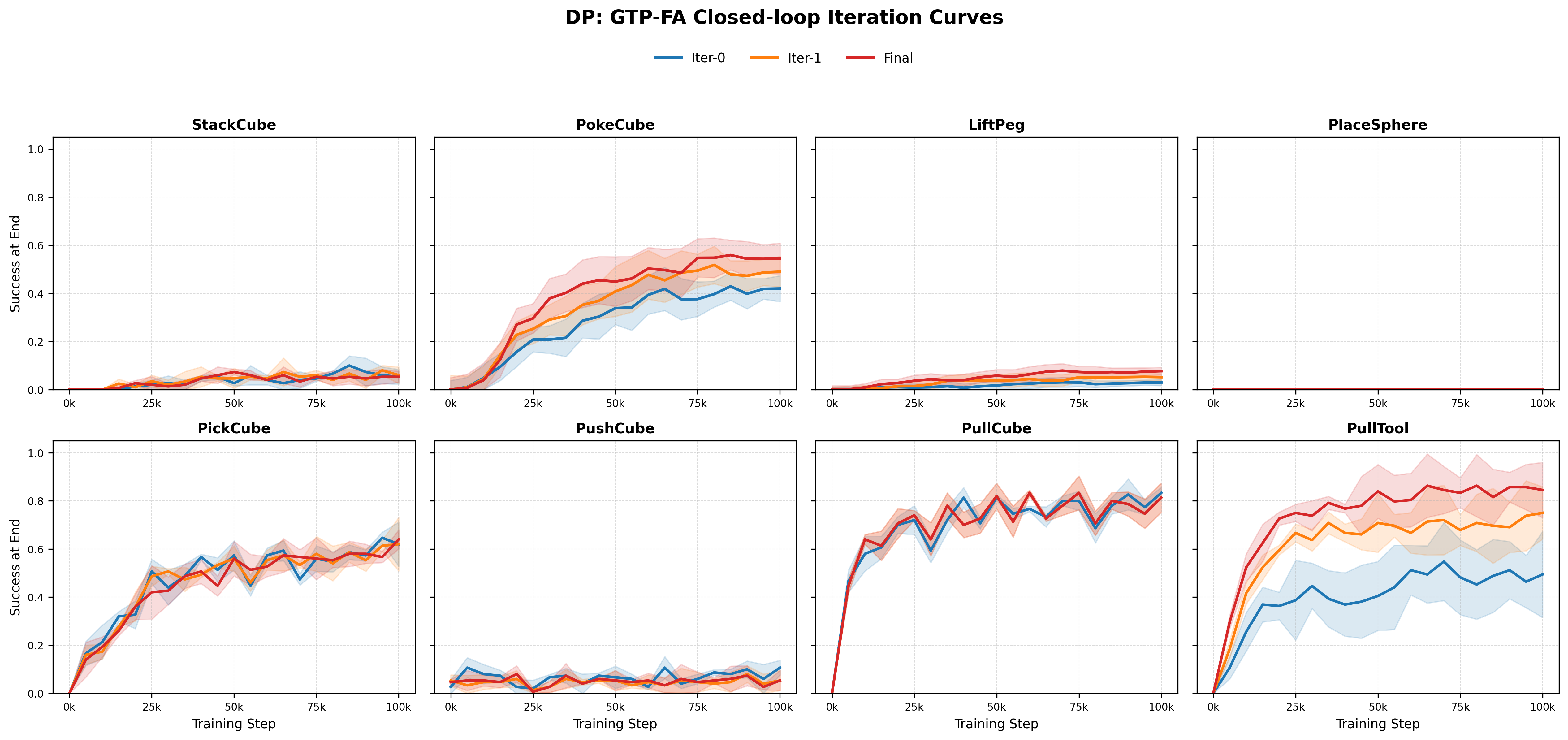}
        \caption{DP}
        \label{fig:a4_iter_dp}
    \end{subfigure}

    \caption{
    GTP-FA closed-loop iteration curves across downstream learners.
    Each subfigure shows the evolution from \texttt{iter\_00} to \texttt{iter\_01} and the final model for one base learner.
    The curves illustrate how the execute--diagnose--update loop uses failure attribution to guide hard-sample mining, distribution reshaping, and subsequent policy updates.
    Overall, later iterations or the final models improve or stabilize performance on many tasks, supporting the effectiveness of diagnosis-driven closed-loop refinement.
    }
    \label{fig:a4_iter_all}
\end{figure*}

Figure~\ref{fig:a4_iter_all} visualizes the closed-loop improvement process of GTP-FA across PPO, SAC, BC, and DP. 
Each subfigure compares \texttt{iter\_00}, \texttt{iter\_01}, and the final model for one downstream learner. 
These curves correspond to the execute--diagnose--update loop: initial rollouts are diagnosed, planning-dominant hard samples are mined, and the grasp-side and downstream policy modules are updated accordingly. 
Across many tasks, later iterations and the final model improve over the initial iteration or yield more stable terminal success, supporting the effectiveness of the diagnosis-driven closed loop. 
Iteration-to-iteration changes vary by task, and some settings show non-monotonic performance.

\subsubsection{VLA training-loss diagnostics}
\label{app:vla_loss_diagnostics}
\FloatBarrier

\begin{figure*}[t]
    \centering
    \includegraphics[width=\linewidth]{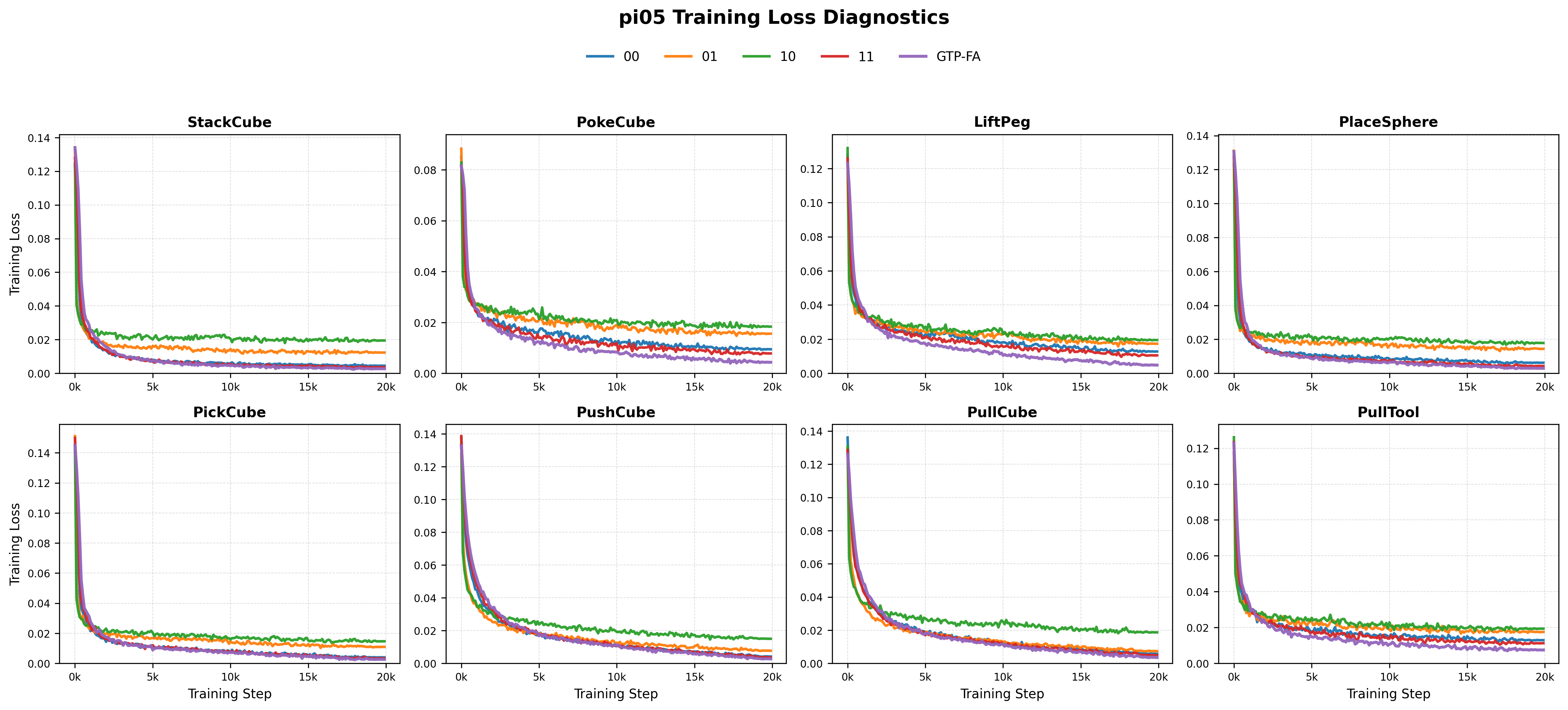}
    \caption{
    VLA ($\pi_{0.5}$) training-loss diagnostics across all simulation tasks.
    The curves show that VLA fine-tuning is stable across settings.
    GTP-FA tends to reach lower or faster-converging training loss, while the final evaluation is still determined by task success rather than loss alone.
    }
    \label{fig:a4_vla_loss_diagnostics}
\end{figure*}

Figure~\ref{fig:a4_vla_loss_diagnostics} reports training-loss curves for $\pi_{0.5}$ under different ablation settings.
All settings exhibit a rapid loss decrease during the early training phase followed by gradual convergence, indicating that the fine-tuning procedure is stable.
GTP-FA often reaches a lower final training loss or enters a low-loss region faster, suggesting that diagnosis-driven data restructuring provides more informative fine-tuning samples.
We emphasize that these loss curves are used only as training diagnostics; the main conclusions are based on task-level success metrics, where GTP-FA-$\pi_{0.5}$ also achieves clear improvements over the original $\pi_{0.5}$ policy.

\paragraph{Summary.}
The supplementary curves lead to three observations.
First, GTP-FA improves final terminal success across a broad range of algorithms and tasks.
Second, the full learning curves show that these gains are reflected not only in final numbers but also in training dynamics, where GTP-FA often improves convergence speed, asymptotic performance, or stability.
Third, the closed-loop iteration curves support the core design of GTP-FA: failure attribution enables the system to identify whether the bottleneck lies in grasping or downstream planning, and to use this diagnosis to guide subsequent data reshaping and policy updates.

\clearpage
\subsection{Real-World Experimental Results}
\label{app:real_world_results}
\FloatBarrier

\begin{figure*}[htp!]
    \centering
    \includegraphics[width=\linewidth]{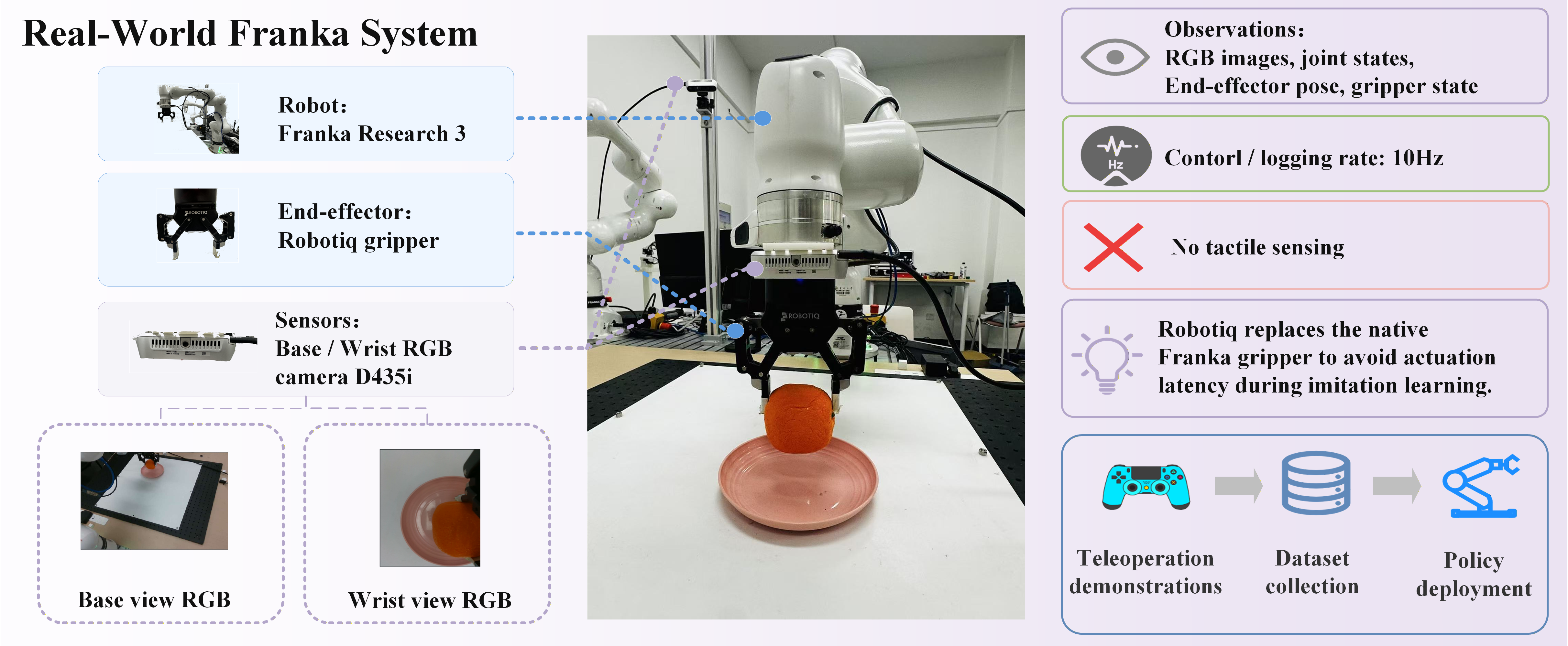}
    \caption{
    Real-world Franka system used in our experiments.
    The figure shows the Franka Research 3 platform, Robotiq gripper, base/wrist D435i cameras, key observation and control settings, and the teleoperation-to-deployment pipeline.
    }
    \label{fig:real_setup}
\end{figure*}

This section provides supplementary real-world results on the Franka Research 3 platform.
The main paper reports quantitative results on five real-world manipulation tasks: placing an orange into a pink tray, stacking a blue cube onto an orange cube, pushing a yellow cube with a stick, pulling a yellow cube with a hook, and grasping the red handle of a gray cup to pour its contents into a blue-gray cup.
Here, we further present real-system observations, grasp-planning visualizations, and representative success and failure rollouts to illustrate the behavioral differences between the original $\pi_{0.5}$ baseline and GTP-FA-$\pi_{0.5}$.
Figure~\ref{fig:real_setup} provides an overview of the real-world hardware platform, sensing configuration, and teleoperation-to-deployment pipeline used in our experiments.

\subsubsection{Real-robot execution interface, tasks, and demonstration data}
\label{app:real_interface}

In this section, we consistently denote the real-robot VLA baseline as the original $\pi_{0.5}$ policy, and our method as GTP-FA-$\pi_{0.5}$.
The real-world experiments use a Franka Research 3 arm equipped with a Robotiq gripper, base/wrist cameras, and a VLA policy deployment interface.
Figure~\ref{fig:real_setup} summarizes the real-world hardware platform used in our experiments.
Here we further focus on the online execution interface and task-level observations.
The interface shows the base-camera view, wrist-camera view, robot state traces, and the task prompt used by the policy.

For $\pi_{0.5}$-based experiments, the simulated tasks use 100 converted expert trajectories for each fine-tuning setting.
For the real-robot setting, both the original $\pi_{0.5}$ baseline and GTP-FA-$\pi_{0.5}$ use the same 300 real expert trajectories per task (1,500 across five tasks).
This ensures that the comparison between the two real-robot methods reflects the effect of explicit grasp selection and diagnosis-driven optimization, rather than differences in demonstration data.

We evaluate five real-world manipulation tasks.
In the orange-to-tray task, the robot receives the instruction ``place the grasped orange into the pink tray'' and must complete a long-horizon manipulation sequence involving approach, grasping, lifting, transport, tray alignment, release, and withdrawal.
In the cube-stacking task, the robot receives an instruction such as ``pick up the blue cube and stack it on the orange cube,'' and must grasp the blue cube, lift it, align it above the orange cube, release it, and maintain a stable stacked configuration.
In the \textsc{PokeCube} task, the robot grasps the red end of a stick and pushes a yellow cube into a red target area.
In the \textsc{PullCubeTool} task, the robot grasps the red part of a hook and pulls a yellow cube into a red target area.
In the \textsc{PourWater} task, the robot grasps the red handle of a gray cup and pours its contents into a blue-gray cup.
These tasks require reliable grasp-conditioned execution, but expose different sources of difficulty: orange-to-tray stresses stable grasping and release, cube stacking stresses small-object localization and precise alignment, tool-use tasks stress functional-part preservation, and pouring further requires a grasp-induced cup pose suitable for transport, alignment, and tilting.

\begin{figure*}[t]
    \centering
    \begin{subfigure}[t]{0.49\linewidth}
        \centering
        \includegraphics[width=\linewidth]{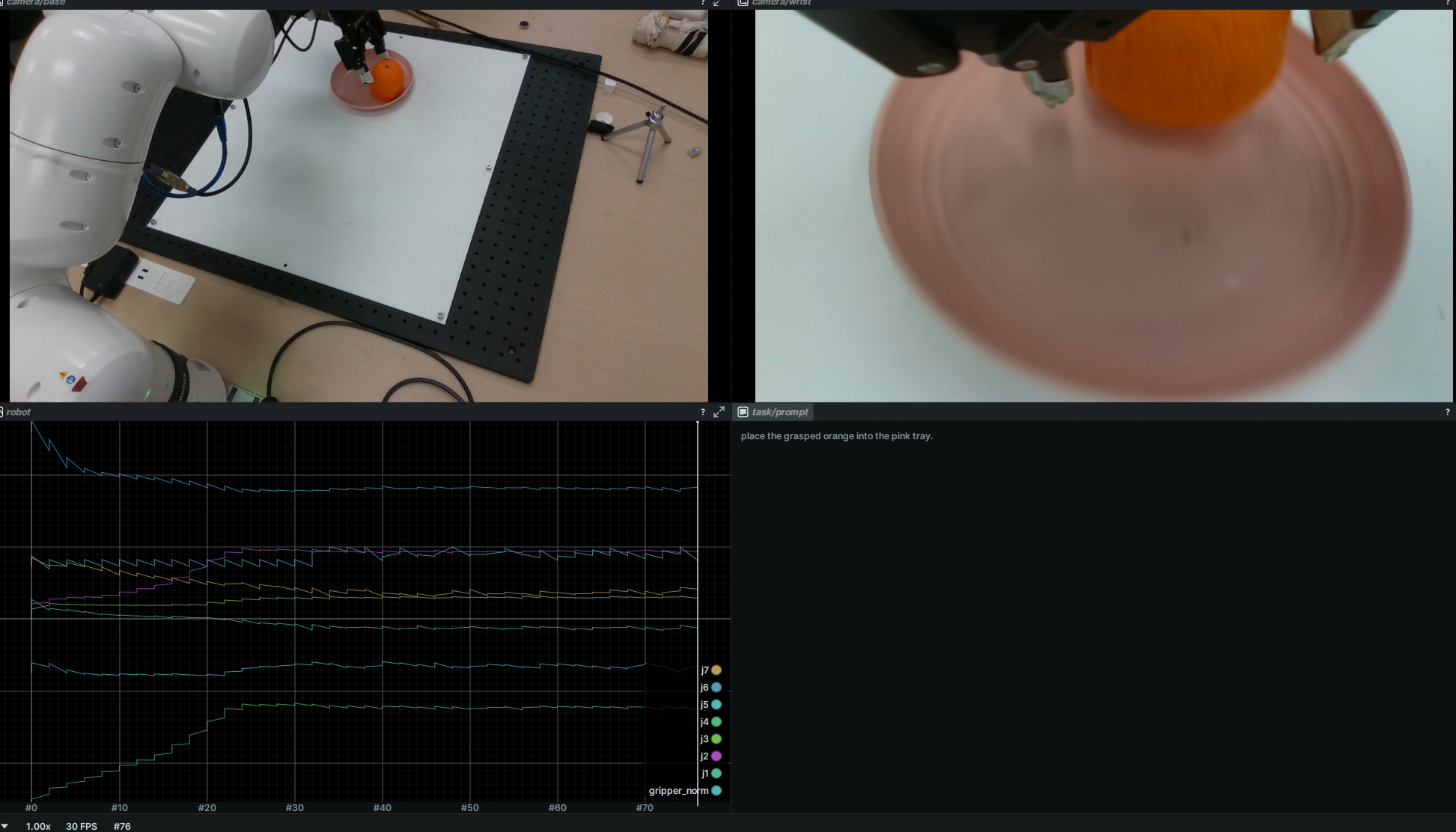}
        \caption{GTP-FA-$\pi_{0.5}$ succeeds on the orange-to-tray task.}
        \label{fig:real_interface_orange_gtpfa}
    \end{subfigure}
    \hfill
    \begin{subfigure}[t]{0.49\linewidth}
        \centering
        \includegraphics[width=\linewidth]{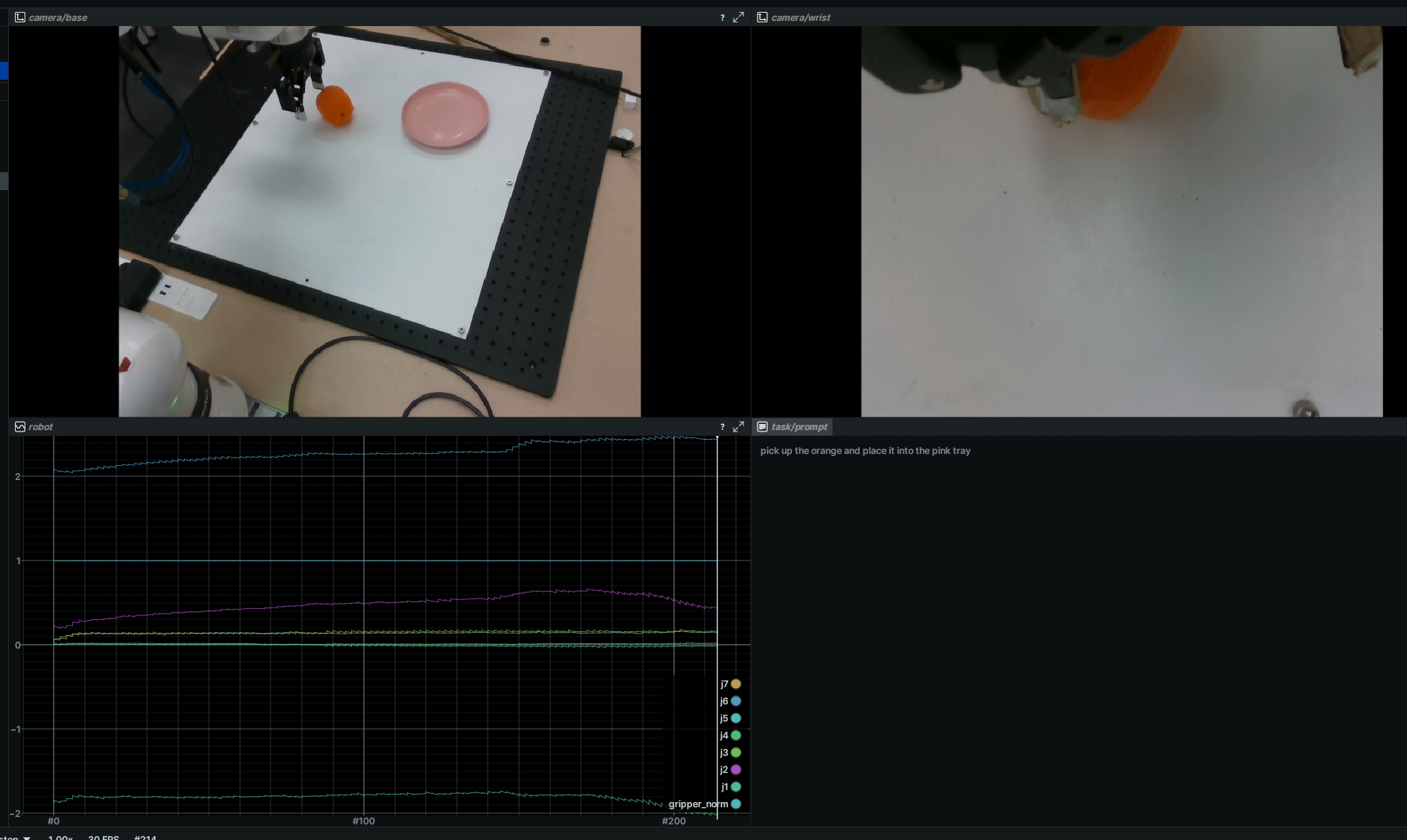}
        \caption{Original $\pi_{0.5}$ fails on the orange-to-tray task.}
        \label{fig:real_interface_orange_vla}
    \end{subfigure}

    \vspace{0.5em}

    \begin{subfigure}[t]{0.49\linewidth}
        \centering
        \includegraphics[width=\linewidth]{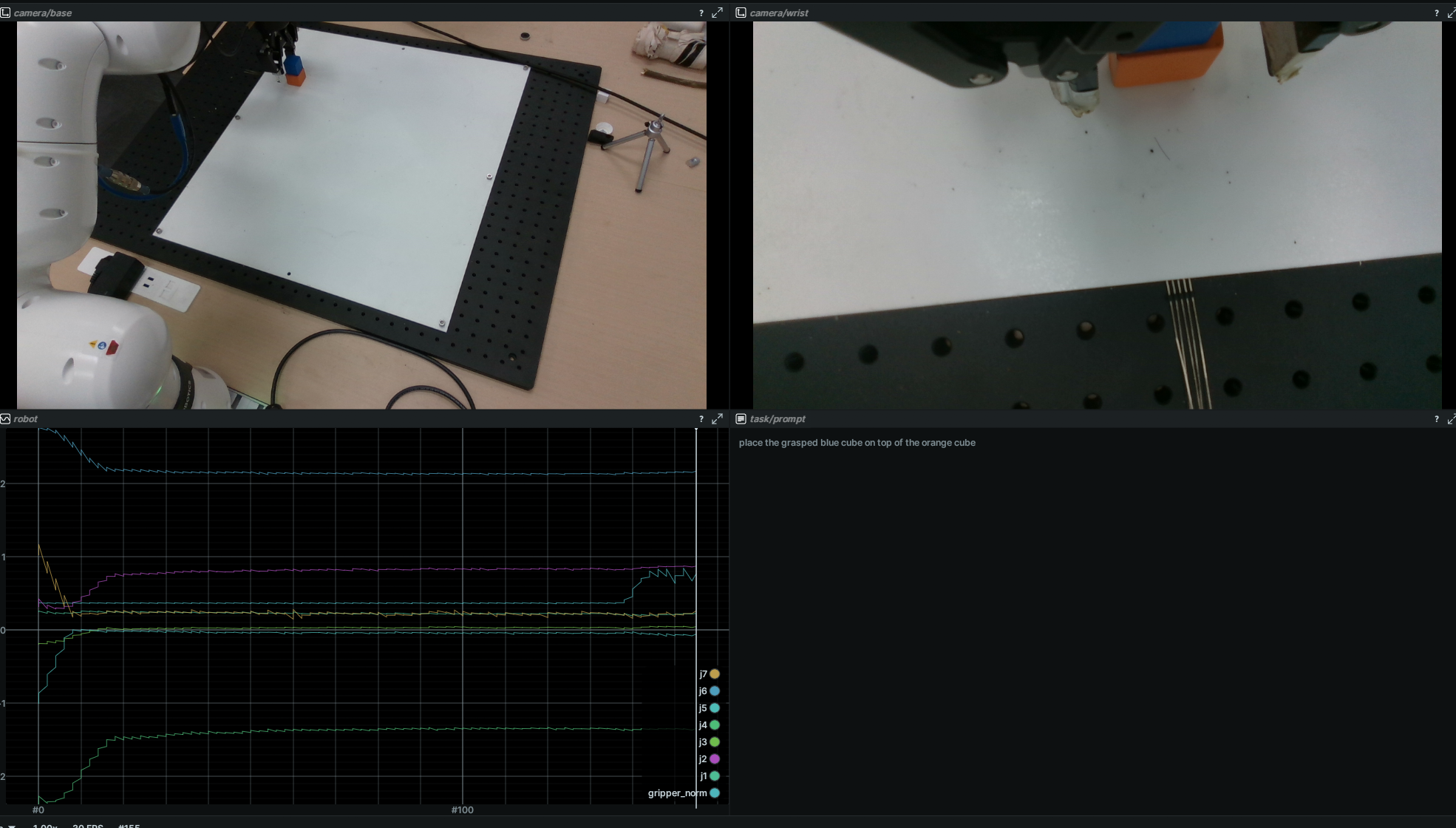}
        \caption{GTP-FA-$\pi_{0.5}$ succeeds on the cube-stacking task.}
        \label{fig:real_interface_stack_gtpfa}
    \end{subfigure}
    \hfill
    \begin{subfigure}[t]{0.49\linewidth}
        \centering
        \includegraphics[width=\linewidth]{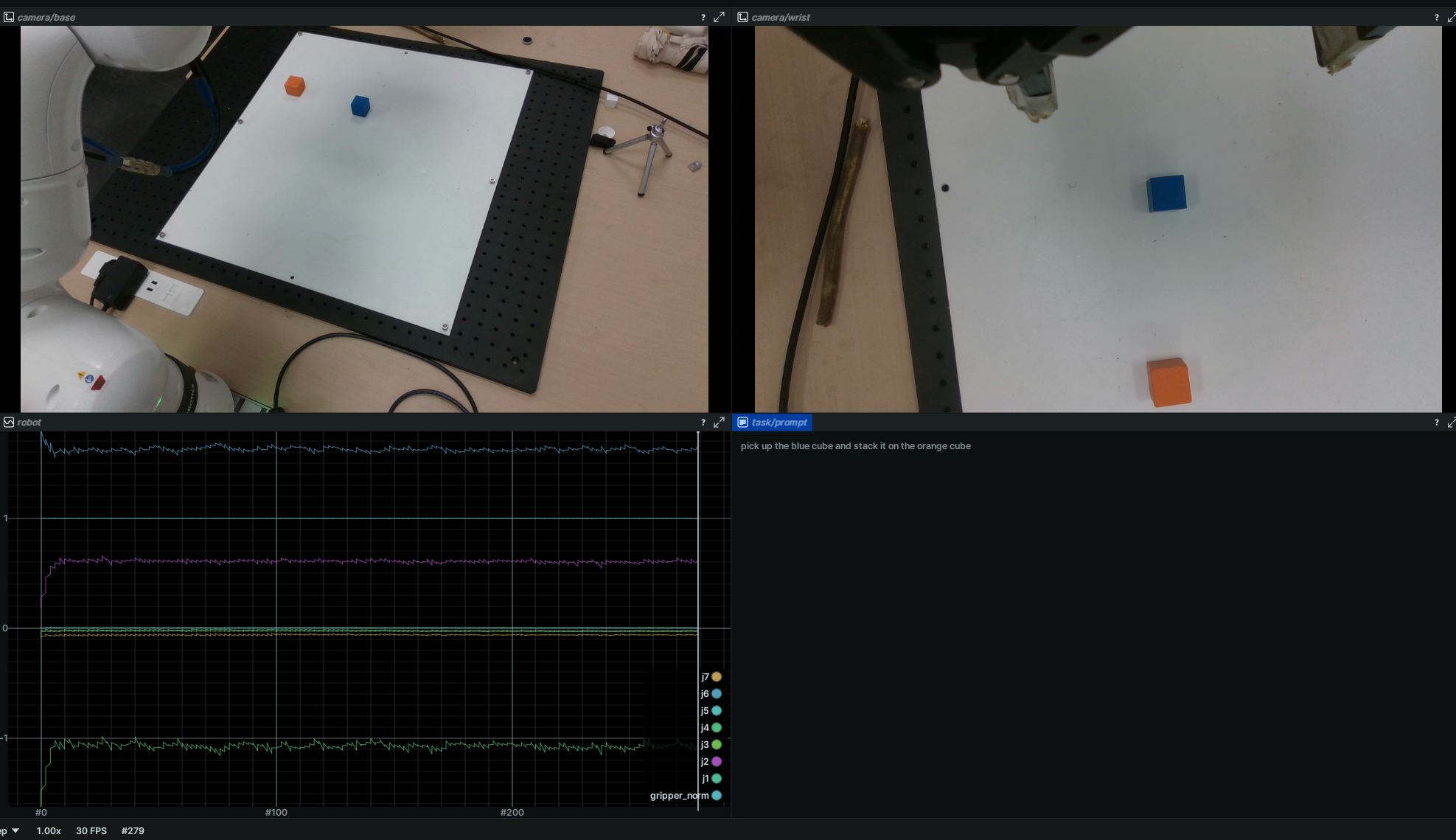}
        \caption{Original $\pi_{0.5}$ fails on the cube-stacking task.}
        \label{fig:real_interface_stack_vla}
    \end{subfigure}

    \caption{
    Real-robot execution interfaces for the two representative manipulation tasks.
    Each interface shows the base-camera view, wrist-camera view, robot state traces, and the task prompt used by the policy.
    The original $\pi_{0.5}$ baseline directly predicts actions from visual observations and language prompts, whereas GTP-FA-$\pi_{0.5}$ first selects a task-compatible grasp and then initializes downstream execution from the selected grasp-conditioned state.
    }
    \label{fig:real_interfaces_two_tasks}
\end{figure*}

\subsubsection{GraspNet candidates versus selected execution grasp}
\label{app:real_grasp_visualization}

\begin{figure*}[t]
    \centering
    \begin{subfigure}[t]{0.24\linewidth}
        \centering
        \includegraphics[width=\linewidth]{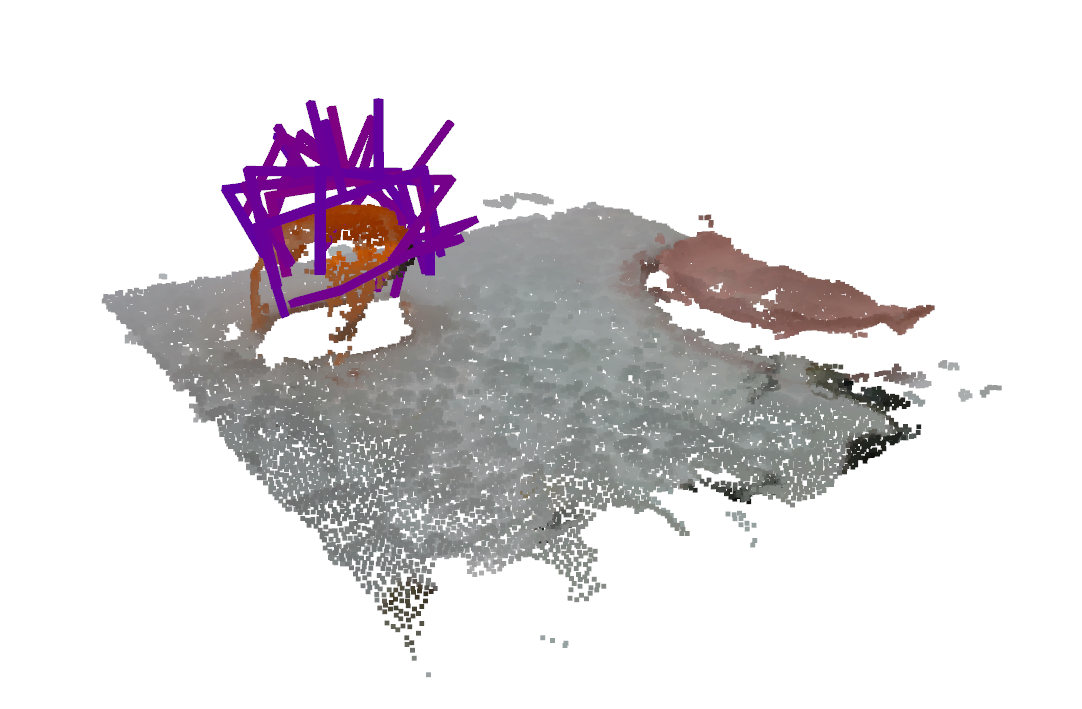}
        \caption{Raw GraspNet candidates for orange-to-tray.}
        \label{fig:orange_grasp_raw}
    \end{subfigure}
    \hfill
    \begin{subfigure}[t]{0.24\linewidth}
        \centering
        \includegraphics[width=\linewidth]{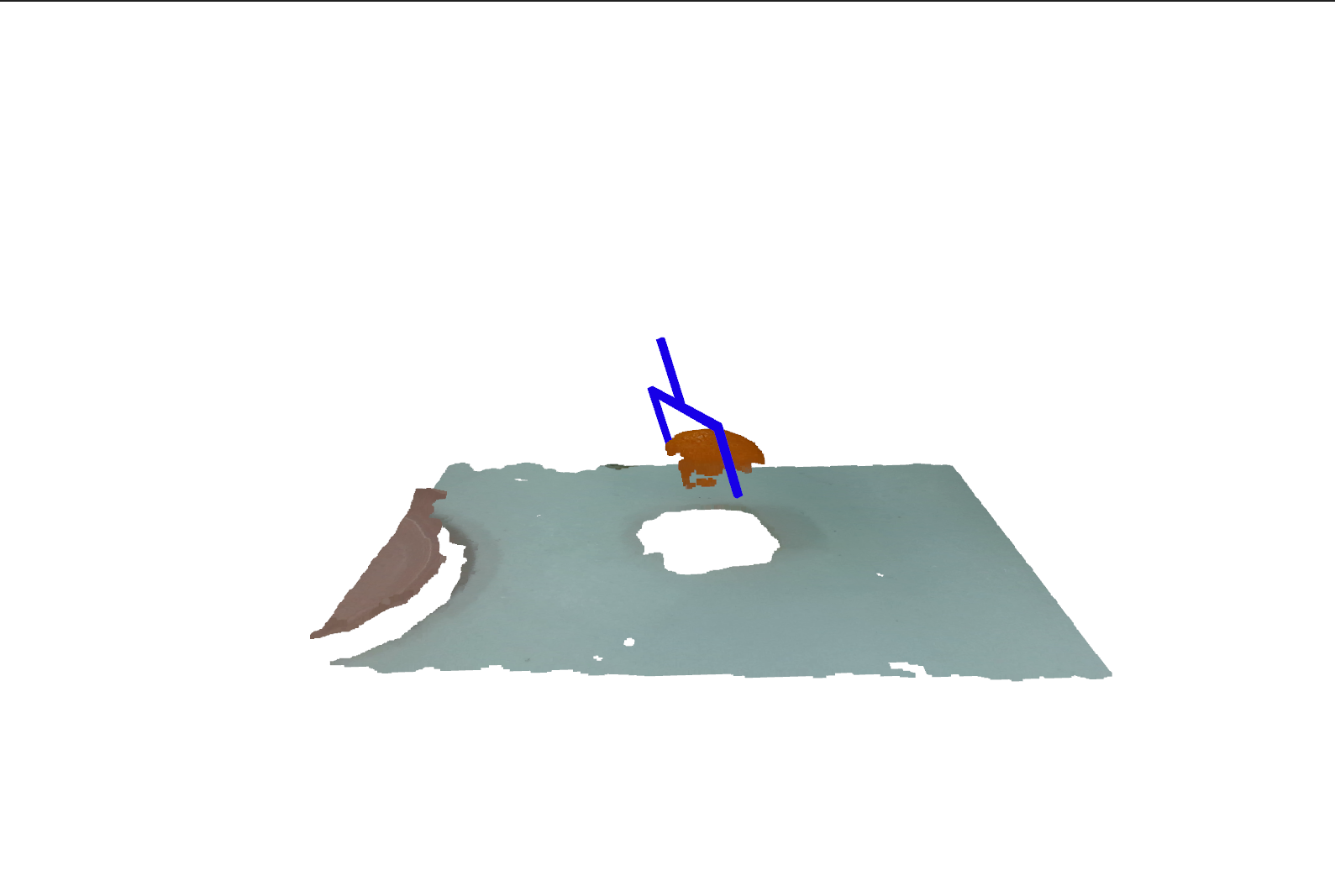}
        \caption{Single selected grasp for orange-to-tray.}
        \label{fig:orange_grasp_selected}
    \end{subfigure}
    \hfill
    \begin{subfigure}[t]{0.24\linewidth}
        \centering
        \includegraphics[width=\linewidth]{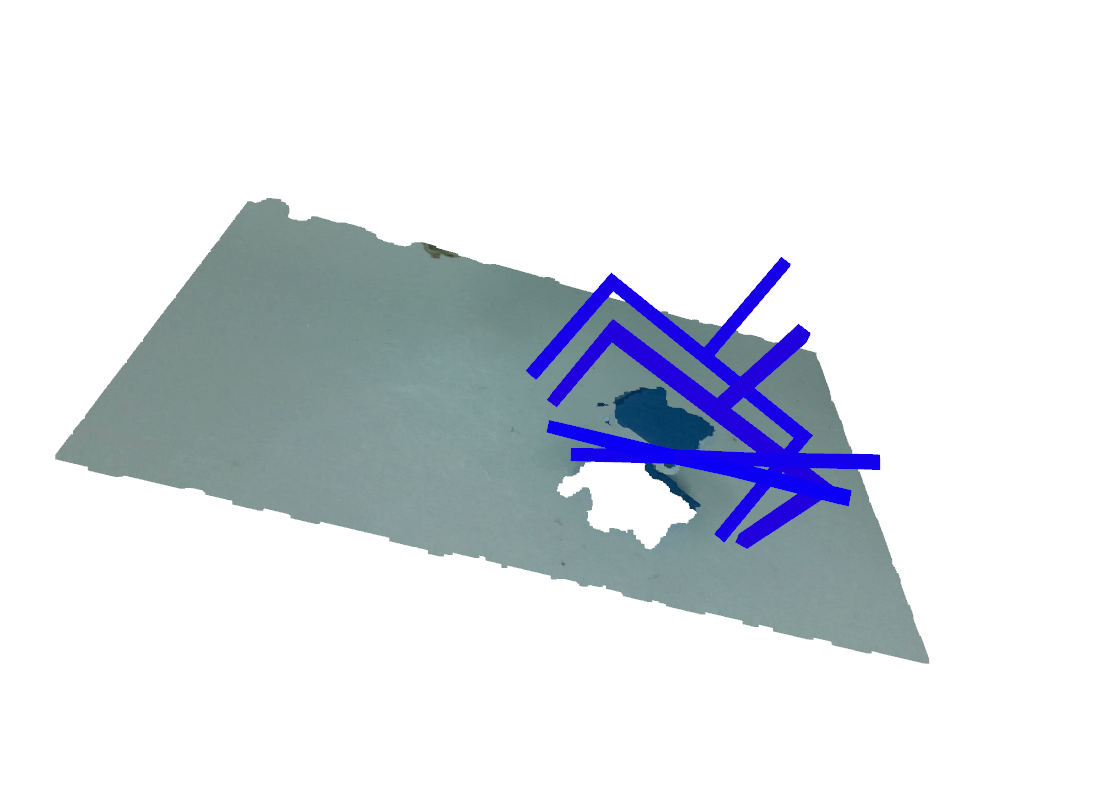}
        \caption{Raw GraspNet candidates for cube stacking.}
        \label{fig:stack_grasp_raw}
    \end{subfigure}
    \hfill
    \begin{subfigure}[t]{0.24\linewidth}
        \centering
        \includegraphics[width=\linewidth]{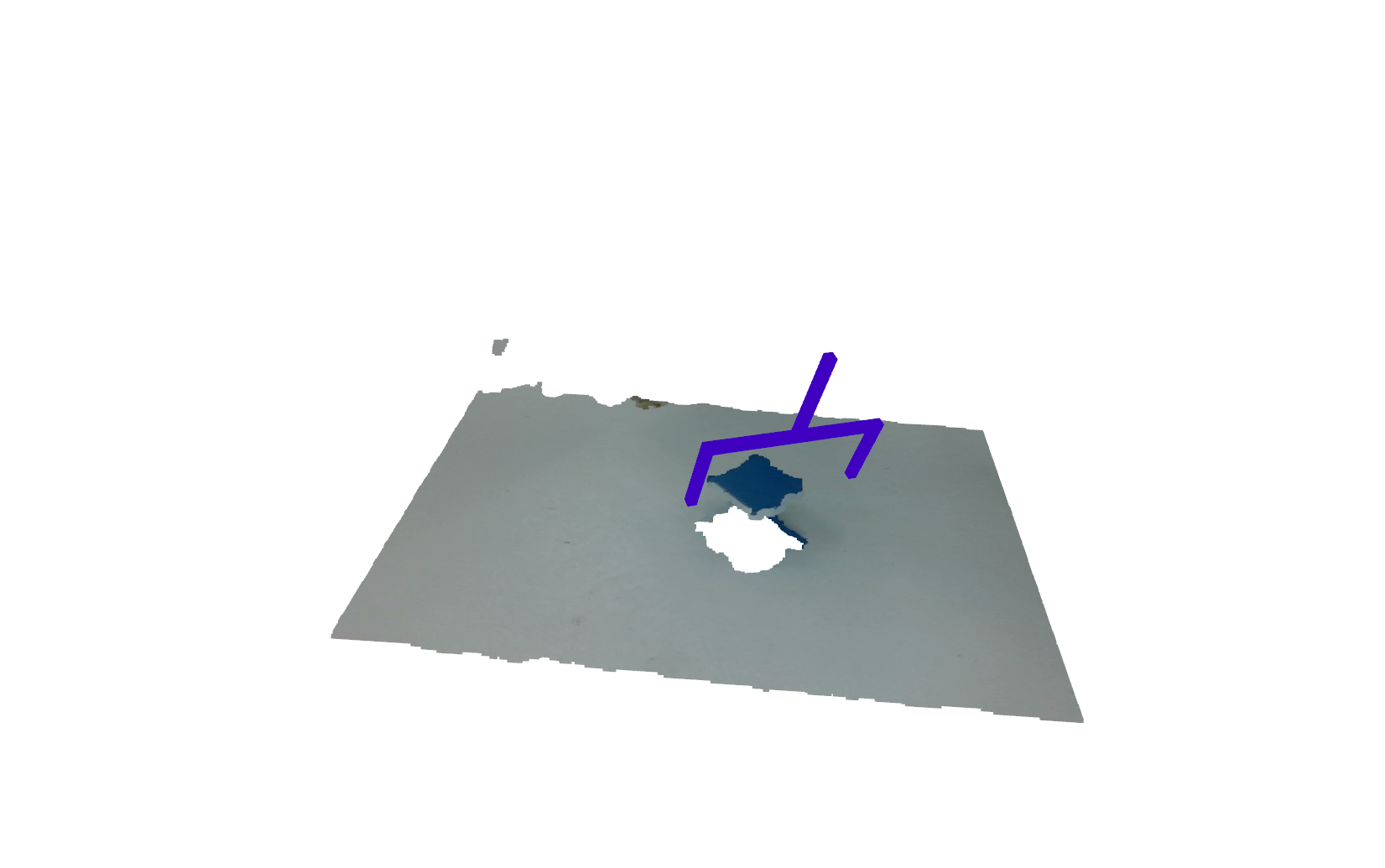}
        \caption{Selected execution grasp for cube stacking.}
        \label{fig:stack_grasp_selected}
    \end{subfigure}

    \caption{
    Raw GraspNet candidates versus the selected unique execution grasp on real observations.
    The raw GraspNet output contains multiple geometrically feasible grasp poses, but not all of them are suitable for the downstream task.
    GTP-FA applies task-prior filtering and diagnostic risk calibration to select a single task-compatible grasp, providing a more reliable post-grasp initial condition for $\pi_{0.5}$ execution.
    }
    \label{fig:real_grasp_candidates_selected}
\end{figure*}

Figure~\ref{fig:real_grasp_candidates_selected} visualizes the difference between the raw GraspNet output and the final grasp used by GTP-FA.
Given real camera observations and the reconstructed target-object point cloud, GraspNet first proposes a set of geometrically feasible candidate grasps.
However, these raw candidates are not necessarily task-compatible: some may approach the object from an unfavorable direction, occupy a region required for subsequent placement or release, or lead to an unstable post-grasp state.
GTP-FA therefore applies task-prior filtering and diagnostic risk calibration to select a single execution grasp that is more suitable for the downstream task.

For the orange-to-tray task, the selected grasp should not only lift the orange stably, but also preserve a post-grasp pose that allows the downstream $\pi_{0.5}$ policy to transport and release the orange into the tray.
For the cube-stacking task, the selected grasp must be more precise: it should grasp the blue cube stably while leaving the cube pose suitable for accurate placement on the orange cube.
GTP-FA-$\pi_{0.5}$ selects one execution grasp from the GraspNet candidates using task-prior filtering and diagnostic risk calibration.

For \textsc{PokeCube}, \textsc{PullCubeTool}, and \textsc{PourWater}, which involve explicit functional parts, the corresponding task-prior grasp-selection visualizations are provided in Appendix~\ref{app:som_real_grasp_selection}.
There, the SoM-guided process shows how VLM grounding identifies task-relevant object regions, functional parts, and preferred grasp areas, and how these semantic constraints are converted into grasp-candidate filtering or scoring rules.

\subsubsection{Orange-to-tray task}
\label{app:real_orange_task}

In the orange-to-tray task, the object and target container are relatively large, making them easier to localize from the base and wrist cameras.
Nevertheless, the original $\pi_{0.5}$ baseline achieves only about 10\% success, while GTP-FA-$\pi_{0.5}$ reaches about 92\%.
This indicates that the main difficulty is not only visual recognition, but also whether the robot can establish a stable and task-compatible grasp before downstream execution.

\begin{figure*}[h]
    \centering
    \setlength{\tabcolsep}{1pt}
    \renewcommand{\arraystretch}{0}
    \begin{tabular}{@{}cccccccc@{}}
        \includegraphics[width=0.122\linewidth]{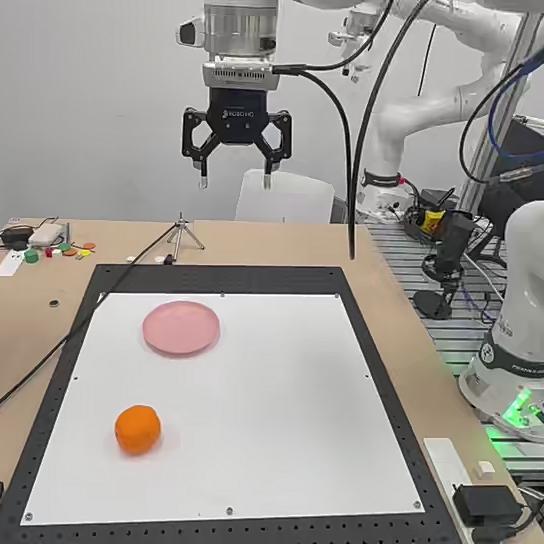} &
        \includegraphics[width=0.122\linewidth]{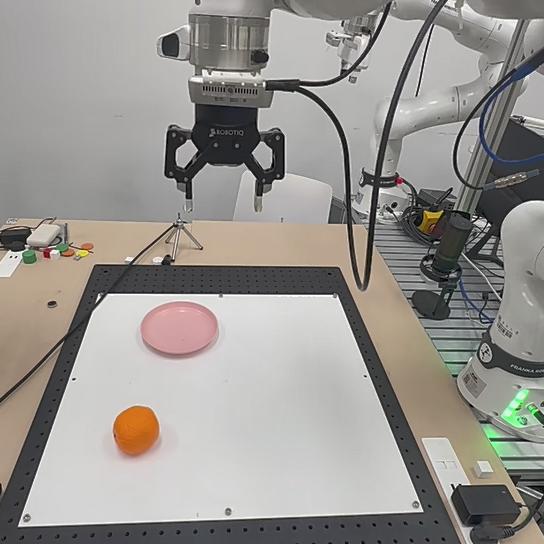} &
        \includegraphics[width=0.122\linewidth]{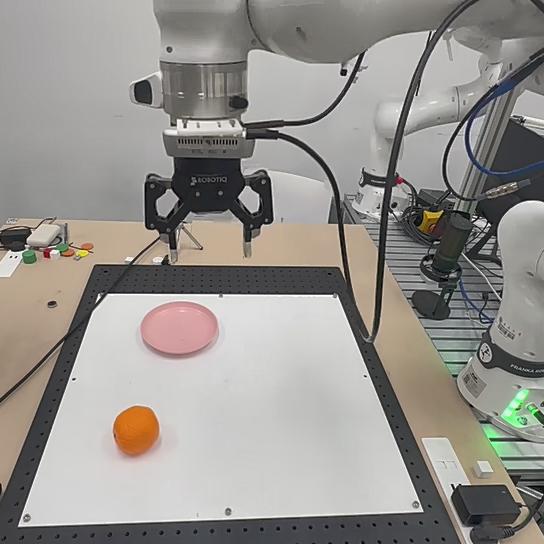} &
        \includegraphics[width=0.122\linewidth]{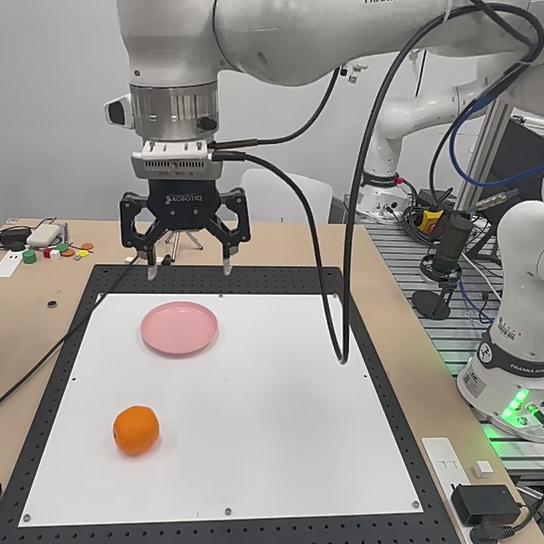} &
        \includegraphics[width=0.122\linewidth]{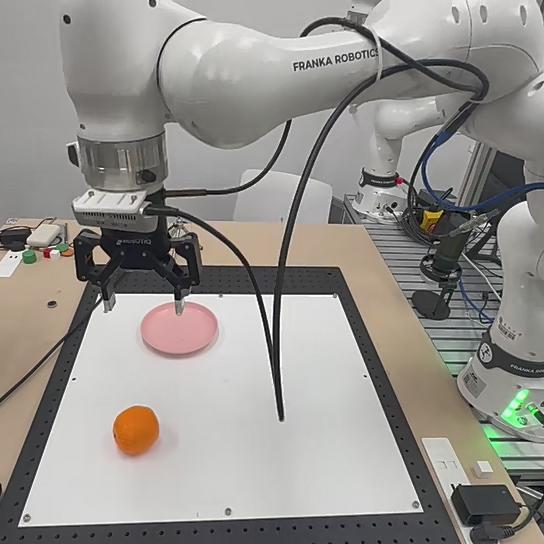} &
        \includegraphics[width=0.122\linewidth]{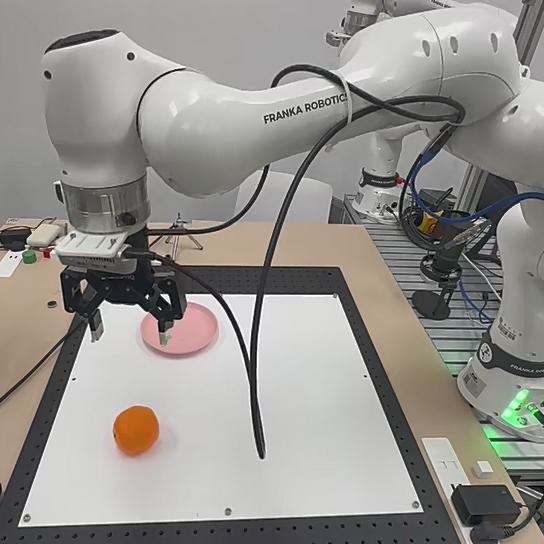} &
        \includegraphics[width=0.122\linewidth]{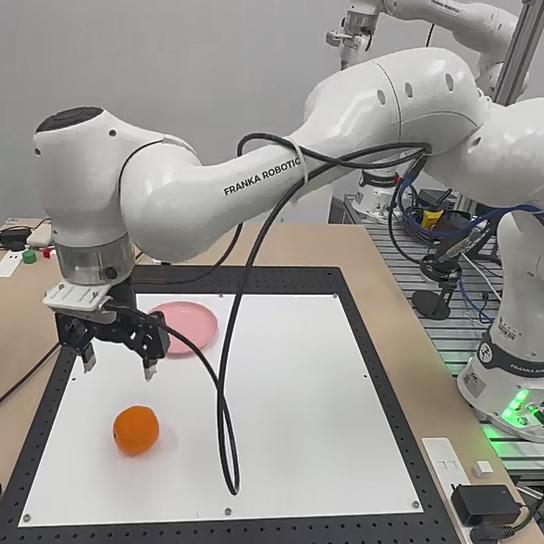} &
        \includegraphics[width=0.122\linewidth]{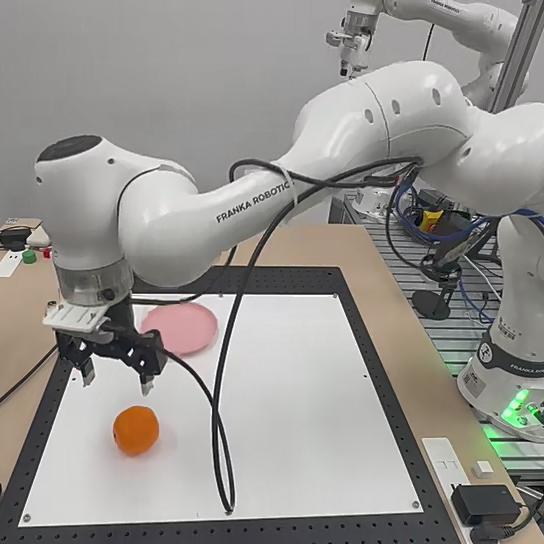} \\
        \includegraphics[width=0.122\linewidth]{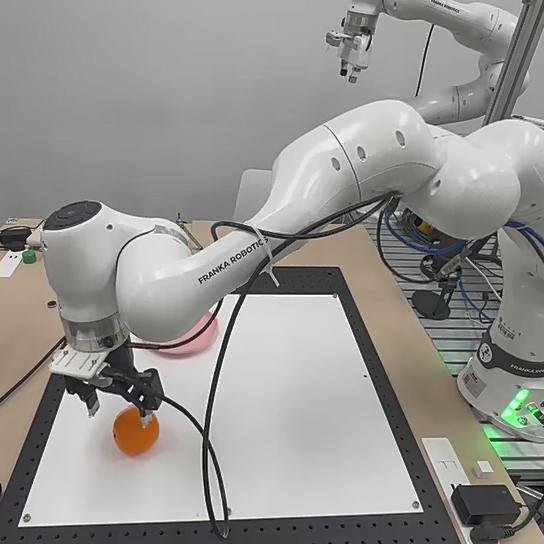} &
        \includegraphics[width=0.122\linewidth]{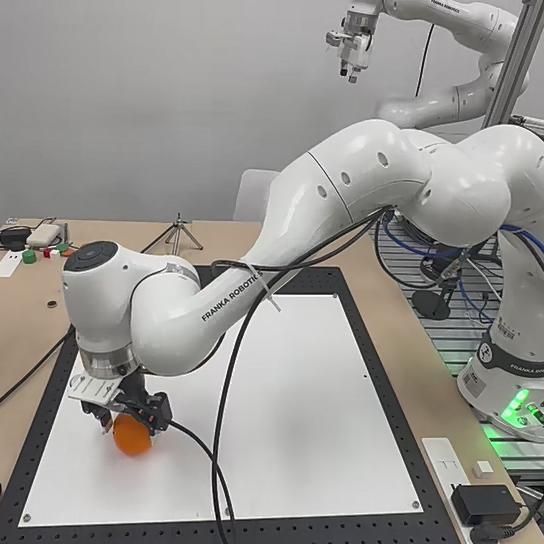} &
        \includegraphics[width=0.122\linewidth]{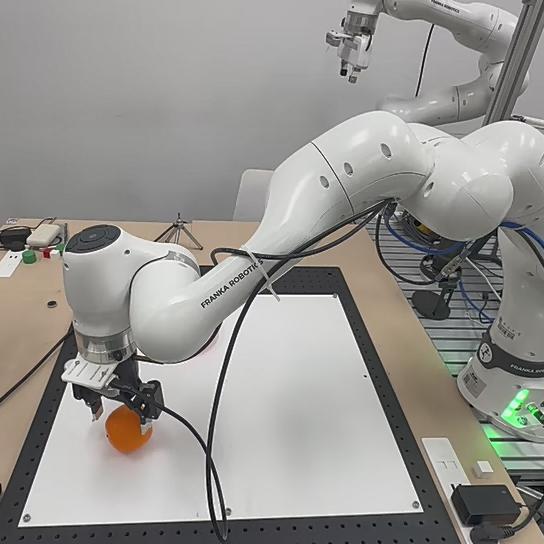} &
        \includegraphics[width=0.122\linewidth]{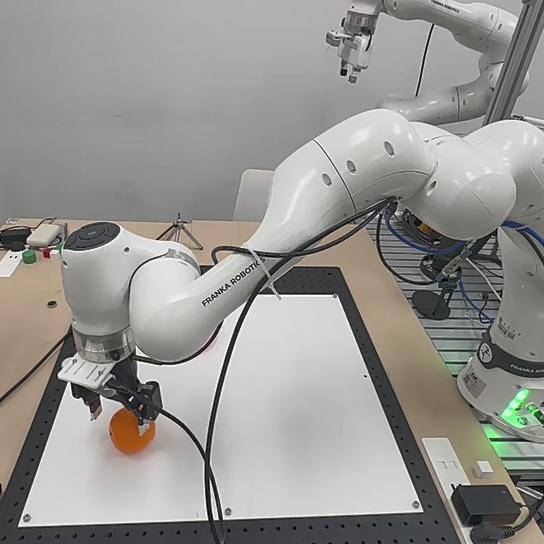} &
        \includegraphics[width=0.122\linewidth]{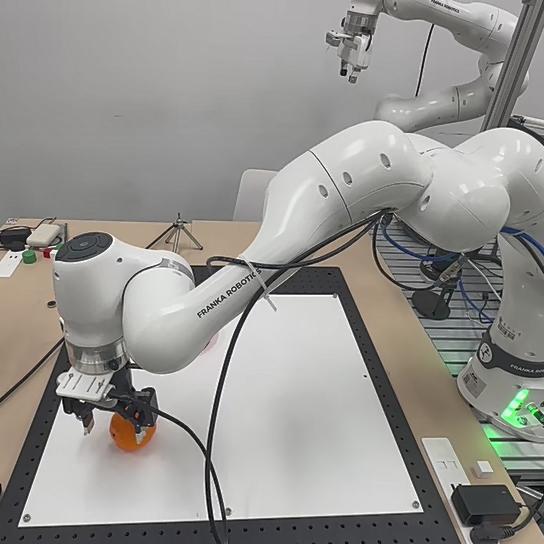} &
        \includegraphics[width=0.122\linewidth]{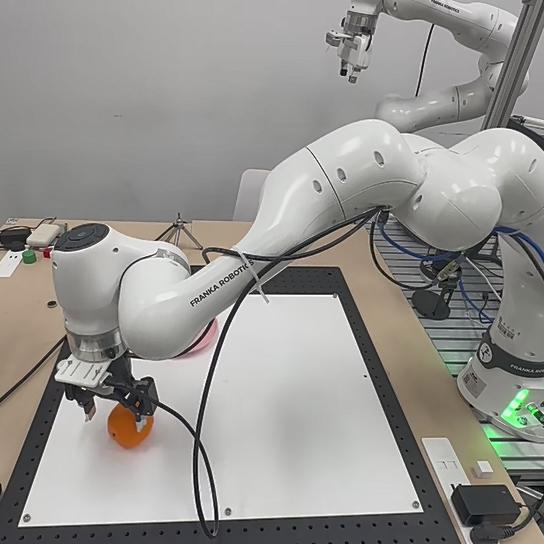} &
        \includegraphics[width=0.122\linewidth]{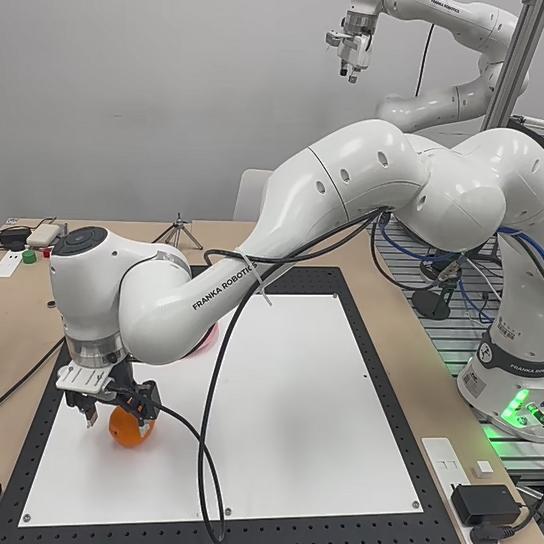} &
        \includegraphics[width=0.122\linewidth]{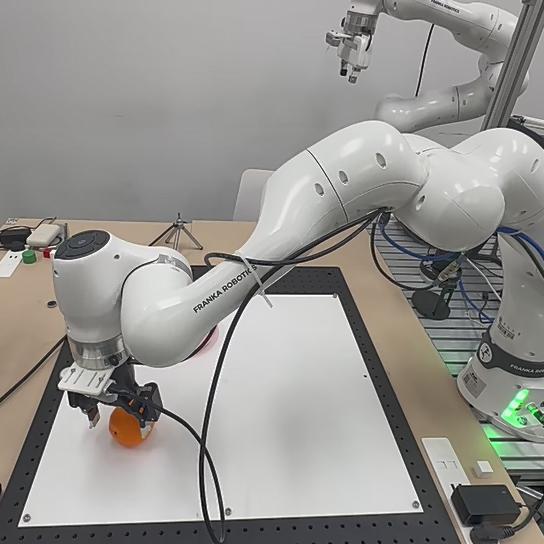}
    \end{tabular}
    \caption{
    Representative failure rollout of the original $\pi_{0.5}$ baseline on the orange-to-tray task.
    Frames are ordered chronologically from left to right and top to bottom.
    The baseline can produce partially reasonable motions, but without explicit grasp-conditioned diagnosis and grasp selection, the execution is more sensitive to grasp instability and post-grasp distribution shift, leading to task failure.
    }
    \label{fig:real_failure_sequence}
\end{figure*}

The original $\pi_{0.5}$ baseline often fails during the grasping stage.
Although the policy receives the correct task prompt and can observe both the orange and the tray, it frequently remains in repeated approach or grasp-attempt behaviors, or produces a grasp that does not support stable transport and release.
In our real-robot setup, the orange is also close to the maximum opening width of the Robotiq gripper, making grasping physically challenging: small approach-pose errors, gripper-closing errors, or contact offsets can prevent a secure closure or produce only marginal contact.
This further amplifies grasp instability for the original $\pi_{0.5}$ policy.
This behavior is illustrated by the failure interface in Figure~\ref{fig:real_interfaces_two_tasks}b and the frame sequence in Figure~\ref{fig:real_failure_sequence}.
Without explicit grasp-candidate filtering and grasp-conditioned diagnosis, the downstream action sequence is highly sensitive to grasp pose errors and post-grasp distribution shift.

\begin{figure*}[b]
    \centering
    \setlength{\tabcolsep}{1pt}
    \renewcommand{\arraystretch}{0}
    \begin{tabular}{@{}cccccccc@{}}
    \includegraphics[width=0.122\linewidth]{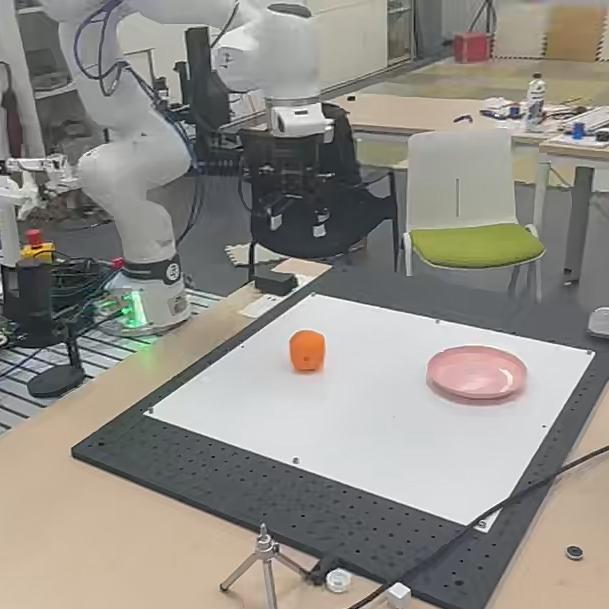} &
    \includegraphics[width=0.122\linewidth]{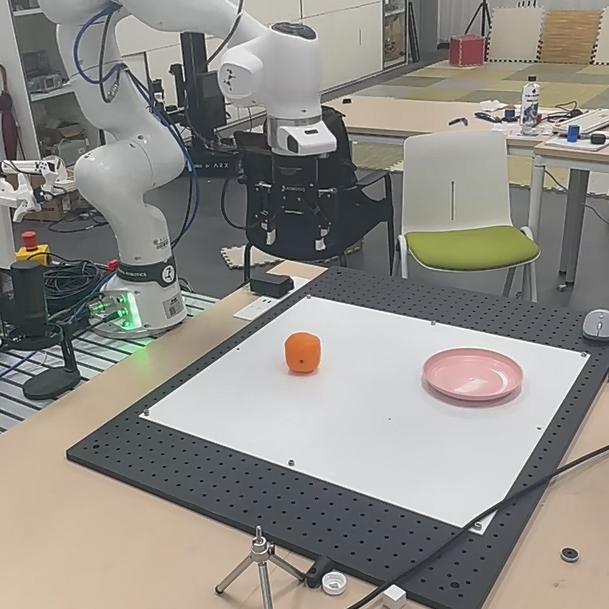} &
    \includegraphics[width=0.122\linewidth]{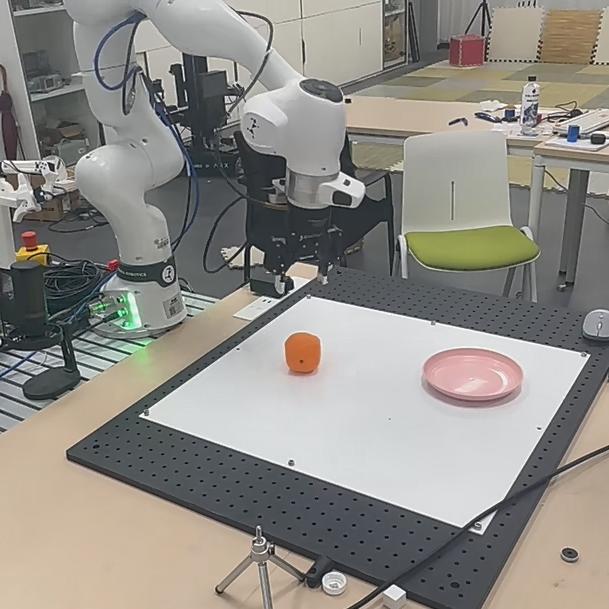} &
    \includegraphics[width=0.122\linewidth]{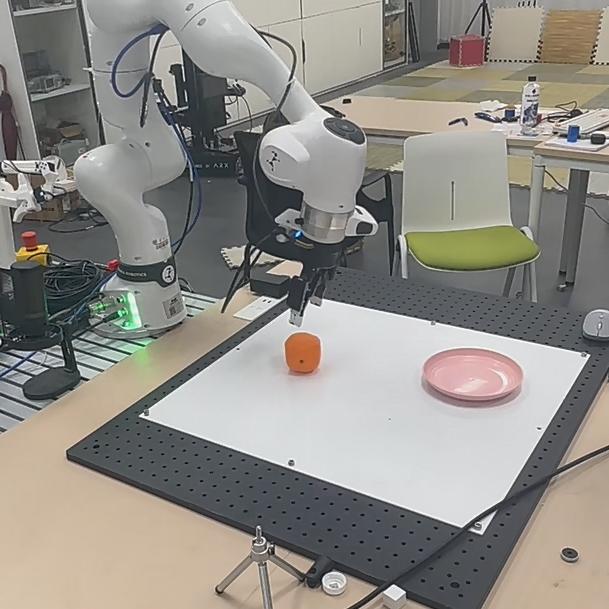} &
    \includegraphics[width=0.122\linewidth]{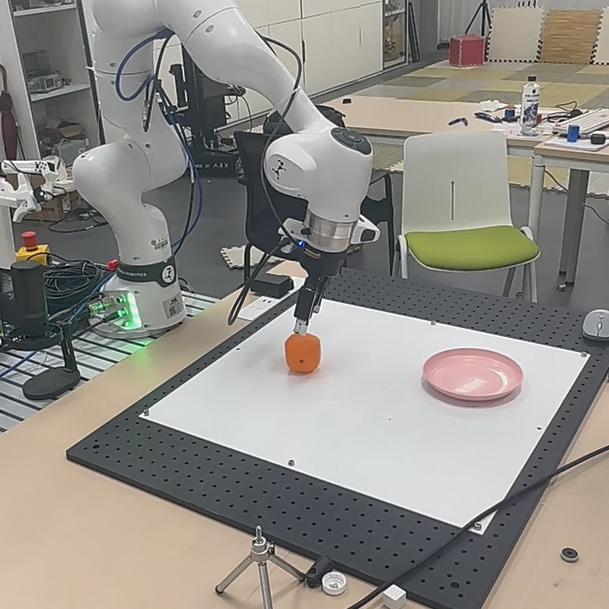} &
    \includegraphics[width=0.122\linewidth]{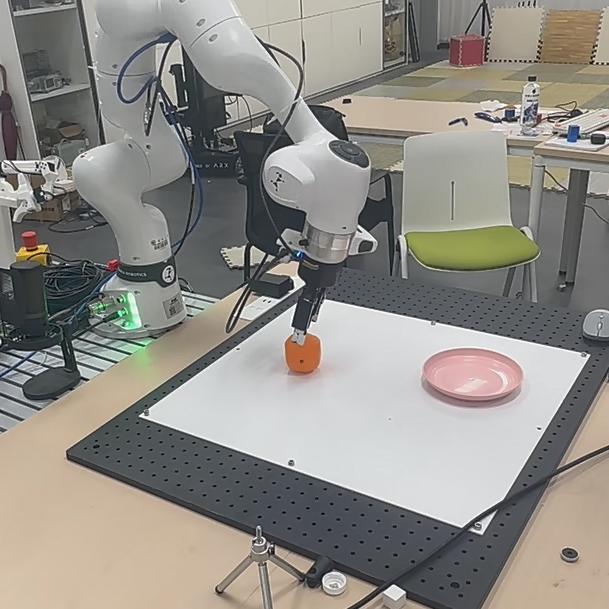} &
    \includegraphics[width=0.122\linewidth]{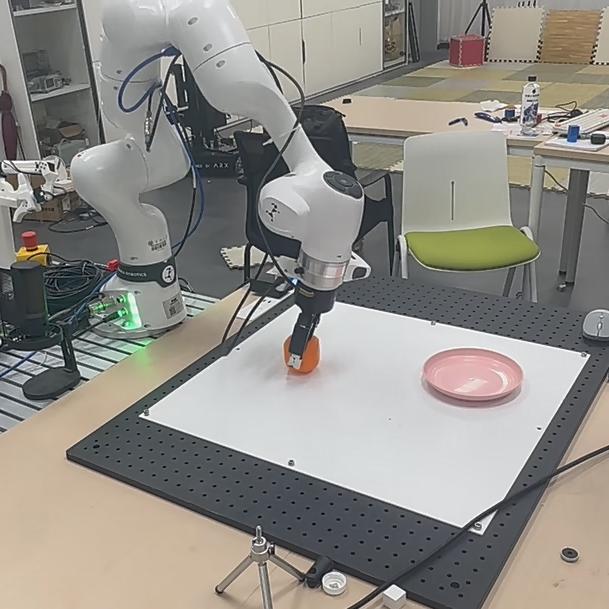} &
    \includegraphics[width=0.122\linewidth]{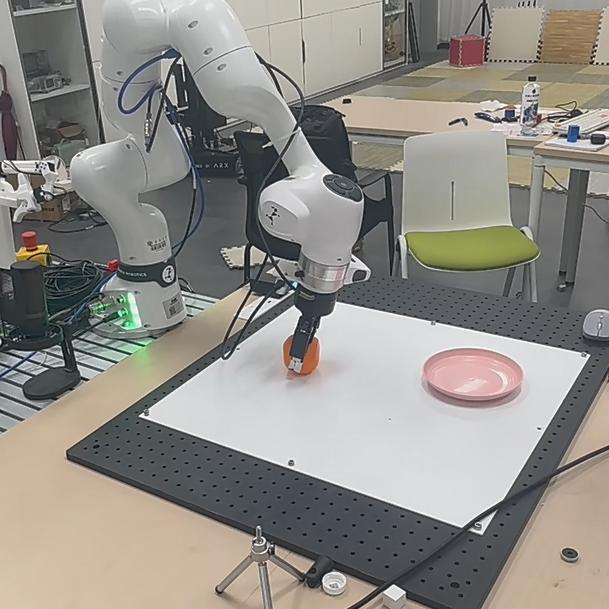} \\
    \includegraphics[width=0.122\linewidth]{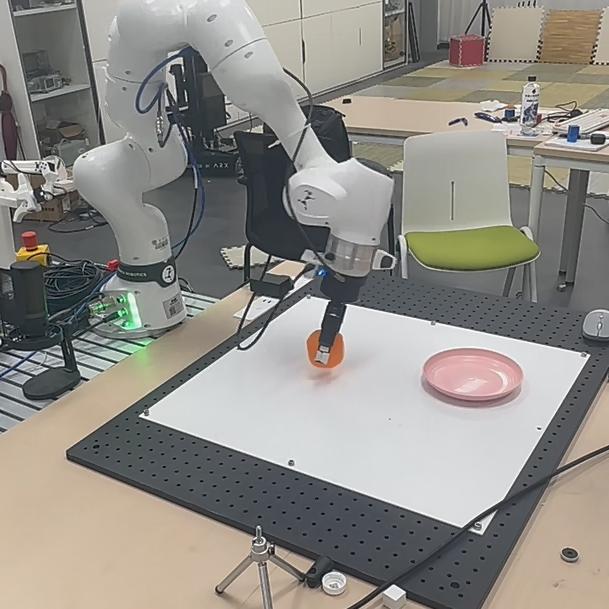} &
    \includegraphics[width=0.122\linewidth]{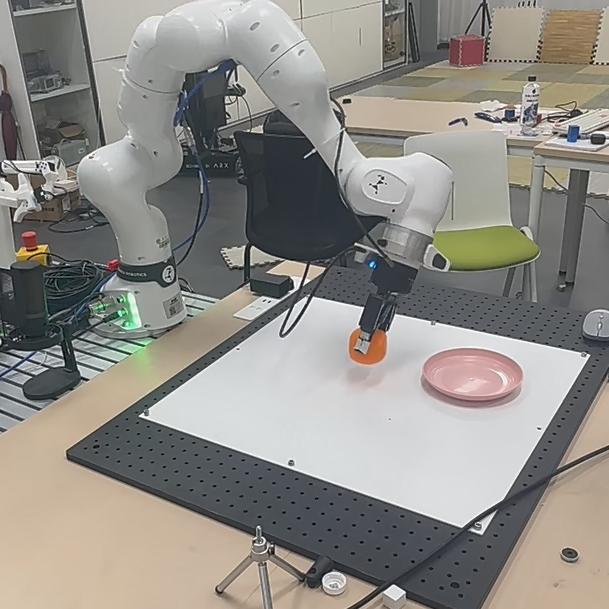} &
    \includegraphics[width=0.122\linewidth]{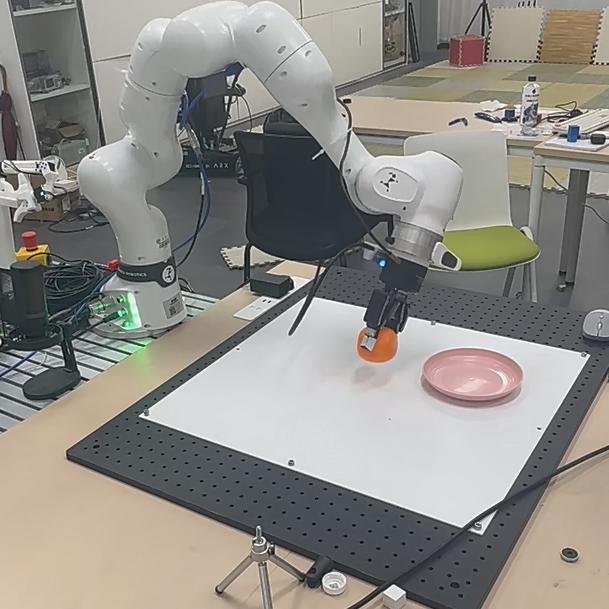} &
    \includegraphics[width=0.122\linewidth]{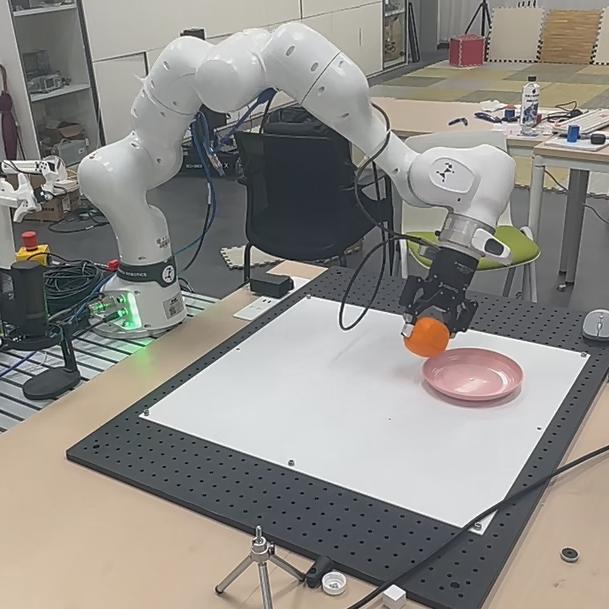} &
    \includegraphics[width=0.122\linewidth]{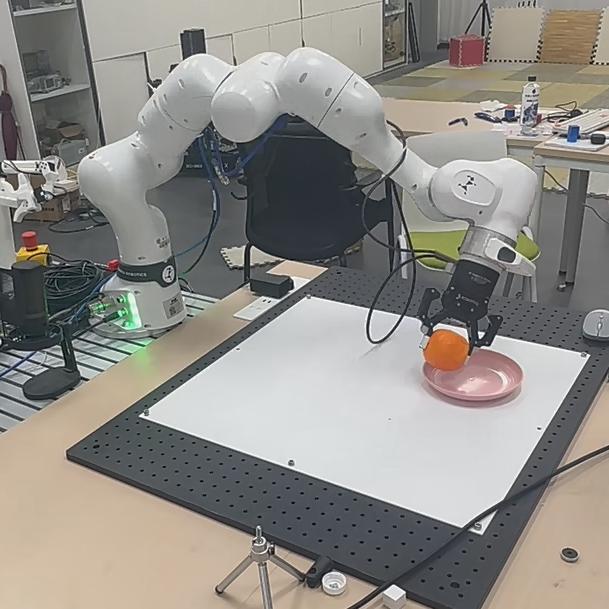} &
    \includegraphics[width=0.122\linewidth]{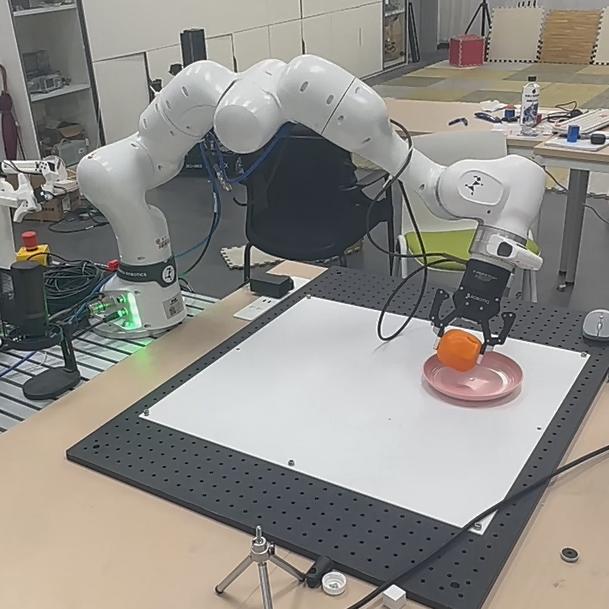} &
    \includegraphics[width=0.122\linewidth]{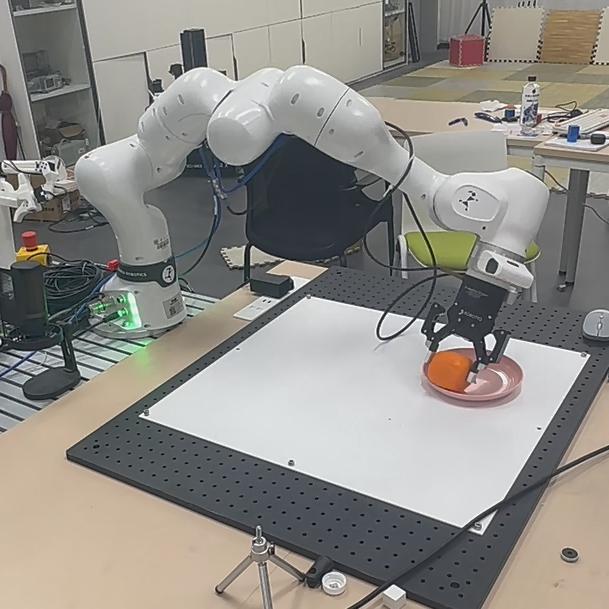} &
    \includegraphics[width=0.122\linewidth]{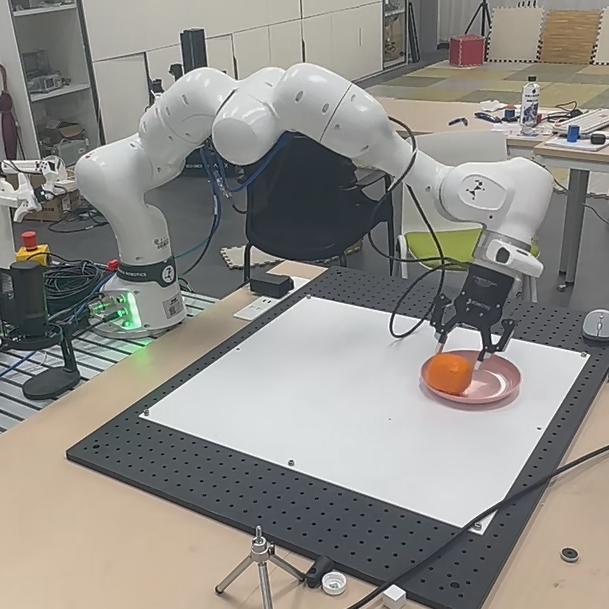}
    \end{tabular}
    \caption{
    Representative successful real-robot rollout of GTP-FA-$\pi_{0.5}$ on the orange-to-tray task.
    Frames are ordered chronologically from left to right and top to bottom.
    The robot approaches the orange, establishes a stable grasp, transports the object above the tray, releases it, and completes the task successfully.
    }
    \label{fig:real_success_sequence}
\end{figure*}

In contrast, GTP-FA-$\pi_{0.5}$ first selects a grasp that is compatible with the subsequent placement stage.
The selected grasp provides a more reliable post-grasp starting condition for the downstream $\pi_{0.5}$ policy, reducing object slip, unstable lifting, and release failure.
As shown in Figure~\ref{fig:real_success_sequence}, the robot approaches the orange, establishes a stable grasp, lifts the object, transports it above the tray, releases it, and completes the task successfully.

\subsubsection{Cube-stacking task}
\label{app:real_stack_task}

We further evaluate the real-robot cube-stacking task, where the robot must pick up the blue cube and stack it on the orange cube.
This task is more sensitive to execution precision than orange-to-tray, because the robot must localize small objects, grasp the blue cube with a suitable pose, align it above the orange cube, and release it without disturbing the stacked configuration.
In this task, the original $\pi_{0.5}$ baseline achieves about 4\% success, whereas GTP-FA-$\pi_{0.5}$ achieves about 74\%.

A key challenge in the real setup is that the base camera is relatively far from the workspace.
As a result, the blue and orange cubes occupy only a small region in the base-camera image.
The original $\pi_{0.5}$ baseline often fails to reliably localize the blue cube or to maintain accurate spatial alignment during grasping and placement.
As shown in the failure interface in Figure~\ref{fig:real_interface_stack_vla}, even when the task prompt is correct, the policy may produce inaccurate approach motions, incorrect grasping behavior, or placement errors.

\begin{figure*}[t]
    \centering
    \setlength{\tabcolsep}{1pt}
    \renewcommand{\arraystretch}{0.8}
    \begin{tabular}{ccccccccc}
        \includegraphics[width=0.105\linewidth]{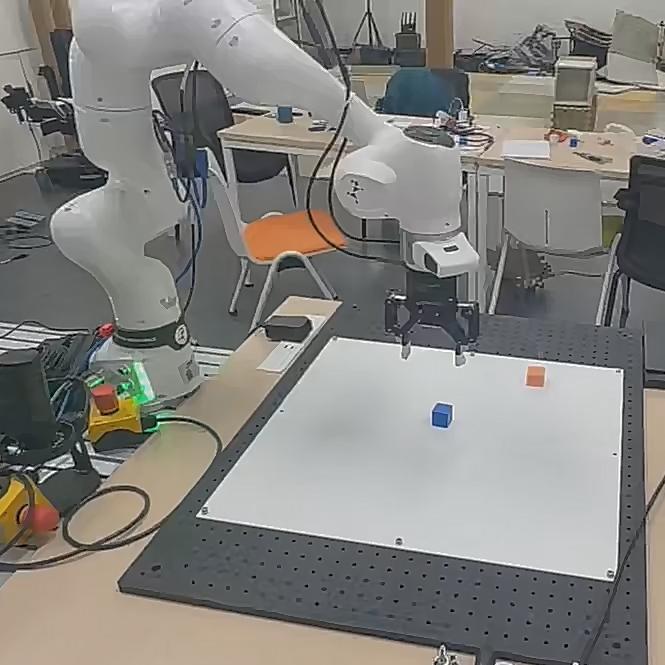} &
        \includegraphics[width=0.105\linewidth]{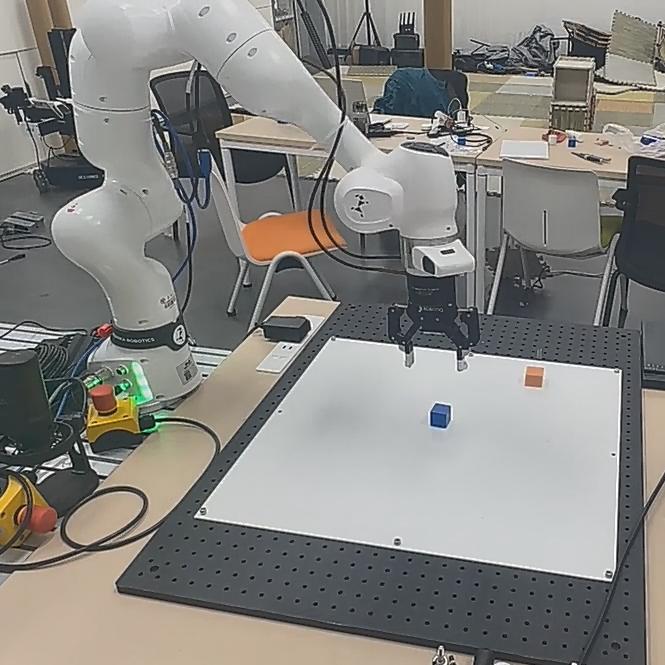} &
        \includegraphics[width=0.105\linewidth]{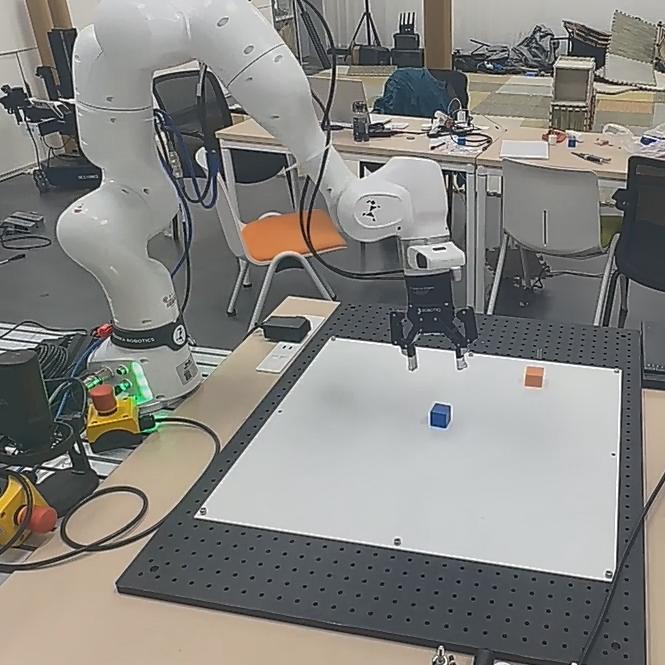} &
        \includegraphics[width=0.105\linewidth]{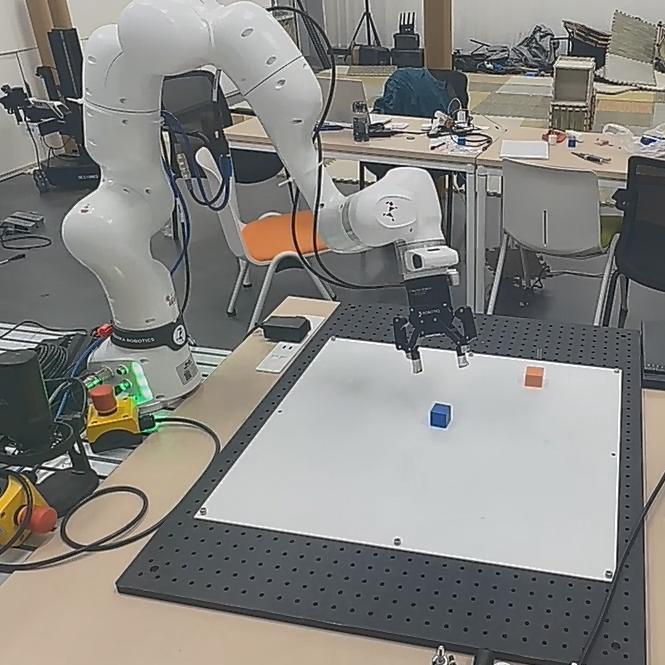} &
        \includegraphics[width=0.105\linewidth]{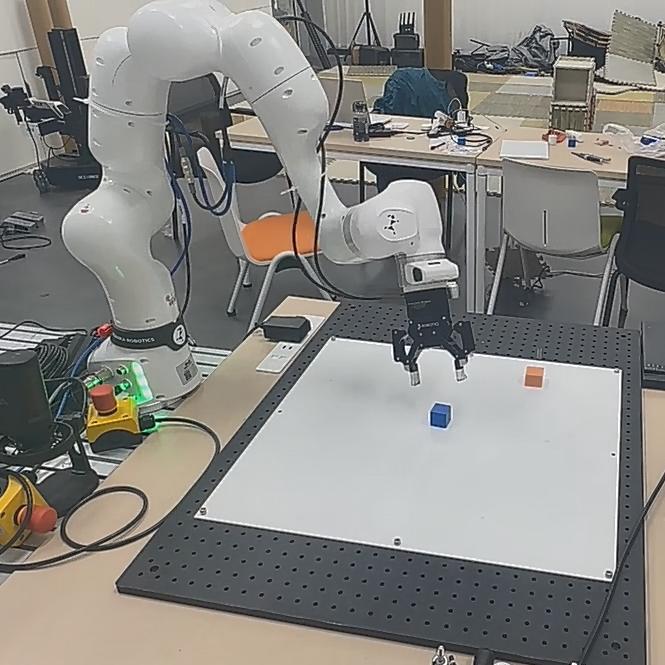} &
        \includegraphics[width=0.105\linewidth]{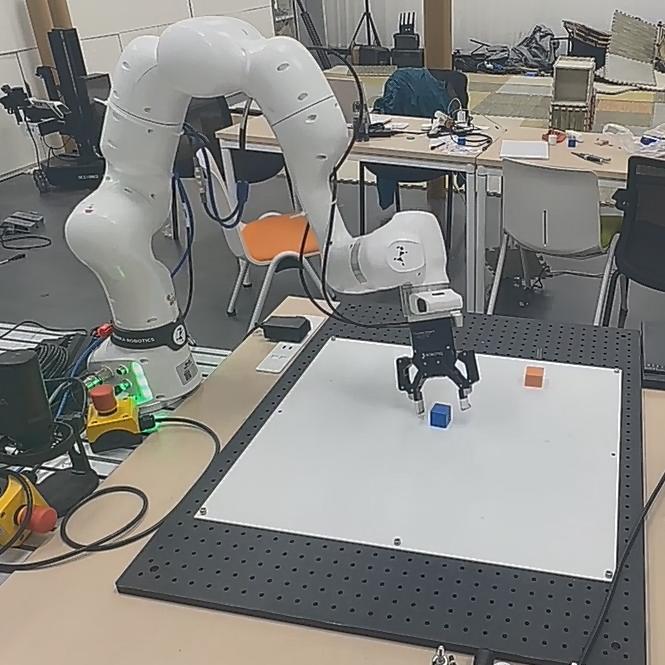} &
        \includegraphics[width=0.105\linewidth]{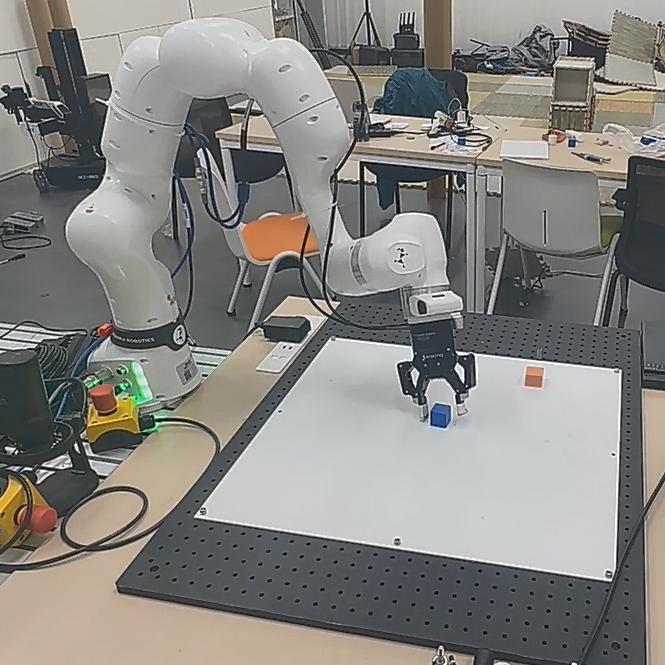} &
        \includegraphics[width=0.105\linewidth]{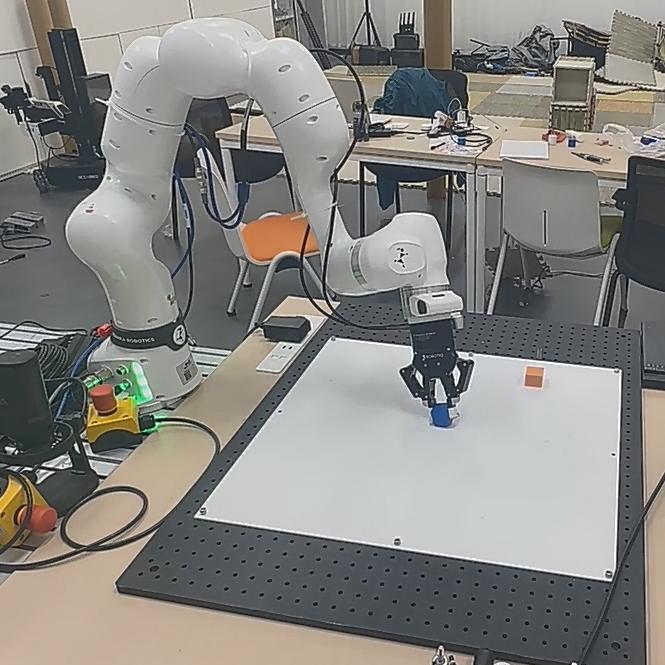} &
        \includegraphics[width=0.105\linewidth]{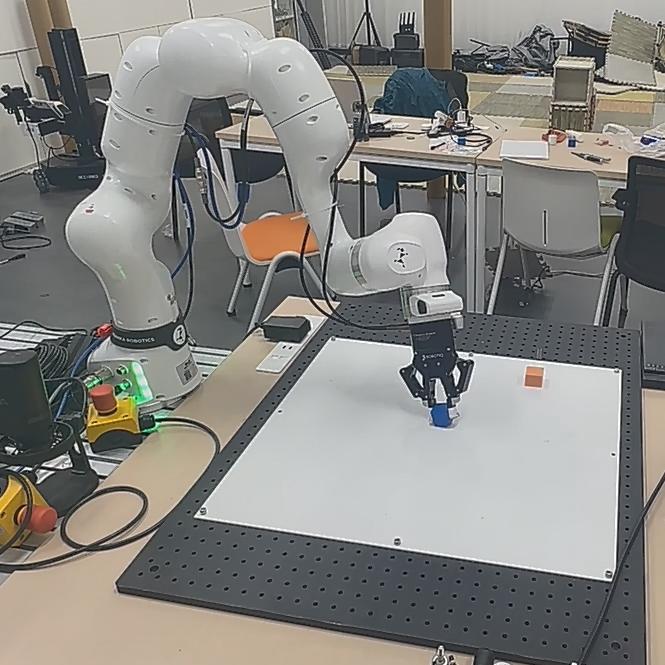} \\
        \includegraphics[width=0.105\linewidth]{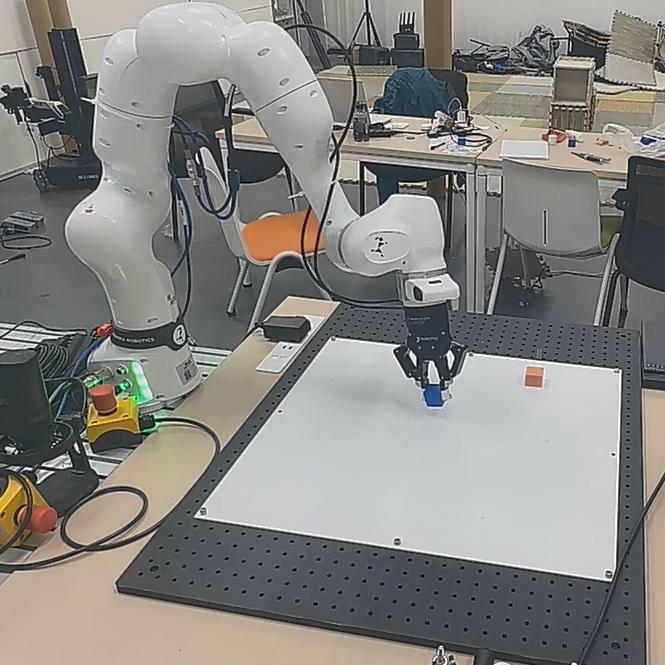} &
        \includegraphics[width=0.105\linewidth]{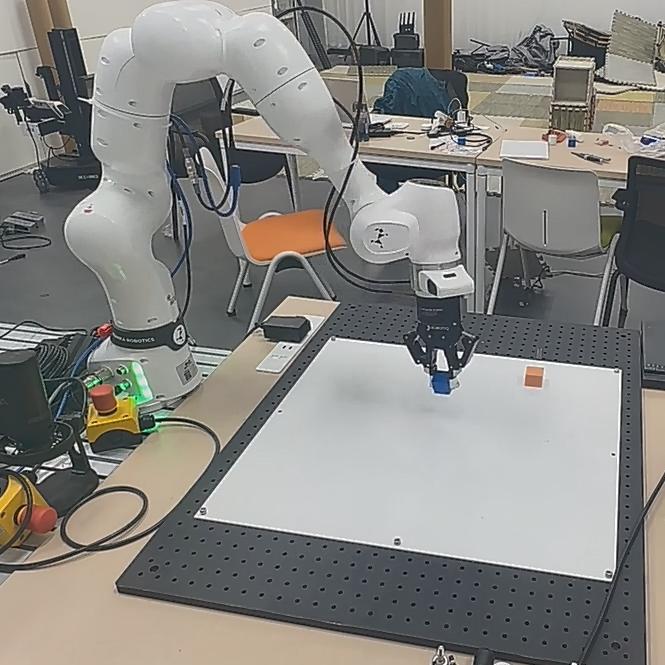} &
        \includegraphics[width=0.105\linewidth]{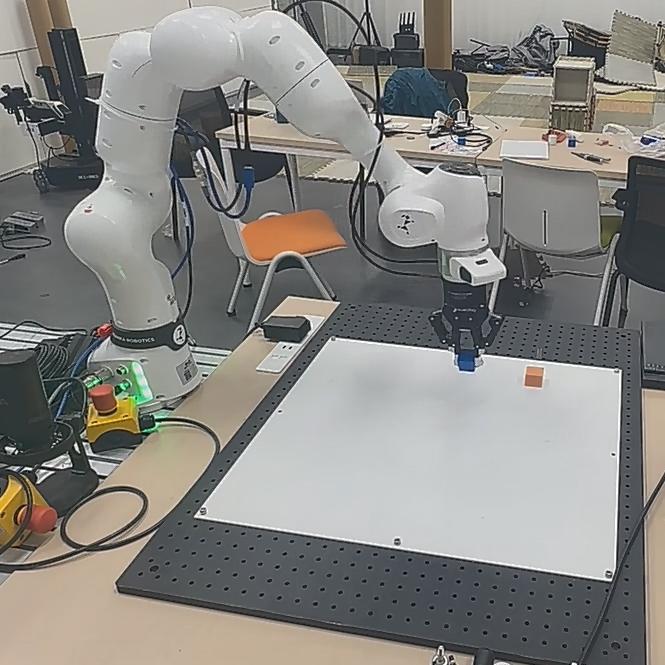} &
        \includegraphics[width=0.105\linewidth]{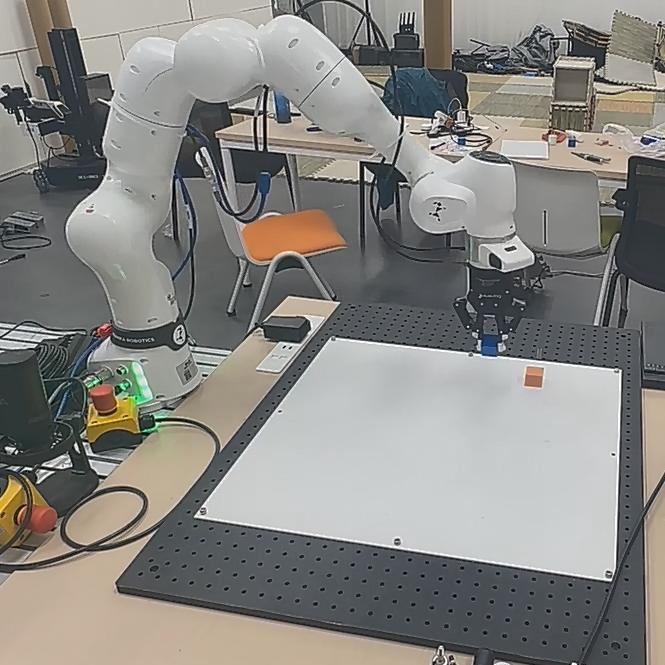} &
        \includegraphics[width=0.105\linewidth]{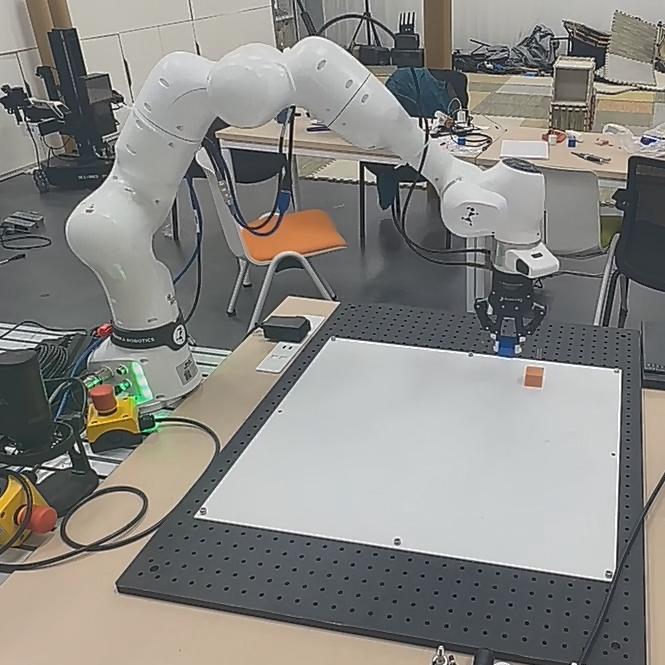} &
        \includegraphics[width=0.105\linewidth]{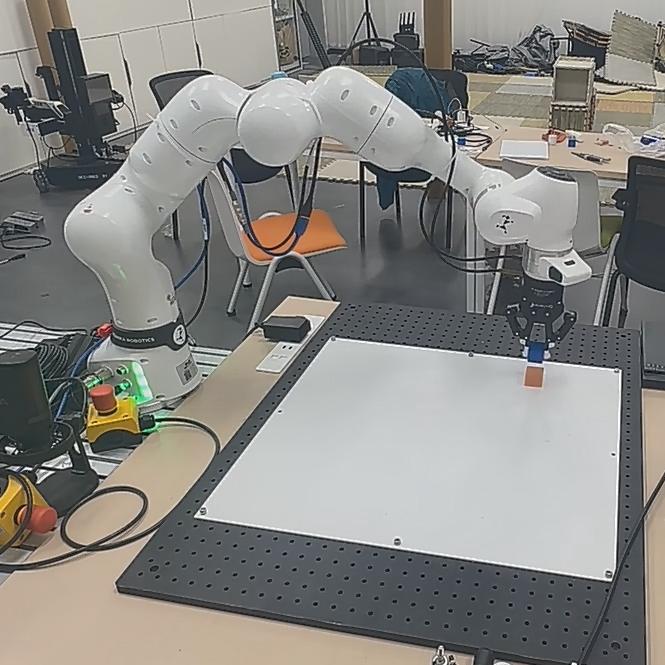} &
        \includegraphics[width=0.105\linewidth]{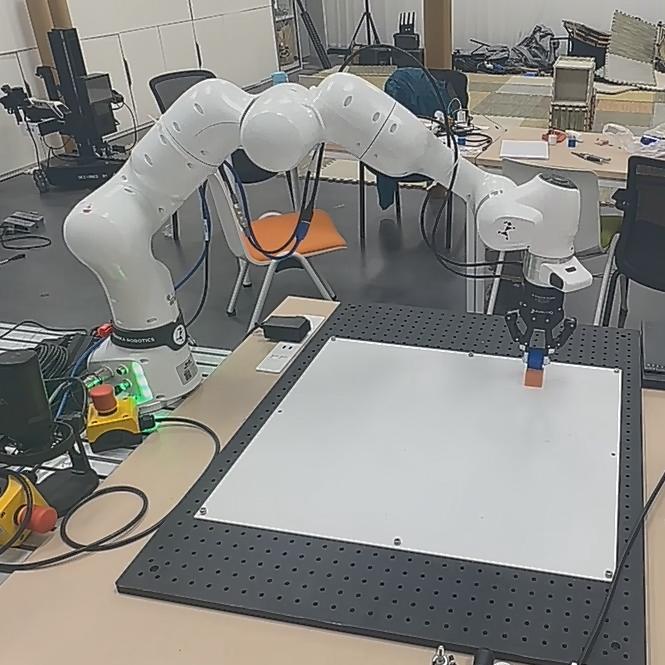} &
        \includegraphics[width=0.105\linewidth]{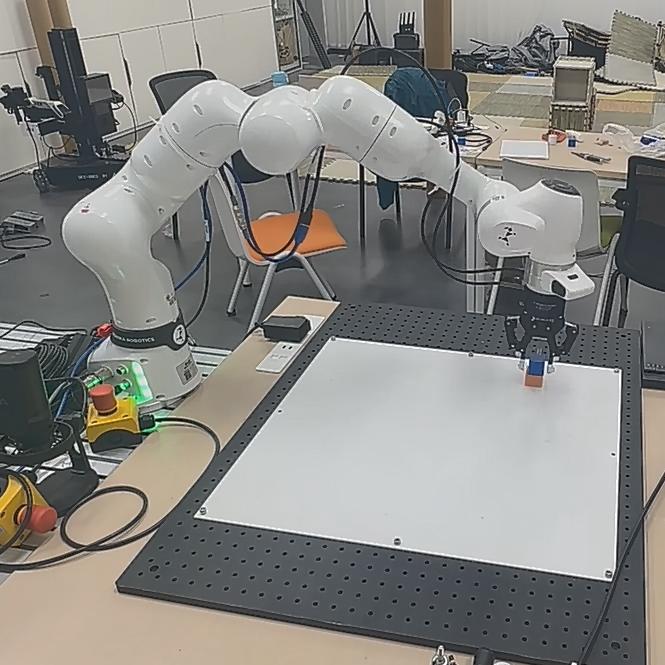} &
        \includegraphics[width=0.105\linewidth]{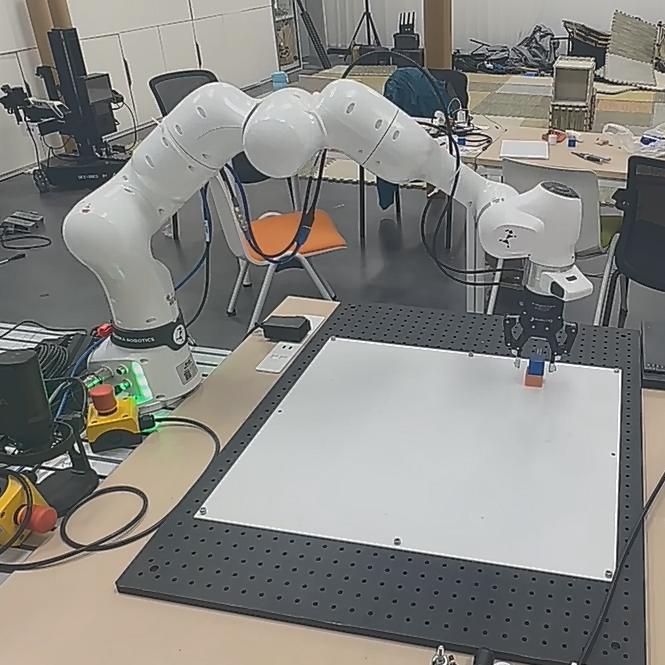}
    \end{tabular}
    \caption{
    Representative successful real-robot rollout of GTP-FA-$\pi_{0.5}$ on the cube-stacking task.
    We select 18 frames from the success video and order them chronologically according to their original frame indices.
    The robot approaches the blue cube, establishes a stable grasp, lifts it, moves above the orange cube, aligns the object, releases it, and completes a stable stack.
    }
    \label{fig:stack_success_rollout}
\end{figure*}

\begin{figure*}[htp!]
    \centering
    \setlength{\tabcolsep}{1pt}
    \renewcommand{\arraystretch}{0.8}
    \begin{tabular}{ccccccccc}
        \includegraphics[width=0.105\linewidth]{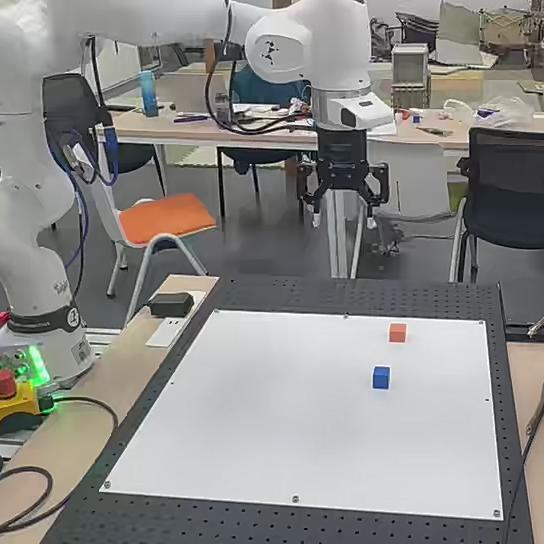} &
        \includegraphics[width=0.105\linewidth]{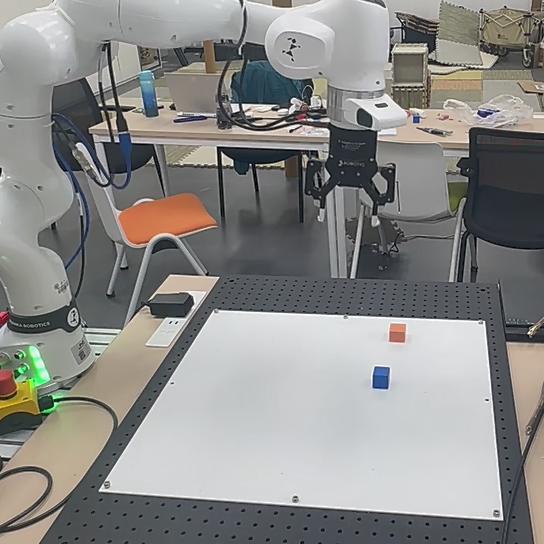} &
        \includegraphics[width=0.105\linewidth]{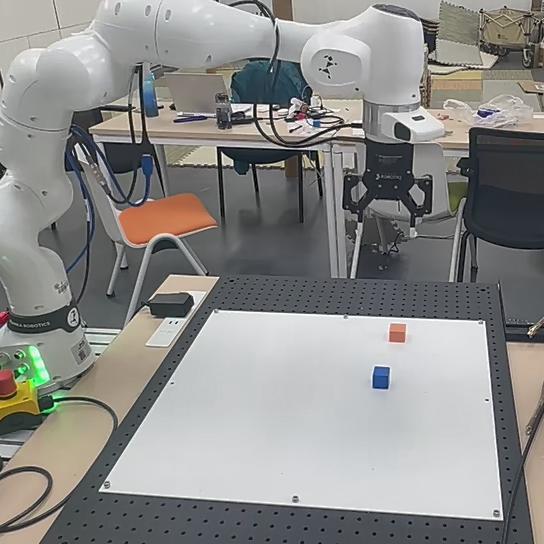} &
        \includegraphics[width=0.105\linewidth]{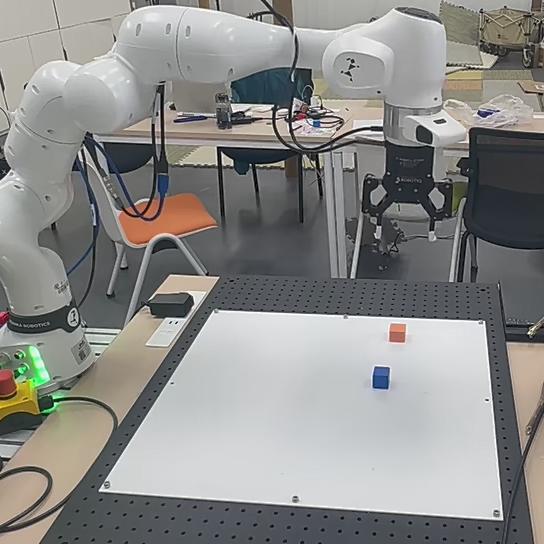} &
        \includegraphics[width=0.105\linewidth]{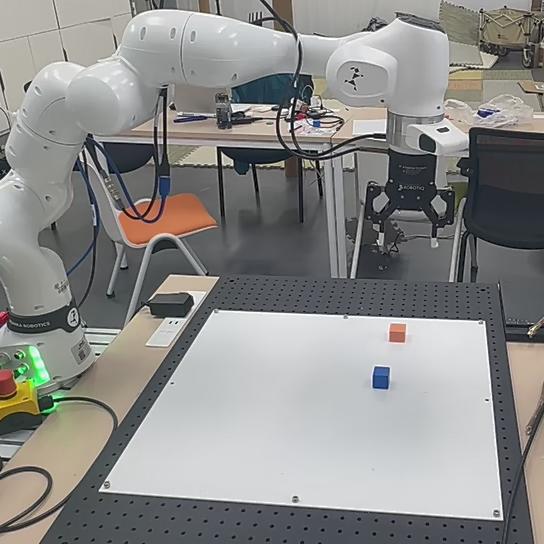} &
        \includegraphics[width=0.105\linewidth]{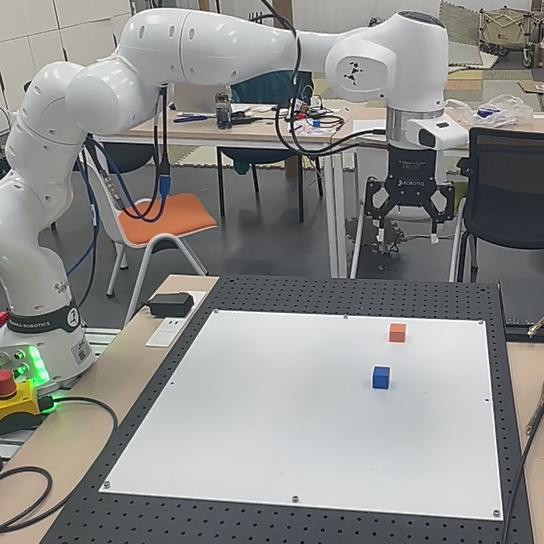} &
        \includegraphics[width=0.105\linewidth]{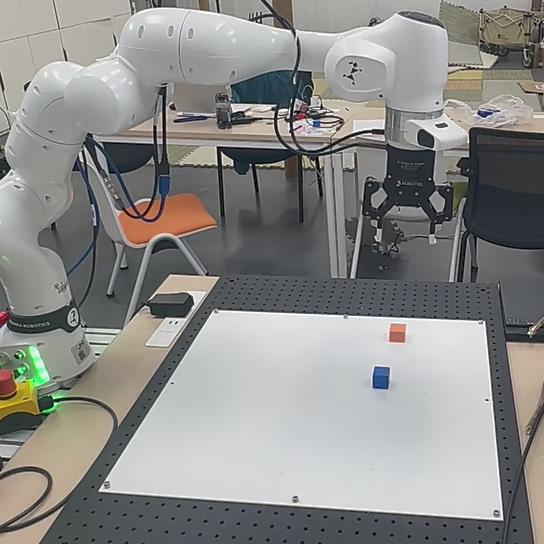} &
        \includegraphics[width=0.105\linewidth]{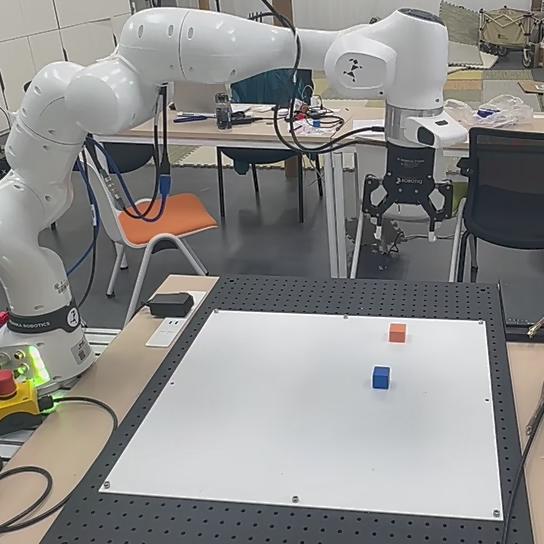} &
        \includegraphics[width=0.105\linewidth]{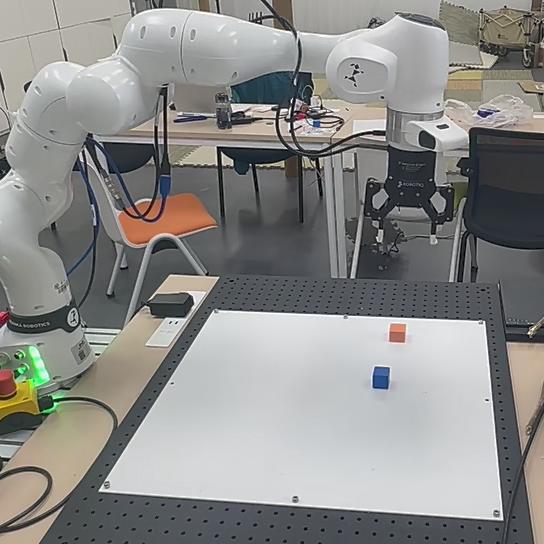}
    \end{tabular}
    \caption{
    Representative failure rollout of the original $\pi_{0.5}$ baseline on the cube-stacking task.
    We select 9 frames from the failure video and order them chronologically according to their original frame indices.
    Due to small-object localization difficulty and the absence of explicit grasp-conditioned diagnosis, the baseline is more likely to produce inaccurate approach, grasping, or placement behavior, causing the stacking task to fail.
    }
    \label{fig:stack_failure_rollout}
\end{figure*}

GTP-FA-$\pi_{0.5}$ mitigates this issue by explicitly grounding the execution on a selected grasp.
The grasp-selection stage first converts raw GraspNet candidates into a unique task-compatible grasp on the blue cube.
This grasp not only supports stable lifting, but also induces a post-grasp cube pose that is more favorable for the subsequent stacking stage.
The downstream $\pi_{0.5}$ policy then starts from a more reliable grasp-conditioned state, improving the chance of successful alignment and placement.
Representative success and failure rollout sequences for this task are shown in Figures~\ref{fig:stack_success_rollout} and~\ref{fig:stack_failure_rollout}, respectively.

\subsubsection{Tool-use and pouring task observations}
\label{app:real_tool_pouring_tasks}

Beyond orange-to-tray and cube stacking, we further evaluate three real-robot tasks that rely more heavily on functional-part selection: \textsc{PokeCube}, \textsc{PullCubeTool}, and \textsc{PourWater}.
The quantitative results are reported in Table~\ref{tab:real_robot_protocol} in the main text.
The original $\pi_{0.5}$ baseline achieves about 24\%, 16\%, and 2\% success on \textsc{PokeCube}, \textsc{PullCubeTool}, and \textsc{PourWater}, respectively, while GTP-FA-$\pi_{0.5}$ improves the success rates to about 86\%, 78\%, and 54\%.
These results further show that real-robot success depends not only on whether $\pi_{0.5}$ understands the language goal, but also on whether the system selects the correct functional grasp region.

In \textsc{PokeCube}, the robot needs to grasp the red end of the stick and use the stick to push the yellow cube into the red target area.
The grasp must allow stable control of the stick while preserving an effective contact geometry for pushing.
If the grasp is unstable or induces an unfavorable stick pose, the downstream policy can fail during the contact-rich pushing phase even if it predicts plausible motions.
GTP-FA restricts grasp selection to regions that better support pushing control through task priors and candidate-grasp filtering, improving the stability of the subsequent tool-use behavior.

In \textsc{PullCubeTool}, the robot needs to grasp the red part of the hook and pull the yellow cube into the red target area.
This task is highly sensitive to the functional end of the tool.
If the grasp occupies the hook tip, the curved contact region, or a region that interferes with the hook-cube interaction, the tool may fail to engage or pull the cube effectively.
GTP-FA preserves the functional interaction region of the hook while selecting a grasp that supports stable tool control, thereby reducing failures caused by functional-region occlusion.

In \textsc{PourWater}, the robot needs to grasp the red handle of the gray cup and pour its contents into the blue-gray cup.
This task is more difficult than simple pushing or pulling because it requires not only stable grasping, but also a post-grasp cup pose that supports transport, alignment, and tilting.
The original $\pi_{0.5}$ baseline has a very low success rate because direct visual-language action prediction can easily produce errors in handle localization, grasp stability, or pouring pose.
GTP-FA explicitly identifies the red handle as the task-relevant grasp region and suppresses unstable or pouring-unfriendly grasps through diagnostic risk calibration, providing a more reliable cup pose for downstream pouring.

These tasks show that the grasp region itself carries task semantics for tool use and container manipulation.
The task prior helps preserve grasps that support subsequent functional interaction and suppresses grasps that occupy functional parts, disrupt tool-object contact, or hinder pouring motions.
The SoM-guided task-prior grasp selection process for these tasks is visualized in Appendix~\ref{app:som_real_grasp_selection}, and additional real-robot videos are provided on the project page.

\subsubsection{Summary and supplementary videos}
\label{app:real_summary}

The real-world visualizations support four observations.
First, explicit grasp selection provides a more stable and task-compatible post-grasp state for downstream $\pi_{0.5}$ execution.
In the orange-to-tray task, although the orange and tray are relatively large and visually observable, the original $\pi_{0.5}$ baseline still often fails during grasping.
This shows that successful task execution requires more than visual-language understanding; it also requires a physically stable grasp, especially when the object size is close to the gripper opening limit.

Second, the cube-stacking task exposes a stronger precision bottleneck.
The target cubes occupy only a small portion of the base-camera view, and the final placement requires accurate alignment.
The original $\pi_{0.5}$ baseline is therefore prone to target-localization, grasp-pose, and placement errors.
By contrast, GTP-FA-$\pi_{0.5}$ uses task-aware grasp selection and diagnosis-driven optimization to produce a more reliable grasp-conditioned starting point for stacking.

Third, \textsc{PokeCube}, \textsc{PullCubeTool}, and \textsc{PourWater} show that real-robot manipulation often requires functional-part-level grasp selection.
For sticks, hooks, and cups, if the grasp occupies the tool interaction end, disrupts contact geometry, or induces a pose unsuitable for pouring, the downstream policy can hardly compensate for the early grasp error through action prediction alone.
GTP-FA combines VLM task priors, GraspNet candidates, and diagnostic risk calibration to select grasps that are more consistent with both task semantics and downstream execution requirements.

Fourth, the comparison between raw GraspNet candidates and the selected unique execution grasp shows that GTP-FA does not merely add a grasp detector before $\pi_{0.5}$ execution.
Instead, it explicitly filters and calibrates grasp candidates according to task compatibility and diagnostic risk, thereby reducing grasp-induced downstream failures.
Together, these results show that $\pi_{0.5}$ policies can benefit substantially from an explicit grasp-then-plan interface with failure attribution, especially in real-world long-horizon manipulation where contact errors, visual scale changes, small-object localization errors, and post-grasp distribution shifts are common.

More real-robot videos are provided on the project page:
\href{https://sites.google.com/view/gtp-fa/}{https://sites.google.com/view/gtp-fa/}.
The videos include successful GTP-FA-$\pi_{0.5}$ executions on all five real-world tasks and representative failure cases of the original $\pi_{0.5}$ baseline, further illustrating the behavioral differences between the two methods.

\subsection{Experimental Protocols and Key Hyperparameters}
\label{app:experimental_protocols}

This section provides additional details on the main training settings, key hyperparameters, computational resources, and real-robot evaluation protocol used in our experiments. Unless otherwise specified, all simulation experiments use the \texttt{pd\_joint\_delta\_pos} control mode and the PhysX CUDA backend. For PPO, SAC, BC, and DP, we run experiments with three random seeds. For VLA experiments, due to the higher cost of fine-tuning and inference, we report task-level terminal success under a fixed evaluation protocol. For VLA-based experiments, the simulated tasks use 100 converted expert trajectories for each fine-tuning setting. For the real-robot setting, both the original VLA baseline and GTP-FA-VLA use the same 300 real expert trajectories per task (1,500 across five tasks), controlling the demonstration budget in this system-level comparison.

\subsubsection{Key Hyperparameters for Original Policies, Ablations, and GTP-FA}
\label{app:key_hyperparameters}

The meanings of \texttt{00}/\texttt{01}/\texttt{10}/\texttt{11} follow the definitions in the main text: \texttt{00} denotes the original policy, \texttt{01} denotes planning-side-only optimization, \texttt{10} denotes grasp-side-only optimization, \texttt{11} denotes naive grasp--plan optimization without failure attribution, and GTP-FA denotes the full method with failure attribution and diagnostic routing. Table~\ref{tab:algo_hyperparams} summarizes the main hyperparameters used for the original policies, ablation variants, and GTP-FA variants.

\begin{table*}[t]
\centering
\caption{
Key hyperparameters for the original policies, ablation variants, and GTP-FA variants.
}
\label{tab:algo_hyperparams}
\small
\setlength{\tabcolsep}{4pt}
\renewcommand{\arraystretch}{1.18}
\begin{tabular}{p{0.08\linewidth} p{0.56\linewidth} p{0.29\linewidth}}
\toprule
\textbf{Method} & \textbf{Main Training Hyperparameters} & \textbf{Evaluation Setting} \\
\midrule

PPO &
Seeds = 0, 1, 2; number of environments = 2048; rollout steps = 16; update epochs = 8; minibatches = 32; total timesteps = 50M; learning rate = $3\times10^{-4}$; discount factor $\gamma=0.8$; GAE $\lambda=0.9$.
&
During training, we use 16 evaluation environments, 50 evaluation steps, and an evaluation frequency of 25. Final evaluation uses 8 parallel environments and 12,500 evaluation steps. \\

\midrule

SAC &
Seeds = 0, 1, 2; number of environments = 64; rollout steps = 50; total timesteps = 1M; replay buffer size = 1M; batch size = 1024; learning starts = 4000; policy learning rate = $3\times10^{-4}$; Q-function learning rate = $3\times10^{-4}$; $\gamma=0.8$; $\tau=0.01$; entropy coefficient $\alpha=0.2$ with automatic entropy tuning.
&
Training evaluation uses 50 episodes. Final evaluation uses 500 episodes. The evaluation frequency is 25. \\

\midrule

BC &
Seeds = 0, 1, 2; total training iterations = 5,000--10,000; batch size = 1024; learning rate = $3\times10^{-4}$; number of demonstrations = 200; maximum episode steps = 50; state normalization enabled.
&
Training evaluation uses 50--100 episodes with 8--16 evaluation environments. Final evaluation uses 500 episodes. \\

\midrule

DP &
Seeds = 0, 1, 2; total training iterations = 100,000; batch size = 1024; learning rate = $1\times10^{-4}$; number of demonstrations = 200; observation horizon = 2; action horizon = 8; prediction horizon = 16; diffusion-step embedding dimension = 64; U-Net dimensions = [64, 128, 256]; number of groups = 8.
&
Training evaluation uses 50 episodes. Final evaluation uses 500 episodes with 10 evaluation environments. The evaluation frequency is 5,000. \\

\midrule

VLA &
Pi0.5 LoRA fine-tuning; number of converted trajectories = 100; training steps = 20,000; batch size = 32; action horizon = 10; input images are resized to $224\times224$; LoRA variants include \texttt{gemma\_2b\_lora} and \texttt{gemma\_300m\_lora}.
&
Evaluation episodes = 100; evaluation seeds = 0, 1, and 2; evaluation chunks = 500; evaluation action horizon = 10. \\

\bottomrule
\end{tabular}
\end{table*}

For the full GTP-FA framework, the closed-loop optimization pipeline additionally includes the failure-attribution discriminator $D$, the grasp-conditioned embedding model $E$, the $D/E$ fusion module, hard-P start screening, and the TaskScore/risk head. Table~\ref{tab:gtpfa_hyperparams} summarizes the shared hyperparameters used across GTP-FA variants.

\begin{table*}[t]
\centering
\caption{
Shared hyperparameters used in GTP-FA.
}
\label{tab:gtpfa_hyperparams}
\small
\setlength{\tabcolsep}{6pt}
\renewcommand{\arraystretch}{1.12}
\begin{tabular}{p{0.24\linewidth} p{0.42\linewidth} p{0.18\linewidth}}
\toprule
\textbf{Module} & \textbf{Hyperparameter} & \textbf{Value} \\
\midrule

Closed-loop optimization & Number of iterations $N_{\mathrm{iters}}$ & 2 \\

\midrule
hard-P injection & Initial hard-P ratio & 0 \\
hard-P injection & Training hard-P ratio $\rho$ after the first iteration & 0.2 \\
hard-P injection & Evaluation hard-P ratio & 0 \\

\midrule
FAD rollout & Number of FAD environments & 512 \\
FAD rollout & Number of repeated trials $K$ per grasp condition & 100 \\

\midrule
Weak-label construction & $\theta_{\mathrm{low}}$ & 0.2 \\
Weak-label construction & $\theta_{\mathrm{high}}$ & 0.8 \\

\midrule
Data split & Split ratio & 0.8 \\
Data split & Split seed & 0 \\

\midrule
Discriminator $D$ & Epochs & 80 \\
Discriminator $D$ & Batch size & 1024 \\
Discriminator $D$ & Learning rate & $3\times10^{-4}$ \\
Discriminator $D$ & Validation ratio & 0.2 \\
Discriminator $D$ & Sampler & weighted \\

\midrule
Embedding model $E$ & Epochs & 200 \\
Embedding model $E$ & Batch size & 256 \\
Embedding model $E$ & Learning rate & $3\times10^{-4}$ \\
Embedding model $E$ & G2 fraction threshold & 0.25 \\
Embedding model $E$ & Success threshold & 0.95 \\

\midrule
$D/E$ fusion & $k$ & 25 \\
$D/E$ fusion & Temperature & 0.2 \\
$D/E$ fusion & Fusion scope & \texttt{g2p\_top2} \\
$D/E$ fusion & Fusion rule & \texttt{switch} \\
$D/E$ fusion & Confidence threshold & 0.97 \\

\midrule
hard-P screening & Top-K & 200 \\
hard-P screening & Minimum count & 5 \\
hard-P screening & $\delta_P$ & 0.65 \\
hard-P screening & $\delta_{G2}$ & 0.25 \\
hard-P screening & $\delta_{G1}$ & 0.20 \\

\midrule
TaskScore / risk head & Epochs & 50 \\
TaskScore / risk head & Maximum samples & 200,000 \\

\bottomrule
\end{tabular}
\end{table*}

Here, $(\theta_{\mathrm{low}}, \theta_{\mathrm{high}})$ are used to construct weak failure-attribution labels, while $(\delta_P,\delta_{G1},\delta_{G2})$ are used to screen planning-dominant hard-P starts from the fused failure-mode probabilities. For PPO and SAC, hard-P injection is implemented as start-state resampling during training. For BC, DP, and VLA, it corresponds to diagnosis-driven restructuring and resampling of training data, rollout data, or fine-tuning data. During standard evaluation, no hard-P starts are injected, ensuring that all methods are compared under the same evaluation distribution.

\subsubsection{Training and Computational Resources}
\label{app:compute_resources}

The hardware platforms differ only in training and inference throughput; all reported results use the same task definitions, ablation settings, evaluation distributions, and success metrics.
PPO, SAC, BC, DP, and their GTP-FA variants are mainly trained on a workstation with $2\times$ RTX 4090 GPUs and an AMD Ryzen 9 9950X CPU. VLA experiments are conducted on a server with $3\times$ A100 GPUs and an AMD EPYC 7453 CPU due to the higher memory and inference cost of VLA fine-tuning. Real-robot experiments are conducted on the Franka Research 3 platform, with policy deployment and perception inference supported by an RTX 4090 workstation. These hardware differences only affect training and inference throughput, and do not change the task definition, evaluation protocol, or success metrics. Table~\ref{tab:compute_resources} summarizes the computational resources used for each experiment type.

\begin{table}[htp!]
\centering
\caption{
Computational resources used in our experiments.
}
\label{tab:compute_resources}
\small
\setlength{\tabcolsep}{5pt}
\renewcommand{\arraystretch}{1.12}
\begin{tabular}{p{0.28\linewidth} p{0.28\linewidth} p{0.34\linewidth}}
\toprule
\textbf{Experiment Type} & \textbf{Methods} & \textbf{Hardware} \\
\midrule
Simulation RL / IL / DP experiments &
PPO, SAC, BC, DP and their GTP-FA variants &
$2\times$ RTX 4090, AMD Ryzen 9 9950X \\

Simulation VLA experiments &
$\pi_{0.5}$-00/01/10/11, GTP-FA-$\pi_{0.5}$ &
$3\times$ A100, AMD EPYC 7453 CPU \\

Real-robot experiments &
Original $\pi_{0.5}$, GTP-FA-$\pi_{0.5}$ &
Franka Research 3 + Robotiq gripper + base/wrist D435i cameras, Intel Xeon Gold 6226R, RTX 4090 \\
\bottomrule
\end{tabular}
\end{table}

\subsubsection{Real-Robot Experimental Protocol}
\label{app:real_robot_protocol}

Real-robot experiments are conducted on a Franka Research 3 robot equipped with a Robotiq gripper, a base Intel RealSense D435i camera, a wrist-mounted Intel RealSense D435i camera, and a VLA policy deployment interface.
Policy inference and perception are supported by an RTX 4090 workstation, as summarized in Appendix~\ref{app:compute_resources}.
For the real-robot setting, we instantiate the base policy with $\pi_{0.5}$ and compare the original $\pi_{0.5}$ baseline with GTP-FA-$\pi_{0.5}$.
Both methods use the same 300 real-robot expert trajectories per task (1,500 in total). This system-level comparison does not isolate failure attribution from the other pipeline components.

We evaluate five real-world manipulation tasks:
(i) placing an orange into a pink tray;
(ii) stacking a blue cube onto an orange cube;
(iii) picking up the red end of a stick and pushing a yellow cube into a red target area;
(iv) picking up the red part of a hook and pulling a yellow cube into a red target area; and
(v) picking up the red handle of a gray cup and pouring the contents into a blue-gray cup.
Each task is evaluated over 50 physical trials, and the task-level success rates are reported in Table~\ref{tab:real_robot_protocol} in the main text.

The original $\pi_{0.5}$ baseline directly predicts actions from visual observations, language instructions, and robot states.
GTP-FA-$\pi_{0.5}$ first performs task-aware grasp selection and then initializes downstream policy execution from the selected grasp-conditioned state.

A trial is counted as successful only if the desired final task condition is satisfied at the end of execution:
the orange must be stably placed inside the pink tray; the blue cube must remain stably stacked on the orange cube after the gripper opens and releases it; the yellow cube must reach the red target area in \textsc{PokeCube} and \textsc{PullCubeTool}; and at least 90\% of the plastic pellets initially contained in the gray cup must be poured into the blue-gray target cup in \textsc{PourWater}.

A trial is counted as a failure if the robot fails to grasp the target object or functional part, drops the object or tool, moves the object to an incorrect location, fails to release the object properly, fails to pour enough contents into the target cup, loses stable control during execution, or terminates in a state that does not satisfy the task goal.
Additional real-world execution interfaces, grasp-planning visualizations, successful rollout sequences, and representative failure cases are provided in Appendix~\ref{app:real_world_results}.

\subsection{Additional Validation of Failure Attribution}
\label{app:additional_validation}

\paragraph{Blind operational audits.}
We evaluated agreement with observable failure responsibilities independently of the model predictions and weak labels. Two authors involved in method and rule design annotated anonymized videos without seeing predictions, probabilities, or weak labels; this is a blind two-author audit, not external validation or a test of complete physical causality. From a disjoint StackCube set of 200 grasp conditions and 20,000 rollouts (5,828 failures), 150 failures spanning 87 conditions were sampled without replacement. The annotators agreed directly on 143/150 cases (95.3\%; Cohen's $\kappa=0.908$). Of these 150 cases, 147 were assigned to the three operational classes and three were Other/Uncertain. On the 147 clear cases, attribution accuracy was 93.9\% and macro-F1 was 90.6\%, with condition-cluster bootstrap 95\% intervals of 89.6--97.4\% and 80.6--95.9\%, respectively. Counting all three uncertain cases as incorrect gives 138/150 (92.0\%) conservative accuracy.

We also audited all 250 held-out GTP-FA real-robot trajectories across the five tasks. Among 58 failures, 51 were clearly classifiable (87.9\% coverage), with 49/51 correct predictions (96.1\% accuracy; 95.9\% macro-F1). Conservative accuracy over all failures was 49/58 (84.5\%). All ten human-labeled G2 cases, comprising six minor slips and four clear drops, were predicted as G2; two planning failures were also predicted as G2 (83.3\% precision and 100\% recall). Other/Uncertain is a human-audit outcome, not a model output. Across the same five tasks, grasp-side failures fell from 161/250 for the original policy to 19/250 for GTP-FA, while planning failures fell from 48/250 to 32/250. This system-level comparison does not isolate attribution from the other components.

\paragraph{Attribution and optimization controls.}
Table~\ref{tab:additional_fa_controls} reports controlled StackCube/PokeCube experiments using seeds $\{0,1,2\}$ with shared evaluation, training budgets, updates, and learner hyperparameters. Shuffled-FA permutes failure--trajectory correspondence while preserving class/confidence distributions and routed/hard-P counts. Reference-FA changes only attribution, using simulator states and operational rules independent of the $D/E$ weak-label construction; it is not a causal oracle. Random-Hard uses the same resampled count and update budget as Full, but selects hard examples randomly. The reductions under Shuffled-FA and Random-Hard support the importance of correct responsibility and targeted exposure. The component effects interact and should not be added as independent contributions.

\begin{table}[htbp]
\centering
\small
\caption{Controlled terminal success (\%) averaged over StackCube and PokeCube. Setting \texttt{11} updates both modules without directed attribution. Single-component removals retain the remaining Full components.}
\label{tab:additional_fa_controls}
\begin{tabular}{lrr}
\toprule
Setting & PPO & BC \\
\midrule
\texttt{00} & 31.8 & 23.0 \\
\texttt{11} & 51.6 & 24.4 \\
Shuffled-FA & 24.0 & 16.1 \\
Reference-FA & 80.4 & 72.0 \\
$D$-only ($\alpha=0$) & 67.7 & 55.3 \\
$E$-only & 47.1 & 21.3 \\
Without risk prediction ($\beta=0$) & 24.5 & 10.9 \\
Random-Hard & 53.7 & 43.9 \\
Without distribution reshaping ($\rho=0$) & 66.3 & 46.8 \\
Without responsibility weighting ($w_P=1$) & 52.7 & 28.9 \\
Full GTP-FA & 76.7 & 66.7 \\
\bottomrule
\end{tabular}
\end{table}

\paragraph{Held-out grasp conditions.}
For each task, 120 Seen and 80 Unseen grasp conditions each generated 100 fresh rollouts. Unseen conditions were excluded from training, validation, and the Seen Embedding Bank. Methods share seeds, splits, rollouts, and initializations. Table~\ref{tab:additional_unseen} evaluates agreement with the predefined weak-label taxonomy; the human audits above separately assess observable failure agreement. Adaptive fusion improves Unseen macro-F1 over $D$-only by 9.6/10.6 percentage points on StackCube/PokeCube while preserving Seen performance. The embedding supplies a condition-level prior and cannot replace direct evidence of the current slip, drop, collision, or alignment event.

\begin{table}[htbp]
\centering
\small
\caption{Failure-attribution macro-F1 (\%) on Seen and condition-level held-out Unseen grasps.}
\label{tab:additional_unseen}
\begin{tabular}{lrrrr}
\toprule
& \multicolumn{2}{c}{StackCube} & \multicolumn{2}{c}{PokeCube} \\
Model & Seen & Unseen & Seen & Unseen \\
\midrule
$D$-only & 95.2 & 80.9 & 94.3 & 78.9 \\
$E$-only & 73.9 & 68.6 & 69.6 & 64.1 \\
Fixed $D+E$ & 90.5 & 85.4 & 93.1 & 87.9 \\
Adaptive $D+E$ & 95.4 & 90.5 & 94.3 & 89.5 \\
\bottomrule
\end{tabular}
\end{table}

\paragraph{Downstream demonstration utilization.}
BC data-budget controls use nested subsets of 50, 100, and 200 independent trajectories per task, with the same update-to-data ratio. The pretrained $D/E$, Seen Bank, risk predictor, hard-P rule, evaluation, seeds, and learner hyperparameters are shared. Uniform retains attribution, risk prediction, hard-P identification, and responsibility weighting, but disables reshaping ($\rho=0$); GTP-FA uses $\rho=0.2$. Table~\ref{tab:additional_budget} shows improved utilization of independent downstream demonstrations. At 50\% budget, indiscriminately resampling all failures (All-Hard) reaches 34.6\%, below Uniform's 36.4\%. These controls do not establish lower total diagnostic interaction, environment interaction, training cost, or compute.

\begin{table}[htbp]
\centering
\small
\caption{BC terminal success (\%) averaged over StackCube and PokeCube under matched data-budget protocols. The 100\% Uniform setting is the no-reshaping control in Table~\ref{tab:additional_fa_controls}.}
\label{tab:additional_budget}
\begin{tabular}{lrrr}
\toprule
Budget & Independent trajectories/task & Uniform & GTP-FA \\
\midrule
25\% & 50 & 27.5 & 39.8 \\
50\% & 100 & 36.4 & 49.6 \\
100\% & 200 & 46.8 & 66.7 \\
\bottomrule
\end{tabular}
\end{table}

\paragraph{Hard-P screening sensitivity.}
The screening thresholds in Appendix~\ref{app:hardP_def} are shared across main tasks and learners. With matched seeds, data, and evaluation initializations, loose/default/strict $(\delta_P,\delta_{G1},\delta_{G2})$ settings of $(0.55,0.30,0.35)$, $(0.65,0.20,0.25)$, and $(0.75,0.15,0.20)$ yield average StackCube/PokeCube terminal success of 72.6/76.7/74.5\%, respectively. The default balances coverage and grasp-side contamination in these experiments; new platforms or distributions may require recalibration.

\FloatBarrier

\subsection{Limitations and Broader Impacts}
\label{app:limitations_impacts}

\paragraph{Limitations.}
Our evidence concerns task-oriented, two-finger grasp-then-plan manipulation in which the selected grasp materially affects downstream feasibility. It does not establish general regrasping, dexterous in-hand manipulation, deformable-object manipulation, or force-controlled contact. The tool-use and pouring results extend the tested task set but do not remove these boundaries. GTP-FA shares diagnosis and optimization routing across learners, while retaining their native objectives; the current offline branch mainly reshapes available data and does not provide online-RL-equivalent planning-side out-of-distribution generation.

FM-G1, FM-G2, and FM-P are operational responsibility categories. Mixed failures may update both grasp risk and downstream learning through soft probabilities, but only high-planning/low-grasp-responsibility conditions receive strong hard-P routing. Discarding or soft-labeling the ambiguous interval during weak-label construction does not implement inference-time rejection. The current model has no Unknown/Other output or open-set detector; severe perception, hardware, or communication failures may be projected into the three classes and misdirect updates.

Real inference uses base/wrist RGB-D sequences, grasp pose, robot/gripper states, logs, relative gripper--object motion, and observable terminal outcomes, without simulator truth or tactile sensing. Severe occlusion, low texture, depth noise, limited frame rate, or unobservable outcomes may weaken diagnosis. The audits in Appendix~\ref{app:additional_validation} measure agreement with observable operational failures and were performed by authors involved in method design. They do not establish complete physical causality or independent external validation. Broader platform testing, open-set rejection, and tactile/force feedback remain future work.

\paragraph{Broader impacts.}
The proposed framework may improve the reliability and sample efficiency of robotic manipulation systems by reducing failure misattribution and unsafe trial-and-error. Potential negative impacts include unsafe deployment in physical environments, unintended object damage, or misuse in automated manipulation systems without appropriate supervision. In our experiments, all real-robot trials are conducted in a controlled laboratory setting with human supervision, limited tabletop objects, and no human subjects.

\paragraph{Existing assets and licenses.}
Our experiments build on publicly available research software and assets, including ManiSkill3 environments, GraspNet-style grasp generation, and open-source implementations of the base learners when available. We cite the corresponding papers in the main text and follow their released licenses and terms of use. The real-robot data used in this work are collected by the authors in a controlled laboratory setting.

\end{document}